\documentclass[journal]{IEEEtran}

\usepackage{amsmath}
\usepackage{amsfonts}
\usepackage{amsthm}
\usepackage{amsbsy}

\usepackage{cite}

\usepackage{todonotes}
\usepackage{xcolor}
\usepackage{tikz}
\usetikzlibrary{shapes,calc}

\usepackage{multicol}
\usepackage[ruled]{algorithm2e}
\usepackage{lipsum} 
\usepackage{mathtools}
\usepackage{ifthen}
\usepackage{subcaption}

\newcommand{\der}[2]{\frac{\partial #1}{\partial #2}}

\newcommand{\lder}[2]{\partial #1 / \partial #2 }

\newcommand{\mbf}{\mathbf}
\renewcommand{\r}{\mathbb{R}}
\newcommand{\esp}[1]{\mathbb{E}\left\{ #1 \right\}}
\newcommand{\expect}{\mathbb{E}}
\newcommand{\trace}[1]{\mathrm{Tr}\left\{ #1 \right\}}

\newcommand{\covar}{\mathsf{cov}}

\newcommand{\Fu}{\mbf F_\mathcal{U}}

\newcommand{\col}{\text{col}}

\newcommand{\cpop}[1]{\left[#1\right]_{\times}}

\newcommand{\FOR}{\mathfrak{F}}

\newcommand{\sdim}{n}

\newcommand{\spanv}[1]{\mathrm{span}\left\{ #1 \right\}}

\newcommand{\jln}[1]{\textcolor{red}{#1}}

\newcommand{\jcm}[1]{\textcolor{brown}{#1}}

\newtheorem{proposition}{Proposition}
\newtheorem{corollary}{Corollary}
\newtheorem{definition}{Definition}
\newtheorem{theorem}{Theorem}
\newtheorem{remark}{Remark}

\newcounter{isdiff}

\ifthenelse{\value{isdiff}>0}
{
\newcommand{\diffblock}{\color{blue}}
\newcommand{\diff}[1]{\textcolor{blue}{#1}}
\newcommand{\diffcaption}[1]{\caption{{\color{blue}#1}}}
}
{
\newcommand{\diffblock}{\color{black}}
\newcommand{\diff}{\textcolor{black}}
\newcommand{\diffcaption}[1]{\caption{#1}}
}
\newcommand{\diffend}{\color{black}}

\renewcommand{\jln}{\diff}
\renewcommand{\jcm}{\diff}

\begin{document}

\title{\diff{Ranging-Based Localizability Optimization\\ for Mobile Robotic Networks}}

\author{Justin~Cano,~\IEEEmembership{Member,~IEEE,}
        and~Jerome~Le~Ny,~\IEEEmembership{Senior~Member,~IEEE}
\thanks{This work was supported by FRQNT under grant 2018-PR-253646 
and by NSERC under grant RGPIN-5287-2018.}
\thanks{The authors are with the Department of Electrical Engineering, 
Polytechnique Montreal, and with GERAD, Montreal, QC H3T 1J4, Canada 
{\tt\small \{justin.cano,jerome.le-ny\}@polymtl.ca}.}
\thanks{Preliminary versions of this paper appeared in \cite{le_ny_jerome_localizability-constrained_2018} and \cite{Cano:ICRA21:constrainedCRLB}.} 
}

\markboth{Preprint Version \#2}%
{}

\maketitle

\begin{abstract}
\diff{In robotic networks relying on noisy range measurements between agents for cooperative localization}, 
the achievable positioning accuracy strongly strongly depends on the network geometry.
This motivates the problem of planning robot trajectories in such multi-robot systems in a way that maintains 
high localization accuracy. We present potential-based planning methods, where localizability potentials are 
introduced to characterize the quality of the network geometry for cooperative position estimation. 
These potentials are based on Cram\'er Rao Lower Bounds (CRLB) and provide a theoretical 
lower bound on the error covariance achievable by any unbiased position estimator.
In the process, we establish connections between CRLBs and the theory of graph rigidity, which has been previously 
used to plan the motion of robotic networks. 
We develop decentralized deployment algorithms appropriate for large networks,
and we use equality-constrained CRLBs to extend the concept of localizability to scenarios 
where additional information about the relative positions of the ranging sensors is known. 
We illustrate the resulting robot deployment methodology through simulated examples \diff{and an experiment}.
\end{abstract}


\IEEEpeerreviewmaketitle

\section{Introduction}

\IEEEPARstart{M}{obile} robots require accurate, computationally efficient and low power 
localization systems to navigate their environment and perform their assigned tasks. 
Positioning can rely on various technologies, e.g., wheel odometry, computer vision 
or long- and short-range radio frequency (RF) systems, each with 
distinct advantages and drawbacks, depending on the environment and requirements.
For example, the most common methods of terrestrial localization rely on RF signals from 
Global Navigation Satellite Systems (GNSS) to achieve meter- to centimeter-level accuracy, 
but these systems do not operate indoors or when the line of sight to the satellites is 
obstructed, and are sensitive to interference.

Multiple robots can collaborate to improve the accuracy and coverage of their 
individual localization solution \cite{sheu_distributed_2010,Prorok:ICRA12:relativeLoc}. 
In particular, they \diff{can leverage information about their proximity to other location-aware nodes} 
\cite{sheu_distributed_2010} or use relative position \cite{Prorok:ICRA12:relativeLoc}, bearing \cite{xu_aoa_2008} 
or distance measurements \cite{wei_noisy_2015,carlino_robust_2018} between them to estimate 
their individual positions in a common reference frame.
Relative bearing measurements can be provided by monocular cameras for example, 
range measurements by short-range RF systems, and relative position measurements 
by LiDARs or stereo cameras. In this paper, we focus on collaborative localization 
in Multi-Robot Systems (MRS) \emph{using only range measurements}. This is motivated by the 
fact that accurate distance measurements can be deduced from Time-of-Flight (ToF) 
measurements obtained from inexpensive short-range RF communication systems, 
\diff{e.g.}, Ultra-Wide Band (UWB) transceivers \cite{sahinoglu_ultra-wideband_2008,mueller_fusing_2015,cano_kalman_2019}.
In particular, such systems associate distance measurements unambiguously with
pairs of robots, simply by having the robots broadcast their IDs. 

Once the robots have measured their relative distances, many algorithms exist to
compute from these measurements an estimate of the robot positions, see, e.g., 
\cite{Buehrer:IEEE18:collaborativeLocSurvey} for a recent survey.
These algorithms can be centralized or decentralized, applicable to static or 
mobile networks, appropriate \diff{or not} for real-time localization, etc. 
Two major factors determine the ability of these 
algorithms to solve the position estimation problem and their accuracy.
First, enough relative distance measurements should be available, which links 
the feasibility of the location estimation problem to the concept of \emph{rigidity} \cite{tay_generating_1985,cao_ratio--distance_2020,Aspnes:TMC06:theoryLoc}
of the \emph{ranging graph} corresponding to these measurements.
Second, satisfying the graph-theoretic condition of rigidity is still insufficient 
to guarantee accurate localization of the individual agents, when measurement noise
is inevitably present. For example, a group of robots that are almost aligned can 
form a rigid formation if enough range measurements are available, but can only 
achieve poor localization accuracy in practice. Indeed, the spatial \emph{geometry} 
of the network strongly influences the accuracy of position estimates in the presence 
of measurement noise \cite{patwari_locating_2005}, a phenomenon known 
as Dilution of Precision (DOP) in the navigation literature \cite[Chap. 7]{groves_principles_2013}.
We call here \textit{localizability} the ability to accurately estimate the positions 
of the individual robots of an MRS in a given geometric configuration, 
using relative measurements.

In contrast to static sensor networks or GNSS, an MRS can actively adjust its geometry, 
\textit{e.g.}, some of the robot positions and orientations, in order to improve
its overall localizability. This results in a coupling between the motion planning
and localization problem for the group.
Maintaining the rigidity of the ranging graph during the motion 
of an MRS is a stronger condition than maintaining its connectivity, 
but similar techniques can be used to address both problems. 
In particular, we can capture the degree of connectivity or rigidity of the graph 
using a function of the first non-zero eigenvalue of a type of Laplacian matrix,
and guide the MRS along paths or configure its nodes in ways that increase this function.
This is the approach adopted for example 
in \cite{Kim:TAC06:maximizingEigenvalue,siciliano_maintaining_2009,decentralized_2010}   
for improving connectivity and in \cite{Shames:Automatica09:LocMinimization,zelazo_rigidity_2012,zelazo_decentralized_2015,sun_distributed_2015}
for improving rigidity.
This article builds on this principle to optimize localizability. 
Following an approach that we initially proposed in 
\cite{le_ny_jerome_localizability-constrained_2018, Cano:ICRA21:constrainedCRLB},
we leverage Cram\'er Rao Lower Bounds (CRLBs) \cite[Chap. \diff{14}]{haug_bayesian_2012} to construct 
localizability potentials, which can then be used as artificial potentials 
\cite{choset_principles_2005} to drive the motion of an MRS toward geometric configurations 
promoting good localization.

The CRLB provides a lower bound on the covariance of any unbiased position estimate 
constructed from the relative range measurements available in the robot network.
Tighter covariance lower bounds exist, such as Barankin bounds \cite{mcaulay_barankin_1971}, 
but an advantage of the CRLB is that it is relatively easy to compute and admits 
a closed-form expression for the problem considered here, 
assuming Gaussian noise \cite{patwari_locating_2005}. 
Moreover, as we show in Section \ref{sec:closed_form_crlb}, the CRLB for Gaussian
noise is in fact closely related to the so-called \emph{rigidity matrix} of the ranging graph. 
This \diff{does not come as a surprise}, 
since the Gaussian CRLB is known to correspond to DOP expressions
for least-squares estimators, which are implicitly derived in \cite{Shames:Automatica09:LocMinimization} for example and also linked to the rigidity matrix.
The CRLB only provides a lower bound on estimation performance and there is generally 
no guarantee that a position estimator actually achieves it.
Nonetheless, using this bound as a proxy to optimize sensor placement is a well accepted 
approach \cite{Ucinski:book04:optimalSensing}.
An important advantage of this approach is that the motion planning strategy becomes
independent of the choice of position estimator implemented in the network.

\diffblock
\textbf{Contributions:} First, this paper formulates a novel motion planning problem allowing 
an MRS to optimize its localizability, by minimizing appropriate cost functions 
based on the Fisher Information Matrix (FIM) appearing in the CRLB.
Second, we establish an explicit connection between localizability and the weighted rigidity matrices 
introduced in \cite{zelazo_decentralized_2015,sun_distributed_2015}. One of the benefits
of establishing this connection is to see that various artificial potentials can be 
constructed from the FIM to capture localizability, as discussed in the literature on optimal 
experimental design \cite{pukelsheim_optimal_2006} or optimal sensing with mobile robots, see, 
e.g., \cite{Ucinski:book04:optimalSensing,LeNy:CDC09:activeSensingGP,carrillo_comparison_2012}.
Some of these functions may be more conveniently optimized than the smallest nonzero 
eigenvalue, which is the standard potential used for connectivity and rigidity maintenance.
Third, by leveraging the structure of the FIM matrix, we propose new distributed algorithms 
enabling the deployment of groups of robots carrying ranging sensors in a scalable and robust manner. 
Fourth, we extend the results to robots carrying multiple ranging sensors, using the theory 
of \emph{constrained} CRLBs \cite{hero_lower_1990} to account for the presence of additional 
rigidity constraints. 
This can be viewed as an alternative and simpler approach to deriving intrinsic
CRLBs on the manifold of rigid motions \cite{bonnabel_intrinsic_2015,chirikjian_wirtinger_2018}.

\diffblock
The structure of the paper is as follows. First, we define the deployment problem in Section \ref{sec:statement},
including localizability potentials further discussed in Section \ref{sec:loca_potentials}. 
Then, we derive in Section \ref{sec:closed_form_crlb} the closed-form expression for the FIM 
and analyze its structure, which allows us to introduce in Section \ref{sec:gradients} 
decentralized methods to estimate the gradients of the localizability potentials. 
Section \ref{sec:extensions_rigidity} extends the analysis to the case of robots carrying multiple ranging sensors. 
The deployment algorithms are validated in two simulated scenarios in Section \ref{sec:sim_standard},
and experimental results using RF range measurements from UWB transceivers are described in Section \ref{sec:exp}.
\diffend

This article builds on the conference paper \cite{le_ny_jerome_localizability-constrained_2018}, 
which introduced the concept of localizability potentials for the deployment of MRS 
in two dimensions. Here we extend the methodology to three dimensions, introduce new 
distributed optimization schemes, discuss useful properties on the FIM and make a clearer 
connection with rigidity theory. 
We also generalize the conference paper \cite{Cano:ICRA21:constrainedCRLB}, 
which considered robots carrying multiple sensors, by developing the results
in three dimensions and integrating the full relative position information
in the CRLB rather than just relative distances, which is significantly more challenging.
We demonstrate in simulation the improvement achievable with this extension.

\textbf{Notation:} We write vectors and matrices with a bold font.
The all-one vector of size $p$ is denoted $\mbf 1_p$.
The notation $\mbf x = \col(\mbf x_1,\ldots,\mbf x_n)$ means that the
vectors or matrices $\mbf x_i$ are stacked on top of each other,
and $\text{diag}(\mbf A_1,\ldots,\mbf A_k)$ denotes a block diagonal matrix
with the matrices $\mbf A_i$ on the diagonal.
The nullspace of a matrix $\mbf A$ is denoted $\ker \mbf A$.
For $\mbf A$ and $\mbf B$ symmetric matrices of the same dimensions, 
$\mbf A \succeq \mbf B$ means that $\mbf A - \mbf B$ is positive semidefinite 
and $\mbf A \succ \mbf B$ that it is positive definite. 
If $\mbf A$ is a symmetric matrix, $\lambda_{\min}(\mbf A)$ and $\lambda_{\max}(\mbf A)$ denote 
its minimum and maximum eigenvalues.
The time derivative of a vector-valued function $t \mapsto \mbf x(t)$ is denoted $\dot{\mbf x}$.
The expectation of a random vector $\mbf x$ is denoted $\expect[\mbf x]$ and its covariance matrix 
$\covar[\mbf x] = \expect \left[ \left(\mbf x - \expect \left[\mbf x \right] \right)
\left(\mbf x - \expect \left[\mbf x\right] \right)^T\right]$. 
For a differentiable function $f: \mathbb R^p \to \mathbb R^q$, $\frac{\partial f(\mbf p)}{\partial \mbf p}$ 
represents the $q \times p$ Jacobian matrix of $f$, with components $\partial f_i(\mbf p)/\partial p_j$ 
for $1 \leq i \leq q$, $1 \leq j \leq p$.
When $q = 1$, $\partial^2 f(\mbf p)/\partial \mbf p \partial \mbf p^T$ denotes the Hessian, 
i.e., the square matrix with components $\partial^2 f(\mbf p)/\partial p_i \partial p_j$.
\diff{Finally, $\mathsf{1}_{\mathsf e}$ is equal to $1$ if the logical expression $\mathsf e$ 
is true and $0$ otherwise, and for a set $\mathcal S$ we also use the alternative notation
$\mathsf{1}_{\mathcal S}(i) \coloneqq \mathsf{1}_{i \in \mathcal S}$.
}

\section{Problem Statement}
\label{sec:statement}

\begin{figure}
    \centering
 	\includegraphics{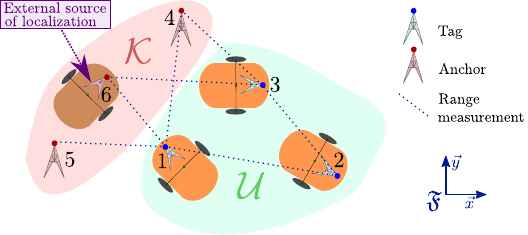}
 	\caption{Illustration of the setup in 2D with 3 mobile tags and 3 anchors, 2 of whom are fixed. 
 	The links for the ranging pairs are shown. The ranging graph includes $3$ additional implicit 
 	links between the anchors, not shown.}
 	\label{fig:setup_deployment}
 	\vspace{-5mm}
\end{figure}
 
Consider a set of $N$ nodes in the $\sdim$-dimensional Euclidean space, where $\sdim = 2$ or 
$\sdim = 3$. We fix a global reference frame  denoted $\mathfrak{F}=(O, \vec{x}, \vec{y}, \vec{z})$ 
if $\sdim = 3$ or $\mathfrak{F}=(O, \vec{x}, \vec{y})$ if $\sdim = 2$.
For $1 \leq i \leq N$, we write the coordinates of node $i$ in that frame 
$\mbf p_i := [x_i,y_i,z_i]^\top$ if $n = 3$ or $\mbf p_i := [x_i,y_i]^\top$ if $\sdim = 2$,
and we let $\mbf p \coloneqq \text{col}(\mbf p_1, \ldots, \mbf p_N) \in \mathbb R^{nN}$ denote 
the global spatial configuration of the nodes, which can vary with time.

As illustrated on Fig. \ref{fig:setup_deployment}, some of these nodes are carried 
by mobile robots, while others could remain at fixed locations.
We suppose that the coordinates of a subset $\mathcal{K}$ of the nodes are 
perfectly known in $\mathfrak{F}$, for $1 < |\mathcal{K}|:=K < N$, 
and refer to these nodes as \emph{anchors}.  
The anchors could be placed at fixed locations or they could be mobile, as long 
as we can precisely localize them via external means, e.g., using accurate GNSS receivers.
The other nodes, also mobile or fixed and whose positions are unknown and need 
to be estimated, are called \emph{tags} in the following. They form a set denoted 
$\mathcal{U}$, with $|\mathcal{U}|:=U=N-K$. 

Next, we assume that $P$ pairs of nodes, called ranging pairs, can measure their distance
(with each such pair containing at least one tag). 
For a \diff{ranging} pair of nodes $(i,j)$, we denote $d_{ij}$ the true distance between the nodes 
and \linebreak $\tilde d_{ij}$ a corresponding measurement, to which both nodes $i$
and $j$ have access. In the following, we consider measurement models assuming either 
additive Gaussian noise
\begin{equation}
\tilde{d}_{ij} = d_{ij}+\nu_{ij}, \; \nu_{ij} \sim \mathcal{N}(0,\sigma^2), 
\label{eq:model_meas}
\end{equation}
or multiplicative log-normal noise
\begin{equation}
\tilde{d}_{ij} = d_{ij} \, e^{\mu_{ij}}, \; \mu_{ij} \sim \mathcal{N}(0,\bar{\sigma}^2),
\label{eq:model_meas_lognormal}
\end{equation}
where the noise realizations $\nu_{ij}$ or $\mu_{ij}$ are independent for all $i, j$ \diff{and $\sigma^2,\bar{\sigma}^2 \in \r^+$ are given covariances}.
We collect all the measured distances $\tilde{d}_{ij}$ \diff{at a given time} in the vector 
$\tilde{\mbf d}=[\dots,\tilde{d}_{ij}, \dots]^\top \in \r^{P}$.
We also define an undirected graph $\mathcal{G}=(\mathcal{E},\mathcal{V})$, 
called the \emph{ranging graph}, whose vertices $\mathcal V$ are the $N$ nodes 
and with an edge in $\mathcal E$ for each ranging pair and for each pair of anchors. 
In particular, the subgraph of $\mathcal G$ formed by the anchors
is a complete graph, which is consistent with the fact that the distances
between anchors are implicitly known from their coordinates.
Two nodes linked by an edge in $\mathcal G$ are called neighbors and we denote by
$\mathcal N_i$ the set of neighbors of $i$ or \emph{neighborhood} of $i$, for $1 \leq i \leq N$.
Let $E = P + \frac{K(K-1)}{2}$ be the total number of edges in $\mathcal G$.

A concrete implementation of the previous \diff{system} 
is as follows.
The nodes could correspond to RF transceivers capable of measuring
their distance with respect to other nodes within their communication radius.
Radiolocation protocols such as Two-Way Ranging (TWR), Time of Arrival (ToA) 
or Time Difference of Arrival (TDoA) \cite{sahinoglu_ultra-wideband_2008,Bensky:book16:wirelesPos} 
use the timestamps of messages exchanged by the transceivers to estimate the ToF of these messages
and deduce distance measurements, which can be assumed to be of the form \eqref{eq:model_meas},
at least under line-of-sight signal propagation conditions.
Another ranging method consists in measuring the strength of a received signal (RSS) 
to deduce the distance to the transmitter using a path loss propagation model \cite{Bensky:book16:wirelesPos}.
This method typically leads to a distance measurement model of the
form \eqref{eq:model_meas_lognormal}, assuming again a simple radio 
propagation environment \cite{coulson_statistical_1998,patwari_locating_2005}.

We assume that the nodes implement a cooperative localization scheme, in order to jointly 
produce an estimate $\hat{\mbf p}$ of all their coordinates $\mbf p$ in $\FOR$, based 
on the noisy measurements $\tilde{\mbf d}$ and the knowledge of the anchor coordinates.
As we explain in Section \ref{sec:loca_potentials}, the value of $\mbf p$ itself
strongly influences the achievable accuracy of its estimate. 
Hence, we introduce in that section some real-valued functions 
$J_\text{loc}:\r^{\sdim N} \to \r$ that can serve as \emph{localizability potentials}, 
i.e., such that a \diff{low value (resp. high value)} for $J_\text{loc}(\mbf p)$ means 
that the performance  
of an estimator at configuration $\mbf p$ is expected to be \diff{good (resp. bad)}. 
A localizability potential can then serve as an artificial potential for motion 
planning \cite{choset_principles_2005}, to guide or constrain the motion of an MRS 
to configurations that are favorable for accurate cooperative localization.
\diff{Concretely, consider a potential function $J(\mbf p) = \alpha J_1(\mbf p) + (1-\alpha) J_\text{loc}(\mbf p)$,
for $\alpha \in (0,1)$,
where $J_1$ may include attractive and repulsive potentials to steer robots toward desired 
locations \cite{Khatib:art86:potentials} and away from obstacles \cite{choset_principles_2005}, 
to maintain network connectivity \cite{decentralized_2010}, to cover an area \cite{Bullo:book09:distributedRobotics}, etc.
}
One can then generate a sequence of configurations $\mbf p(0),\mbf p(1), \dots,$ for the MRS by following 
the gradient descent scheme
\begin{equation}    \label{eq:descent_per_agent}
\mbf p_{i,k+1} = \mbf p_{i,k} - \gamma_k \left( \der{J(\mbf p_k)}{\mbf p_{i}} \right )^T,
\end{equation}
for each mobile node $i$, with $\{\gamma_k\}_{k \geq 0}$ a sequence of appropriate stepsizes.
\diff{The presence of $J_\text{loc}$ in the overall potential favours configurations that
have higher localizability, and this effect becomes more pronounced as $\alpha$ increases.
Alternatively, one can also minimize $J_1$ subject to a constraint on the maximum tolerable
value of $J_\text{loc}$.}
\diff{Note however that as in most cases where artificial potentials are used to plan the motion
of an MRS, the gradient descent scheme \eqref{eq:descent_per_agent} typically only leads to locally
optimal configurations.}

A key issue when relying on artificial potentials to provide goal configurations to an MRS is 
to ensure that each mobile node $i$ can compute the gradient 
$\left(\partial J_{\text{loc}}(\mbf p(k))/\partial \mbf p_i\right)^T$ with respect to its coordinates 
in \eqref{eq:descent_per_agent} by exchanging information only with its immediate neighbors 
in the communication network, \emph{which we assume here to coincide with the ranging graph}
(although in general the anchors will not need to communicate with each other).
This ensures scalability to large networks and improves the robustness of
the network against the loss of nodes. The design of distributed gradient descent schemes 
for the localizability potentials is discussed in Section \ref{sec:gradients}.

In summary, the problem considered in this paper is to first define appropriate
functions that can serve as localizability potentials and then design distributed
gradient descent algorithms for these potentials in order to deploy an MRS with
ranging sensors while ensuring that its cooperative localization scheme \diff{remains precise.} 
In addition, we show in Section \ref{sec:extensions_rigidity} how to adapt the definition 
of the localizability potentials and the gradient descent scheme to a more complex
situation where multiple tags can be carried by the same robot. This introduces 
additional constraints on the positions $\mbf p$, which should be taken into account 
by localization and motion planning algorithms.
\diff{These constraints} can be used in practice to provide more accurate full pose estimates for 
the robots. 

\begin{remark}  \label{rmk: estimate in the loop}
In practice, the tags have access to their position $\mbf p$ only through their estimates $\hat{\mbf p}$. 
As a result, \diff{when using artificial potentials for motion planning},
the gradient descent scheme \eqref{eq:descent_per_agent} cannot be directly implemented, 
and the standard approach is to compute and follow the gradient at the current estimate, 
i.e., use $\partial J(\hat{\mbf p}_k)/\partial \mbf p_i$ in \eqref{eq:descent_per_agent}.
\diff{Since including a localizability potential aims to improve the accuracy of the position
estimates along the robots' paths, it contributes to making this approximation of
ignoring position uncertainty less problematic.} 
Alternatively, \eqref{eq:descent_per_agent} can also be used to compute a sequence of steps, 
i.e., plan a future trajectory for the MRS, in which case we assume at the planning stage
that the agents will be able to track that trajectory perfectly.
\diff{Moreover, we empirically study the behavior of the scheme \eqref{eq:descent_per_agent}
with gradients evaluated at the imperfect position estimates, 
both in simulations 
in Section \ref{sec:sim_standard} and through experiments in Section \ref{sec:exp}.
In particular, our experiment confirms the intuitive fact that enhancing the localizability
is important to ensure that the robots are able to reliably follow their desired trajectories.
} 
\end{remark}
 
\diffblock
\begin{remark}
In general, the ranging graph $\mathcal{G}$ could change over time as nodes move in their 
environment. In this case, the algorithms presented later could still be implemented 
at each period over the current ranging graph, but localizability could become poor 
if critical ranging pairs become disconnected. 
To address this issue, ranging between specific pairs can be maintained by adding 
connectivity potentials to the function $J_1$ above. Alternatively, when we use
the model \eqref{eq:model_meas_lognormal} or alternative models where the
variance degrades with distance \cite{Cano:IROS22:distanceCRLB},
then $J_{loc}$ increases when the links become longer, a consequence of 
the result \eqref{eq:ourfim} stated in the next section.
Hence, in a manner similar to the use weighted graph models for 
connectivity \cite{Kim:TAC06:maximizingEigenvalue} and rigidity 
\cite{zelazo_decentralized_2015}, maintaining ranging distance 
between nodes can be promoted directly through the localizability potential.
\end{remark}
\diffend

\section{Localizability Potentials}
\label{sec:loca_potentials}

This section is concerned with defining artificial potentials that can be used
as localizability potentials. The proposed definitions require that we first
recall some \diff{notions} 
from estimation theory related to the CRLB. 
 
\subsection{Constrained Cram\'er-Rao Lower Bound}
\label{ss:pstat_crlb}

We assume that the position estimator implemented by the MRS is unbiased, i.e., satisfies
$\expect[\hat{\mbf p}] = \mbf p$. We then focus on finding configurations $\mbf p$
for which the error covariance matrix 
$\expect \left[ (\hat{\mbf p} - \mbf p)(\hat{\mbf p} - \mbf p)^\top \right]$ for $\hat{\mbf p}$, 
which is then also the covariance matrix $\covar[\hat{\mbf p}]$,
is ``small'' in some sense.
More precisely, since the error covariance depends on the specific estimator used 
and can be difficult to predict analytically, 
we use the CRLB, a lower bound on the covariance of any unbiased estimator, to quantify 
the quality of a configuration $\mbf p$.
Although this implicitly assumes that an estimator can be constructed to achieve or approach
this lower bound, this methodology is commonly used in optimal experiment design
and sensor placement \cite{pukelsheim_optimal_2006,Ucinski:book04:optimalSensing}.
In general, the CRLB corresponds to the inverse of the Fisher Information Matrix (FIM), 
which we define below.
 
\begin{definition}[FIM]     \label{def: FIM}
Let $\mbf x \in \r^p$ be a deterministic parameter vector and $\mbf y \in \r^q$ 
a random observation vector, for some positive integers $p,q$.  
Define $f : \r^q \times \r^p \to \r^+$ the Probability Density Function (PDF) of $\mbf y$,
which depends on the parameter $\mbf x$, so that we write $f(\mbf y;\mbf x)$. 
Under some regularity assumptions on $f$ (see \cite[Chap. 14]{haug_bayesian_2012}), 
the ${p\times p}$ Fisher Information Matrix (FIM) of this PDF is defined as
\begin{equation}    \label{eq: FIM def}
\mbf F(\mbf x) = - \expect_{\mbf y} \left[ \der{^2 \ln f(\mbf y;\mbf x)}{\mbf x \partial \mbf x^\top} \right].
\end{equation}
The matrix $\mbf F(\mbf x)$ is symmetric and positive semi-definite.
\end{definition} 

In the position estimation problem, the parameters of interest are the node coordinates 
in the vector $\mbf p \in \r^{\sdim N}$, 
whereas the random observations are contained in the vector $\tilde{\mbf d}$.
As computed in \cite{patwari_locating_2005},
the FIM of the PDF $f(\tilde{\mbf d};\mbf p)$ is an $\sdim N\times \sdim N$ matrix 
that depends on $\mbf p$ and can be decomposed into $\sdim \times \sdim$ blocks
$\mbf F_{ij}$ such that 
\begin{align}
\begin{split}
\mbf F_{ij}(\mbf p) &= \mbf F_{ij}(\mbf p_{ij}) = - \frac{1}{d_{ij}^{2 \kappa}\sigma^2} \mbf p_{ij} \mbf p_{ij}^\top \, 
\mathsf{1}_{\mathcal N_i}(j), \text{ if } i\neq j, \\
\mbf F_{ii}(\mbf p) &= -\sum_{j \neq i} \mbf F_{ij},
\end{split}
\label{eq:ourfim}
\end{align}
where $\mbf p_{ij} \coloneqq \mbf p_i - \mbf p_j$, 
and $\kappa=1$ for the additive noise model \eqref{eq:model_meas} 
or $\kappa=2$ for the multiplicative noise model \eqref{eq:model_meas_lognormal}. 
\diff{The result \eqref{eq:ourfim} can be obtained using the Slepian-Bangs 
formula \cite[Section 3.9]{kay_fundamentals_1993} or by direct calculation.}
 
Note however that estimating the anchor positions is not needed, since the locations 
of these nodes are known. The fact that $\hat{\mbf p}_i := \mbf p_i$ for all $i \in \mathcal{K}$, 
with 
$\mbf p_i$ known, should be taken into account by an estimator of the 
tag positions, and hence should also be taken into account when bounding the covariance of these 
estimators. We can rely on the theory of CRLBs with equality constraints on the estimated 
parameters in order to include these trivial constraints on the anchor positions and later 
in Section \ref{sec:extensions_rigidity} also additional rigid constraints on the tag positions.

\begin{theorem}[Equality constrained CRLB \cite{hero_lower_1990}] \label{thm:gorman}
Let $\mbf x \in \r^p$ be a deterministic parameter vector and $\mbf y \in \r^q$ 
a random observation vector, for some positive integers $p,q$. 
Let $\mbf h: \r^p \to \r^c$, for $c\leq p$, be a differentiable function such that 
$\mbf h(\mbf x)= \mbf 0$. 
Let $\hat{\mbf x}$ be an unbiased estimate of $\mbf x$ also satisfying $\mbf h(\hat{\mbf x}) = \mbf0$ 
and with finite covariance matrix. 
Define $\mbf F_c \coloneqq \mbf A^\top \mbf F \mbf A$, the \emph{constrained Fisher Information Matrix}, 
where $\mbf A$ is any matrix whose columns span $\ker \der{\mbf h}{\mbf x}$,
and $\mbf F$ is the FIM defined in \eqref{eq: FIM def}.
Then, the following inequality holds
\begin{equation}
\covar[\hat{\mbf x}] \succeq 
\mbf A \left(\mbf F_c\right)^\dagger \mbf A^\top =: \mbf B_c
\label{eq:crlb}
\end{equation}
where $\dagger$ denotes the Moore-Penrose pseudo-inverse \cite[p. 21]{petersen_matrix_2012}.
\end{theorem}

Consider now the problem of estimating the vector of tag coordinates $\mbf p_{\mathcal U} \in \r^{nU}$ 
based on the distance measurements $\tilde{\mbf d}$ and knowledge of the anchor coordinates
$\mbf p_{\mathcal K} \in \r^{nK}$. 
Order the nodes so that $\mbf p = \text{col}(\mbf p_{\mathcal U}, \mbf p_{\mathcal K}$),
and partition the FIM defined in \eqref{eq:ourfim} accordingly as
\begin{equation}    \label{eq: FIM partitioning}
\mbf F = \begin{bmatrix}
\mbf F_\mathcal{U} &  \mbf F_\mathcal{UK} \\
\mbf F_\mathcal{UK}^\top & \mbf F_\mathcal{K}
\end{bmatrix}, 
\end{equation}
with in particular $\mbf F_\mathcal{U}$ a symmetric positive semi-definite matrix
of size $nU \times nU$. We then have the following result.

\begin{proposition} \label{prop: constrained CRLB for network}
Let $\hat{\mbf p}_{\mathcal U}$ be an unbiased estimate of the tag positions $\mbf p_{\mathcal U}$, 
based on the measurements $\tilde{\mbf d}$ and the knowledge of the anchor positions 
$\mbf p_{\mathcal K}$. Then
\begin{align}   \label{eq: cov tags CRLB simple}
\mathsf{cov}[\mbf{\hat{p}}_\mathcal{U}] \succeq \mbf F_\mathcal{U}^\dagger(\mbf p).
\end{align}
\end{proposition}

\begin{proof}
This result is a corollary of Proposition \ref{prop: constrained CRLB for network refined}
stated below, with $\mbf f_c \equiv \mbf 0$ in \eqref{eq:feasible set} and so 
$\mbf A_\mathcal{U} = \mbf I_{\sdim U}$.
\end{proof}

\subsection{Localizability Potentials and Optimal Design}
\label{section: localizability potentials unconstrained}

Given \eqref{eq: cov tags CRLB simple}, the following functions are possible candidates 
to define potential functions that penalize configurations of the ranging network leading 
to poor localizability
\begin{align}
J_A(\mbf p) &= \trace{\mbf F_\mathcal{U}^{-1}(\mbf p)} \;\; \text{(A-Optimal Design)},
\label{eq:pot:A} \\
J_D(\mbf p) &= - \ln \det \{\mbf F_\mathcal{U}(\mbf p)\} \;\; \text{(D-Optimal Design)}, \label{eq:pot:D} \\
J_E(\mbf p) &= - \lambda_{\min} \{\Fu (\mbf p)\} \;\; \text{(E-Optimal Design)},  \label{eq:pot:E}
\end{align}
assuming in the first two cases that $\mbf F_{\mathcal U}(\mbf p)$ is invertible.
In the following, we refer to the functions $J_A$, $J_D$ and $J_E$ as the
A-Opt, D-Opt and E-Opt potentials respectively, using standard terminology
from optimal experiment design \cite{pukelsheim_optimal_2006}. 

In each case, configurations $\mbf p$ for which $J(\mbf p)$ takes large values correspond 
to geometries for which the error covariance matrix of an unbiased position estimator will 
necessarily be ``large'' in a sense defined by the choice of potential. Hence,
for \eqref{eq:pot:A}, we have from \eqref{eq: cov tags CRLB simple} that
$J_A(\mbf p)$ is a lower bound on $\trace{\covar[\hat{\mbf p}_{\mathcal U}]}$, which
represents the total mean-squared error (MSE) of the unbiased estimator $\hat{\mbf p}_{\mathcal U}$. 
Similarly, \eqref{eq:pot:D} corresponds to a lower bound 
on $\ln \det (\covar[\hat{\mbf p}_{\mathcal U}])$, which would be equal (up to a constant) 
to the statistical entropy of $\hat{\mbf p}_{\mathcal U}$, 
if this estimate were to follow a normal distribution. Finally, still assuming 
$\mbf F_\mathcal{U} \succ 0$, minimizing $J_E$ in \eqref{eq:pot:E} aims to minimize 
the maximum eigenvalue of $\mbf F_\mathcal{U}^{-1}$ (equal to $1/\lambda_{\min}({\mbf F_\mathcal{U})}$), 
which is a lower bound on the maximum eigenvalue or
induced $2$-norm of $\covar[\hat{\mbf p}_{\mathcal U}]$.
Potentials like $J_E$ are often used to maintain the connectivity
\cite{Kim:TAC06:maximizingEigenvalue, decentralized_2010,siciliano_maintaining_2009}
or rigidity \cite{zelazo_decentralized_2015,sun_distributed_2015} of an MRS, which
are closely related problems.

Once a potential has been chosen, it can be used to move the
nodes to configurations of low potential values, where the localization accuracy is
expected to be high. This can be done for example by descending the gradient of
the potential, as discussed in Sections \ref{sec:gradients} and \ref{sec:extensions_rigidity}.

\begin{remark}
Another a priori possible potential is 
\[
J_T(\mbf p) = - \trace{\Fu(\mbf p)}. 
\]
Configurations $\mbf p$ that minimize this potential are called T-optimal designs \cite{pukelsheim_optimal_2006}.
However, in our case we can compute
\[
J_T(\mbf p) = - \alpha \sum_{\{i,j\}\in \mathcal{E}} d_{ij}^{2-2\kappa}, 
\]
with $\alpha$ a positive constant. 
In the case of additive Gaussian noise \eqref{eq:model_meas},
$\kappa = 1$ and $J_T$ is constant, so that it cannot be used to optimize $\mbf p$. 
In the case of multiplicative noise \eqref{eq:model_meas_lognormal}, we have $\kappa=2$ so 
$J_T(\mbf p) = -\alpha \sum_{\{i,j\} \in \mathcal{E}} d^{-2}_{ij}$ becomes a simple attractive
potential. In this case, $J_T$ cannot be used alone as a potential, since its global minimum 
is trivially achieved when all agents occupy the same position. 
In view of these remarks, $J_T$ is not considered further in the following.
\end{remark}

\section{Properties of the Fisher Information Matrix}
\label{sec:closed_form_crlb}

In this section, we study certain algebraic properties of the FIM
that are useful for the design of algorithms in the next sections.
In particular, we establish connections between the FIM and rigidity theory. 

\subsection{Infinitesimal Rigidity}
\label{ss:rigidity_theory}

For the ranging graph $\mathcal{G}=(\mathcal{E},\mathcal{V})$, the incidence matrix
$\mbf{H} \in \mathbb{Z}^{E \times N}$
is defined by first assigning an arbitrary direction $i \to j$ to each edge $\{i,j\}$ of $\mathcal{E}$,
and then setting each element as follows: 
\[
\text{for } \{i,j\} \in \mathcal E, k \in \mathcal V,
H_{i \to j,k} = 
\begin{cases}
1   &   \text{ if } k = i, \\
-1  &   \text{ if } k = j, \\
0   &   \text{ otherwise}.
\end{cases} 
\]
We use throughout the paper the lexicographic ordering to order the 
edges $i \to j$
and hence the rows of $\mbf H$. As a result, the rows of $\mbf H$ corresponding
to pairs of tags (in $\mathcal U \times \mathcal U$) appear first, followed by pairs in
$\mathcal U \times \mathcal K$ and finally by pairs of anchors, in $\mathcal K \times \mathcal K$.

\diffblock
\begin{remark}
Some references define $\mbf H$ as an $N \times E$ matrix, 
transposing the $E \times N$ matrix above. Our choice of convention is motivated by the
fact that it makes the connection to the rigidity matrix and the FIM clearer below.
\end{remark}
\diffend

Given a ranging graph $\mathcal{G}$, a \emph{framework} is a pair $(\mathcal{G},\mbf p)$, 
where the vector $\mbf p \in \r^{\sdim N}$ contains the positions of all agents.
The \textit{rigidity function} $\mbf r : \r^{\sdim N} \to \r^{E}$ of a framework 
$(\mathcal{G},\mbf p)$ is defined componentwise by
\begin{equation}
[\mbf r(\mathcal G,\mbf p)]_{i \to j}
= \frac{1}{2} \|\mbf p_{ij}\|^2, \;\; \forall \{ i,j \} \in \mathcal E,
\label{eq:rigidity_function}
\end{equation}
and its \emph{rigidity matrix} $\mbf R(\mathcal G, \mbf p) \in \r^{E \times \sdim N}$ is
the Jacobian $\partial \mbf r/\partial \mbf p$ of the rigidity function
\cite{tay_generating_1985,zelazo_decentralized_2015}, which can be written explicitly as 
\begin{equation}
\mbf{R}(\mathcal G, \mbf p) = \text{diag}(\dots ,\mbf p_{ij}^\top,\dots) \, [\mbf H \otimes \mbf I_\sdim].
\label{eq:rigidity_matrix}
\end{equation}
In other words, the row 
$i \to j$
of $\mbf R(\mathcal G, \mbf p)$ is
\[
\begin{bmatrix} \mbf 0 & \ldots & \mbf 0 & \mbf p_{ij}^T & \mbf 0 & \ldots & \mbf 0 
& -\mbf p_{ij}^T & \mbf 0 \ldots & \mbf 0 \end{bmatrix}
\]
with $\mbf p_{ij}^T$ occupying the $i^{th}$ block of $n$ coordinates and $-\mbf p_{ij}^T$ the $j^{th}$ block.
Next, when the node positions vary with time, consider motions that do not change the distances 
between nodes in ranging pairs, in other words, motions that keep the rigidity
function constant. These motions must then satisfy
\[
\frac{d \mbf r(\mathcal G,\mbf p)}{d t} = \mbf R(\mathcal G, \mbf p) \frac{d \mbf p}{d t} = \mbf 0,
\]
i.e., the corresponding velocity vectors $d \mbf p/dt$ must lie in the kernel of 
$\mbf R(\mathcal G, \mbf p)$. This constraint is rewritten more explicitly in the following
definition.

\begin{definition}[Infinitesimal motion of a framework]
\label{def: inf. motions}
An infinitesimal motion of a framework $(\mathcal G, \mbf p)$ is
any vector $\mbf v = \col(\mbf v_1,\ldots,\mbf v_N)$ in $\mathbb R^{\sdim N}$,
such that $\mbf v \in \ker \mbf R(\mathcal G, \mbf p)$.
Equivalently, for each edge $\{i,j\} \in \mathcal{E}$, we have
$
\mbf p_{ij}^T (\mbf v_i - \mbf v_j) = \mbf 0.
$ 
\end{definition}
Any framework admits a basic set of infinitesimal motions, namely, the \emph{Euclidean} 
infinitesimal motions of the framework \cite{tay_generating_1985,bonin_matroids_1996}, 
which can be defined for $n=3$ as
\[
\text{Eucl}^3_{\mbf p} = \left \{ \col(\mbf v + \boldsymbol{\omega}\times \mbf p_1,\ldots, 
\mbf v + \boldsymbol{\omega}\times \mbf{p}_N) \, | \, \mbf v, \boldsymbol{\omega} \in \r^3 \right\}, 
\]
and for $n=2$, with the notation $\mbf p_i = [x_i,y_i]^T$,
\begin{align*}
\text{Eucl}^2_{\mbf p} = \Big\{ \col\left( \mbf v + \omega \begin{bmatrix} y_1 \\ -x_1 \end{bmatrix},\ldots, 
\mbf v + \omega \begin{bmatrix} y_N \\ -x_N \end{bmatrix} \right) \Big | \\ 
\mbf v \in \r^2, \omega \in \mathbb R \Big \}. 
\end{align*}
These infinitesimal motions correspond to the global rigid translations and rotations 
of the whole framework, and it is immediate to verify that the subspace $\text{Eucl}_{\mbf p}$ 
is always contained in $\ker \mbf R(\mathcal G,\mbf p)$.
Infinitesimally rigid frameworks do not admit other infinitesimal motions,
which would correspond to internal deformations.

\begin{definition}[Infinitesimal rigidity]
\label{def: infinitesimal rigidity}
A framework $(\mathcal G,\mbf p)$ in $\mathbb R^{\sdim N}$ 
is called infinitesimally rigid if 
all its infinitesimal motions are Euclidean, i.e., if
$
\ker \mbf R(\mathcal G,\mbf p) = \text{Eucl}^\sdim_{\mbf p}.
$
\end{definition}

The following result provides a basis of $\text{Eucl}^{\sdim}_{\mbf p}$ and is used
in Section \ref{sec:extensions_rigidity}. When $n=3$, with $\mbf e_x$, $\mbf e_y$, $\mbf e_z$ 
the standard unit vectors in $\mathbb R^3$, define 
$\mbf v_{T_\xi} = \mbf 1_N \otimes \mbf e_\xi$ as well as 
$\mbf v_{R_\xi} = \col(\mbf e_\xi \times \mbf p_1,\ldots,\mbf e_\xi \times \mbf p_N)$, 
for $\xi \in \{x,y,z\}$. 
Similarly, if $n=2$ and $\mbf e_x$, $\mbf e_y$ are the standard unit vectors in $\mathbb R^2$, define $\mbf v_{T_x} = \mbf 1_N \otimes \mbf e_x$,
$\mbf v_{T_y} = \mbf 1_N \otimes \mbf e_y$ and
\[
\mbf v_{R_z} = \col \left( \begin{bmatrix} -y_1 \\ x_1 \end{bmatrix}, \ldots, 
\begin{bmatrix} -y_N \\ x_N \end{bmatrix} \right).
\]

\begin{proposition}
\label{prop:kerthree}
Suppose that $N \geq n$. If $n=2$ and at least $2$ nodes are at distinct locations, 
the dimension of $\text{Eucl}^{2}_{\mbf p}$ is $3$ and a basis of this subspace 
is given by $(\mbf v_{T_x},\mbf v_{T_y},\mbf v_{R_z})$.
If $n=3$ and we have at least $3$ nodes that are not aligned, the dimension of 
$\text{Eucl}^{3}_{\mbf p}$ is $6$ and a basis of this subspace is given by 
$(\mbf v_{T_x},\mbf v_{T_y},\mbf v_{T_z},\mbf v_{R_x},\mbf v_{R_y},\mbf v_{R_z})$.
\end{proposition}

\begin{proof}
We provide a proof for $n=3$, the case $n=2$ is similar.
The fact that the vectors in the proposition span $\text{Eucl}^{3}_{\mbf p}$ 
is clear by definition, so it is sufficient to prove their independence.
Consider a linear combination equal to zero
\begin{align*}
&\alpha_1 \mbf v_{T_x} + \alpha_2 \mbf v_{T_y} + \alpha_3 \mbf v_{T_z} 
+ \alpha_4 \mbf v_{R_x} + \alpha_5 \mbf v_{R_y} + \alpha_6 \mbf v_{R_z} \\
&= \col(\mbf v + \boldsymbol{\omega}\times \mbf p_1,\ldots,
\mbf v + \boldsymbol{\omega}\times \mbf p_n) = \mbf 0, 
\end{align*}
where $\mbf v = [\alpha_1,\alpha_2,\alpha_3]^T$ and
$\boldsymbol \omega = [\alpha_4,\alpha_5,\alpha_6]^T$. Suppose that the nodes indexed
by $i$, $j$ and $k$ are not aligned. We have from the equation above 
$\mbf v = - \boldsymbol \omega \times \mbf p_i$, and so
\[
\boldsymbol \omega \times (\mbf p_j - \mbf p_i) = \boldsymbol \omega \times (\mbf p_k - \mbf p_i) = \mbf 0.
\]
Since $(\mbf p_j - \mbf p_i)$ and $(\mbf p_k - \mbf p_i)$ are by assumption independent,
this gives $\boldsymbol \omega = \mbf 0$ and hence $\mbf v = \mbf 0$. This proves the independence
of the vectors in the proposition, which therefore form a basis of $\text{Eucl}^{3}_{\mbf p}$.
\end{proof}

\subsection{Relations between the Rigidity Matrix and the FIM}

Throughout this section, we consider the set of nodes (tags and anchors) to be at positions $\mbf p$,
with corresponding ranging graph $\mathcal G$. This defines a framework $(\mathcal G,\mbf p)$,
as discussed \diff{in} the previous section. The FIM $\mbf F$ is given by \eqref{eq:ourfim}, whereas the rigidity
matrix $\mbf R \coloneqq \mbf R(\mathcal G, \mbf p)$ is given by \eqref{eq:rigidity_matrix}.

\begin{proposition} 
\label{prop: FIM - R relation}
We have $\mbf F = \mbf R^\top \mbf Q \mbf R$, where 
$\mbf Q = \text{diag} \left( \ldots, 1 / (d_{ij}^{2\kappa} \sigma^2), \ldots \right) \in \r^{E \times E}$,
and $\kappa\in\{1,2\}$ is the parameter appearing in \eqref{eq:ourfim}.
\end{proposition}

To explain this result, remark that $\mbf F$ in \eqref{eq:ourfim} has a structure similar to the 
Laplacian matrix $\mbf L$ of the graph $\mathcal G$ \cite[Chapter 12]{Godsil:book01:algebraicGraphTheory}.
The expression of Proposition \ref{prop: FIM - R relation} then corresponds to the standard relationship 
$\mbf L = \mbf H^\top \mbf H$ between the incidence matrix and the usual Laplacian matrix of an undirected graph.
Hence, the FIM $\mbf F$ can be considered as \diff{a weighted Laplacian matrix, noting 
the relation \eqref{eq:rigidity_matrix} between $\mbf H$ and $\mbf R$.}
In \cite{zelazo_decentralized_2015}, matrices of the form $\mbf R^\top \mbf Q \mbf R$, 
for any diagonal matrix $\mbf Q$, are called (weighted) ``symmetric rigidity matrices''. 
Hence, with this terminology, Proposition \ref{prop: FIM - R relation} says that the FIM is 
a symmetric rigidity matrix, for a specific set of weights in $\mbf Q$ determined by the
properties of the measurement noise model. In particular, these weights depend inversely on
the (true) distances between ranging nodes.

\begin{proof} 
Starting from \eqref{eq:rigidity_matrix}, we have
\[
\mbf R^\top \mbf Q \mbf R = (\mbf H^\top \otimes \mbf I_n) \,
\text{diag} \left(\ldots,\frac{\mbf p_{ij} \mbf p_{ij}^\top}{d_{ij}^{2\kappa} \sigma^2},\ldots \right)
(\mbf H \otimes \mbf I_n).
\]
Hence, for $i \neq j$, the block $i,j$ of $\mbf R^\top \mbf Q \mbf R$ is 
\[
[\mbf R^\top \mbf Q \mbf R]_{ij} = \sum_{e \in \mathcal E} H_{ei} H_{ej} \mbf Q_{ee} 
= -\frac{\mbf p_{ij} \mbf p_{ij}^\top}{d_{ij}^{2\kappa} \sigma^2} \mathsf{1}_{\mathcal N_i}(j) = \mbf F_{ij},
\]
using the fact that $H_{ei} H_{ej} = -1$ if 
$e$ is $i \to j$ and $0$ otherwise. Similarly, for all $i$
\[
[\mbf R^\top \mbf Q \mbf R]_{ii} = \sum_{e \in \mathcal E} H_{ei} H_{ei} \mbf Q_{ee} 
= \sum_{j \in \mathcal N_i}
\frac{\mbf p_{ij} \mbf p_{ij}^\top}{d_{ij}^{2\kappa} \sigma^2} = \mbf F_{ii}.
\]
\end{proof}

The following result then follows immediately from the fact that $\mbf Q \succ \mbf 0$
in Proposition \ref{prop: FIM - R relation}.
\begin{corollary} \label{prop:samekernel}
We have $\ker \mbf F = \ker \mbf R$.
\end{corollary}

\diff{
The following result states that infinitesimal rigidity provides a sufficient condition for 
the invertibility 
of the symmetric positive semi-definite matrix $\mbf F_\mathcal U$ appearing in \eqref{eq: FIM partitioning}.
}

\begin{theorem} \label{thm: Fu invertible}
\diff{Suppose that} the framework $(\mathcal G,\mbf p)$ is infinitesimally rigid and contains at least $n$ anchors
at distinct locations. \diff{Moreover, when $n=3$, suppose that at least $3$ of these anchors are not
aligned}. Then $\mbf F_{\mathcal U}$ \diff{is invertible}.
\end{theorem}

\begin{proof}
We give the proof in the more involved case $n=3$.
With the assumed ordering of nodes and edges, the rigidity matrix has the following
block structure
\[
\mbf R = \begin{bmatrix}
\mbf R_1 & \mbf R_2 \\
\mbf 0 & \mbf R_{3}
\end{bmatrix}, \text{ with } \mbf R_1 \in \mathbb R^{P \times U}, 
\mbf R_3 \in \mathbb R^{\frac{K(K-1)}{2} \times K}.
\] 
In other words, the rows of the matrix $\mbf R_1$ correspond to the edges
internal to $\mathcal U$ and between $\mathcal U$ and $\mathcal K$, whereas
$\mathbf R_3$ is the rigidity matrix of the complete subgraph formed by the 
anchors and the links between them. Now, we have 
$\mbf F_\mathcal U = \mbf R_1^\top \mbf Q_1 \mbf R_1$, with $\mbf Q_1$ diagonal and
invertible, as in Proposition \ref{prop: FIM - R relation}, so 
$\ker \mbf F_{\mathcal U} = \ker \mbf R_1$.
Consider some vector $\mbf x_1 \in \mathbb{R}^U$ with $\mbf x_1 \in \ker \mbf R_1$. 
Then, 
\begin{equation}
\mbf R \begin{bmatrix} \mbf x_1 \\ \mbf 0 \end{bmatrix} = \begin{bmatrix}
\mbf R_1 & \mbf R_2 \\
\mbf 0 & \mbf R_3
\end{bmatrix}
\begin{bmatrix} \mbf x_1 \\ \mbf 0 \end{bmatrix}
= \mbf 0,
\end{equation}
hence $\text{col}(\mbf x_1,\mbf 0)$ is in $\ker \mbf R$. Since $\mathcal G$ is infinitesimally 
rigid, there must exist $\mbf v$, $\boldsymbol \omega$ in $\mathbb R^3$ such that 
\[
\begin{bmatrix} \mbf x_1 \\ \mbf 0 \end{bmatrix} = 
\text{col}(\mbf v+\boldsymbol \omega\times \mbf p_1,\ldots,\mbf v + \boldsymbol \omega \times \mbf p_N).
\] 
In particular, for the 3 anchors that are not aligned, indexed by $i$, $j$ and $k$, we must have
\[
\mbf v+ \boldsymbol \omega \times \mbf p_i = \mbf v + \boldsymbol \omega \times \mbf p_j 
= \mbf v + \boldsymbol \omega \times \mbf p_k = \mbf 0.
\]
From this, we conclude as in the proof of Proposition \ref{prop:kerthree} that
$\mbf v = \boldsymbol \omega = \mbf 0$, which in turns implies $\mbf x_1 = \mbf 0$.
Hence $\ker \mbf F_{\mathcal U} = \{\mbf 0\}$, i.e., $\mbf F_{\mathcal U} \succ \mbf 0$.  
\end{proof}

\begin{remark}
If we have only one tag, then one can show that $\mbf F_{\mathcal U}$ is invertible 
if and only if we have at least $n$ anchors and the nodes' locations span an affine space 
of full dimension $n$ (i.e., we have $3$ non aligned nodes if $n=2$, and $4$ non coplanar
nodes if $n=3$). Note that if we have only $n$ anchors, we cannot localize uniquely the tag
in general, even with perfect measurements, because the intersection of $n$ spheres 
in $\mathbb R^n$ gives two possible locations. Hence, even when $\mbf F_{\mathcal U}$ is invertible,
the localization problem might not be uniquely solvable. Unicity of the localization solution
can be characterized by the stronger notion of global rigidity \cite{Aspnes:TMC06:theoryLoc},
which however is more complex to check if $n=2$ and for which no exact test is currently known 
if $n=3$.
\end{remark}

Theorem \ref{thm: Fu invertible} can be used to produce an initial node placement
and choose ranging links to guarantee that $\mbf F_{\mathcal U}$ is already 
invertible at the start of the deployment. For this, we should ensure that $(\mathcal G, \mbf p)$
is infinitesimally rigid. One convenient way to satisfy this condition (in fact, the 
stronger condition of global rigidity) is to construct a \emph{triangulation graph} 
\cite{Aspnes:TMC06:theoryLoc,Moore:Sensys04:networkLoc}: starting from a set of at least
$n+1$ anchors, we add tags one by one, with each new tag connected to at least $n+1$ previous 
nodes that are in general position ($3$ non-aligned nodes if $n=2$, $4$ non-coplanar nodes if $n=3$).
Although this construction requires more anchors and links than the strict minimum necessary for
the invertibility of $\mbf F_{\mathcal U}$, the resulting network supports efficient distributed
localization algorithms that are robust to measurement noise \cite{Moore:Sensys04:networkLoc}.

\section{Distributed Gradient Computations for the Localizability Potentials}
\label{sec:gradients}

In order to implement the gradient descent scheme \eqref{eq:descent_per_agent}, 
in Section \ref{section: gradient FIM} we provide analytical forms for
the gradients of the localizability potentials \eqref{eq:pot:A}, \eqref{eq:pot:D} and \eqref{eq:pot:E}. 
Then, in Sections \ref{section: decentralized D and A} and \ref{section: decentralized E}, 
we describe decentralized deployment algorithms 
by showing how each agent can compute its components of the gradient of the chosen localizability 
potential, using its own local information as well as data obtained from its neighbors in the ranging graph.

\subsection{Partial Derivatives of the FIM}
\label{section: gradient FIM}

Irrespective to the potential considered, we need to evaluate the derivative 
of the FIM $\Fu$ in \eqref{eq: FIM partitioning}
with respect to any coordinate $\xi_i \in \{x_i,y_i,z_i\}$ of a mobile agent $i$ 
(anchor or tag) located at $\mbf p_i = [x_i,y_i,z_i]^\top$. 
We provide formulas for the case $n=3$, the case $n=2$ being similar.
Define the notation $\xi_{ij}=\xi_i - \xi_j$ and 
$\gamma_{ij} = \frac{\kappa}{\sigma^2d_{ij}^{2(\kappa+1)}} \mathsf{1}_{\mathcal N_i}(j)$. 
For $\mbf F_{ij}$, $i\neq j$, the $3 \times 3$ blocks introduced in \eqref{eq:ourfim}, 
we find
\begin{equation*} 
\der{\mbf F_{ij}}{x_i}
= 
\gamma_{ij}
\begin{bmatrix}
 x_{ij}^{3} - \frac{d^{2}_{ij} x_{ij}}{\kappa} 
& x_{ij}^{2} y_{ij} - \frac{d^{2}_{ij} y_{ij}}{2\kappa} 
&  x_{ij}^{2} z_{ij} -  \frac{d^{2}_{ij} z_{ij}}{2\kappa}  \\
\star & 
x_{ij} y_{ij}^{2} &
x_{ij} y_{ij} z_{ij}\\
\star& 
\star & 
x_{ij} z_{ij}^{2}
\end{bmatrix}
\end{equation*}
\begin{equation*}
\der{\mbf F_{ij}}{y_i}
= \gamma_{ij}
\begin{bmatrix}
	x_{ij}^{2} y_{ij} & 
	x_{ij} y_{ij}^{2} - \frac{d^{2}_{ij} x_{ij}}{2\kappa} & 
	x_{ij} y_{ij} z_{ij}\\
	\star &
	 y_{ij}^{3} - \frac{d^{2}_{ij} y_{ij}}{\kappa} &
	 y_{ij}^{2} z_{ij} - \frac{d^{2}_{ij} z_{ij}}{2\kappa}\\
	 \star & 
	 \star & 
	 y_{ij} z_{ij}^{2}
\end{bmatrix},
\end{equation*}
\begin{equation}    \label{eq:derfim}
\der{\mbf F_{ij}}{z_i}
= \gamma_{ij}
\begin{bmatrix}
x_{ij}^{2} z_{ij} & 
x_{ij} y_{ij} z_{ij} &
x_{ij} z_{ij}^{2}  - \frac{d^{2}_{ij} x_{ij}}{2\kappa}\\
\star &
y_{ij}^{2} z_{ij} & 
y_{ij} z_{ij}^{2} - \frac{d^{2}_{ij} y_{ij}}{2\kappa} \\
\star &
\star & 
z_{ij}^{3} - \frac{d^{2}_{ij} z_{ij}}{\kappa}
\end{bmatrix},
\end{equation}
where the symbol $\star$ replaces symmetric terms. These expressions are sufficient to compute 
the whole matrix $\partial \mbf F_{\mathcal U} / \partial \xi_i$,
because $\mbf F_{ji} = \mbf F_{ij}$, $\mbf{F}_{aa} = -\sum_{b \in \mathcal N_a} \mbf F_{ab}$,
and $\partial \mbf F_{ab}/\partial \xi_i = \mbf 0$ if $a \neq b$ and $i \notin \{a,b\}$.

Using standard differentiation rules \cite{petersen_matrix_2012}, the partial derivatives 
of the A-Opt potential \eqref{eq:pot:A} are
\begin{equation}
    \der{J_A(\mbf p)}{\xi_i} =\der{\trace{\Fu^{-1}}}{\xi_i} = -\trace{\Fu^{-2}\der{\mbf F_{\mathcal U}}{\xi_i}}.
    \label{eq:der:Aopt}
\end{equation}
Similarly, we can compute the derivatives of the D-Opt potential \eqref{eq:pot:D} as
\begin{equation}
    \der{J_D(\mbf p)}{\xi_i} = - \der{\ln \det \Fu}{\xi_i}  = -\trace{\Fu^{-1}\der{\mbf F_{\mathcal U}}{\xi_i}}.
    \label{eq:der:Dopt}
\end{equation}
Finally, if 
$\lambda_{\min}(\Fu)$ is a non-repeated eigenvalue with associated unit norm eigenvector $\mbf v$, 
we can compute the derivative of the E-Opt potential \eqref{eq:pot:E} as
\cite[p. 565]{harville_matrix_1997} 
\begin{align}   \label{eq:der:Eopt - 1}
\der{J_E(\mbf p)}{\xi_i} = -\der{\lambda_{\min}(\mbf p)}{\xi_i} = -\mbf v^\top \der{\Fu}{\xi_i} \mbf v.
\end{align}
Hence, we can in principle compute the gradient of the chosen localizability 
potential, using the expressions for the FIM and its derivatives.
However, in practice we would also like to be able to implement these computations in a
distributed manner, in order to obtain deployment strategies that 
can \diff{be used by an MRS with incomplete ranging graph $\mathcal{G}$,
assuming communication over this ranging graph is also possible.}

\subsection{Decentralized Gradient Computations for the D- and A-Opt Potentials}
\label{section: decentralized D and A}
We propose now a new method to estimate in a distributed way the gradient of the D- 
and A-Opt potentials at a given configuration $\mbf p$, which have similar expressions, 
see \eqref{eq:der:Aopt} and \eqref{eq:der:Dopt}. 
As mentioned in Remark \ref{rmk: estimate in the loop}, 
\diff{we assume that each node $i$ has access to its position $\mbf p_i$, 
which could be its true position (e.g., for anchors) or an estimate obtained
after executing a localization algorithm such as the one in \cite{Moore:Sensys04:networkLoc}. 
In the latter case, the algorithms presented here will simply produce the gradient
of $J_{\text{loc}}$ at the estimated position.
}
 \diff{In the following}, we omit $\mbf p$ from the notation, writing $\Fu$ instead of $\Fu(\mbf p)$.
The method essentially relies on 
inverting $\Fu$ in a decentralized manner, which we discuss first.

\subsubsection{Auxiliary Problem}
Suppose that each tag $i \in \mathcal U$ knows initially a matrix $\mbf E_i \in \r^{n \times m}$,
for some integer $m$, and the tags need to compute $\Fu^{-1} \mbf E$ in a distributed manner
over the network $\mathcal G$, where $\mbf E = \col(\mbf E_1,\ldots,\mbf E_U) \in \r^{nU \times m}$.
This is equivalent to solving in a decentralized manner the linear 
system $\Fu \mbf X = \mbf E$, with the \diff{matrix} variable $\mbf X \in \r^{nU \times m}$.
A special case of this problem is to compute $\Fu^{-1}$, when $\mbf E = \mbf I_{nU}$.

Consider the following system of differential equations
\begin{equation}
\dot{\mbf X}(t) = -\mbf F_\mathcal{U} \mbf X(t) + \mbf E, \; \mbf X(0) = \mbf X_0.
\label{eq:diffsyst}
\end{equation}
If $\mbf F_\mathcal{U} \succ \mbf 0$, as guaranteed 
by 
Theorem \ref{thm: Fu invertible}, then $-\mbf F_\mathcal{U}$ has strictly negative 
eigenvalues, i.e., is stable, so the solution $\mbf X(t)$ to the system \eqref{eq:diffsyst} 
converges to the solution $\mbf F^{-1}_\mathcal{U} \mbf E$ of the linear system as $t \to \infty$, 
no matter the choice of initial condition $\mbf X_0$.
A discrete-time version of the flow \eqref{eq:diffsyst} can be implemented for $l \geq 0$ as
\[
\mbf X_{l+1} = \mbf X_{l} - \eta_l \, (\mbf F_\mathcal{U} \mbf X_l - \mbf E),
\]
for some stepsizes $\eta_l$, which reads more explicitly for each tag $1 \leq i \leq U$
\begin{align}
\mbf X_{i,l+1} = & \, \eta_l \sum_{j \in \mathcal N_i \cap \mathcal U} \mbf F_{ij} 
(\mbf X_{i,l} - \mbf X_{j,l}) \nonumber \\  
&+ \left( \mbf I_n + \eta_l \sum_{j \in \mathcal N_i \cap \mathcal K} \mbf F_{ij}\right) 
\mbf X_{i,l} + \diff{\eta_l}\mbf E_i. 
\label{eq:diffsyst_i}
\end{align}
Again, the iterates $\mbf X_k$ converge to the desired solution $\Fu^{-1} \mbf E$ 
if we choose for example $\eta_l = \eta$ constant and sufficiently small 
(namely, as long as $\eta < 2/\lambda_{\max}(\Fu)$). 
The iterations \eqref{eq:diffsyst_i} can be implemented in a decentralized manner
by the tags, i.e., at each step $l$ tag $i$ only \diff{needs} to exchange its
matrix $\mbf X_i$ with its neighboring tags. This also requires that tag $i$ knows 
$\mbf F_{ij}$ for $j \in \mathcal N_i$, which is the case if prior to the iterations, 
the nodes (tags and anchors) broadcast their position (estimates) to their neighbors.
When the iterations have converged, the $n \times m$ matrix $\mbf X_i$ at tag $i$
represents the $i^{\text{th}}$ block of rows of $\Fu^{-1} \mbf E$, i.e., 
$\Fu^{-1} \mbf E = \col(\mbf X_1,\ldots,\mbf X_U)$.

\begin{remark}
The iterations \eqref{eq:diffsyst_i} correspond to Richardson iterations
to solve the linear system $\Fu \mbf X = \mbf E$ in a decentralized way \cite{Bertsekas:2015:parallel}. 
Other distributed iterative methods could be used, such as the Jacobi over-relaxation iterations
\begin{align*}
\mbf X_{i,l+1} = \, &(1-\eta) \, \mbf X_{i,l} + 
\eta \, \mbf F_{ii}^{-1} \, \left( \mbf E_i - \sum_{j \in \mathcal N_i \cap \mathcal U} \mbf F_{ij} \mbf X_{j,l} \right),
\end{align*}
with potentially better convergence properties,
but a detailed discussion of such alternatives, which can be found in \cite[Chapter 2]{Bertsekas:2015:parallel}, 
is outside of the scope of this paper. 
\end{remark}

\subsubsection{Application to compute $\partial J_D / \partial \xi_i$} 

To implement the gradient descent scheme \eqref{eq:descent_per_agent} for D-optimization,
each mobile node $i$ (tag or anchor) needs to compute 
$\partial J_D / \partial \xi_i$ for $\xi_i \in \{x_i,y_i,z_i\}$, which is given by \eqref{eq:der:Dopt}.
Denote $\mbf M = \Fu^{-1} \in \r^{nU \times nU}$ and its $n \times n$ blocks $\mbf M_{ij}$,
for $1 \leq i,j \leq U$. 
First, the tags run the iterations \eqref{eq:diffsyst_i}, with the matrix $\mbf E = \mbf I_{nU}$.
That is, tag $j$ uses the matrix $\mbf E_j = \mbf e^\top_j \otimes \mbf I_n$, where $\mbf e_j$ is 
the $j^{\text{th}}$ unit vector in $\r^U$. After convergence, tag $j$ stores an approximation 
of the matrix $\mbf M_j = [\mbf M_{j1},\ldots,\mbf M_{jU}] \in \mathbb R^{n \times nU}$. 
\diff{A stopping condition $\max_{i\in \mathcal{N}_j}\|\mbf X_{i,l+1}-\mbf X_{i,l}\|/\|\mbf X_{i,l}\|<\epsilon$ 
can be implemented at each node $j$, for a threshold $\epsilon>0$.}

Next, note from \eqref{eq:derfim} that the only $n \times n$ non-zero blocks 
$\partial \mbf F_{ab}/\partial \xi_i$, with $0 \leq a,b \leq U$, are those for 
which: i) $a=b$ and $a \in \mathcal N_i$; ii) $a = b = i$; 
iii) $a = i$ and $b \in \mathcal N_i$; or iv) $b = i$ and $a \in \mathcal N_i$.
Moreover, if $i$ is a mobile anchor (so $i \geq U+1$), only case i) can occur.
From this remark, we can derive the following expressions. If $i \in \mathcal U$
\begin{align}    
\frac{\partial J_D(\mbf p)}{\partial \xi_i} = &\sum_{j \in \mathcal N_i \cap \mathcal U}
\trace{ \left( \mbf M_{jj} + \mbf M_{ii} - 2 \mbf M_{ij} \right)
\frac{\partial \mbf F_{ij}}{\partial \xi_i}} \nonumber \\
&\diff{+
\sum_{j \in \mathcal{N}_i \cap \mathcal{K}} \trace{\mbf M_{ii} \der{\mbf F_{ij}}{\xi_i}}
}
,
\label{eq:der:Dopt tag}
\end{align}
and if $i \in \mathcal K$
\begin{equation}    \label{eq:der:Dopt anchor}
\frac{\partial J_D(\mbf p)}{\partial \xi_i} = \sum_{j \in \mathcal N_i \cap \mathcal U}
\trace{ \mbf M_{jj} \frac{\partial \mbf F_{ij}}{\partial \xi_i}}.
\end{equation}
Assuming that each node knows an estimate of its coordinates and of its neighbors'
coordinates, node $i$ can obtain from its neighbor tags $j$ the terms 
$\trace{\mbf M_{jj} \partial \mbf F_{ij}/\partial \xi_i}$, and also compute the terms 
\diff{
$\trace{\mbf M_{ii} \partial \mbf F_{ij}/\partial \xi_i}$ 
and $\trace{\mbf M_{ij} \partial \mbf F_{ij}/\partial \xi_i}$
}
if $i \in \mathcal U$.
Hence, overall this provides a method allowing each mobile node $i$ to compute 
$\partial J_D / \partial \xi_i$ by communicating only with its neighbors. 
\diff{Nevertheless, 
it requires significant data exchanges between the agents 
(exchanges to reach the convergence in \eqref{eq:diffsyst_i} and sending of the approximations 
of $\mbf M_j$, of size $n \times nU$, to the neighbors), which can  limit its scalability to large MRS.}
Algorithm \ref{algo:dopt} summarizes the distributed gradient computation procedure 
for D-optimization.

\begin{algorithm}[htbp!]
	\KwData{Each node $i$ knows an estimate of its $\mbf p_i$ from a localization 
	algorithm, or exactly if $i \in \mathcal K$}
	\KwResult{Each mobile node $i$ knows $\partial J_D(\mbf p)/\partial \mbf p_i$}
	Each node $i \in \mathcal U \cup \mathcal K$ broadcasts $\mbf p_i$ to its neighbors\;
	The tags run the iterations \eqref{eq:diffsyst_i} until convergence, 
	with $\mbf E_j = \mbf e^\top_j \otimes \mbf I_n$ for tag $j$,
	and each tag $j$ stores the resulting matrix $\mbf M_j$\;
    Each mobile tag $i$ computes $\sum_{j \in \mathcal N_i}
    \trace{ \left( \mbf M_{ii} - 2 \mbf M_{ij} \diff{\mathsf 1_{\mathcal K}(j)} \right) \frac{\partial \mbf F_{ij}}{\partial \xi_i}}$\;
	Each tag $j$ computes and sends $\trace{ \mbf M_{jj} \frac{\partial \mbf F_{ij}}{\partial \xi_i}}$
	to each of its mobile neighbors $i \in \mathcal N_j$ ($i$ tag or anchor)\;
	Each mobile node $i$ computes its gradient using \eqref{eq:der:Dopt tag} or \eqref{eq:der:Dopt anchor}\;
	\caption{D-Opt distributed gradient computation}
	\label{algo:dopt}
\end{algorithm}

The same steps can be used to compute the gradient \eqref{eq:der:Aopt} 
at each mobile node for A-optimization.
The only difference is that the matrices $\mbf M_i$ above should represent rows 
of $\Fu^{-2}$ instead of $\Fu^{-1}$. For this, the tags first compute the rows $\mbf{\tilde M}_i$ 
of $\Fu^{-1}$ using the iterations \eqref{eq:diffsyst_i}. Then, we restart these iterations 
but now replacing the matrices $\mbf E_i = \mbf e^\top_i \otimes \mbf I_{n}$ by $\mbf{\tilde M}_i$. 
This computes an approximation of $\Fu^{-1} \Fu^{-1} = \Fu^{-2}$, as desired. 
\diff{However, the resulting distributed A-Opt scheme requires more computational resources 
and communication exchanges and is thus less applicable for large MRS.}

\subsection{Decentralized Computation of E-Opt Gradient}
\label{section: decentralized E}

The decentralized computation of the gradient of the E-Opt potential can be done
using the methodology developed in \cite{decentralized_2010} for the standard Laplacian,
also used in \cite{zelazo_decentralized_2015} for the symmetric rigidity matrix.
Hence, our presentation is brief and focuses on adapting this methodology to $\Fu(\mbf p)$.

Using the sparsity of $\Fu$, if $i \in \mathcal{U}$, we can rewrite \eqref{eq:der:Eopt - 1} as
\begin{align}
\der{J_E(\mbf p)}{\xi_i} 
= &\sum_{j \in \mathcal{N}_i \cap \mathcal{U}} 
(\mbf v_i - \mbf v_j)^T \der{\mbf F_{ij}}{\xi_i} (\mbf v_i - \mbf v_j)^T \nonumber \\
& + \mbf v_i^T \left( \sum_{j \in \mathcal{N}_i \cap \mathcal{K}} \der{\mbf F_{ij}}{\xi_i} \right)
\mbf v_i,
\label{eq:gradientJe:sparsity}
\end{align}
and if $i \in \mathcal{K}$
\begin{equation}
\der{J_E(\mbf p)}{\xi_i} = \sum_{j \in \mathcal{N}_i \cap \mathcal{U}}
\mbf v_j^\top \der{\mbf F_{ij}}{\xi_i}  \mbf v_j,
\label{eq:gradientJe:sparsity2}
\end{equation}
where $\mbf v = \col(\mbf v_1,\dots, \mbf v_U) \in \r^{\sdim U}$.
Computing
these expressions requires a decentralized
algorithm to estimate the components of $\mbf v$, a unit norm eigenvector 
associated with
$\lambda_1 := \lambda_{\min}(\mbf F_\mathcal{U})$.

\subsubsection{Power-iteration eigenvector estimator}
To compute $\mbf v$ in a decentralized manner, consider the solution $t \mapsto \mbf w(t) \in \mathbb R^{nU}$ 
to the following differential equation, adapted from \cite{decentralized_2010},
\begin{equation} 
\dot{\mbf w} = -[\beta \Fu + \mu((\sdim U)^{-1}\|\mbf w(t)\|^2 - 1)\mbf I_{\sdim U}]\mbf w(t),
\label{eq:eopt:scheme}
\end{equation}
with an initial condition $\mbf w_0 := \mbf w (0)$ 
and $\beta, \mu> 0$.

\begin{proposition}
If $\mu > \lambda_1 \beta$ and $\mbf w_0 ^\top \mbf v \neq  0$, then the solution $\mbf w(t)$ to \eqref{eq:eopt:scheme} 
converges to an eigenvector $\mbf w_\infty$ of $\Fu$, associated with $\lambda_1$ and proportional to $\mbf v$.  
\label{prop:eopt_conv}
\end{proposition}
\begin{proof}
This follows from the argument in the appendix of \cite{decentralized_2010}.
\end{proof}
In practice, we can choose $\mbf w_0$ randomly to fulfill the condition 
$\mbf w_0 ^\top \mbf v \neq  0$ with probability one. To set the gains $\beta, \mu$, 
note that $\trace{\Fu} > \lambda_1$ since $\Fu \succ 0$.
Then, for the additive measurement noise model \eqref{eq:model_meas}, we have 
$\trace{\Fu} \leq \frac{2 P}{\sigma^2}$. So, if we choose $\beta \diff{~\geq~} \sigma^2/(2 P)$
and $\mu > 1$, the condition of Proposition \ref{prop:eopt_conv} is satisfied.
For the log-normal model \eqref{eq:model_meas_lognormal}, we have 
$\trace{\Fu} \leq \frac{2}{\sigma^2}\sum_{\{i,j\} \in \mathcal{E}, i \in \mathcal{U}} d_{ij}^{-2}$. 
Hence, if we set again $\beta \diff{~\geq~} \sigma^2/(2 P)$ and now $\mu > 1/d_{min}^2$,
such that $d_{ij} \geq d_{\min}$ for all $i, j$, then the condition of Proposition
\ref{prop:eopt_conv} is satisfied. The minimum distance $d_{\min}$ between robots
could be enforced as part of a collision avoidance scheme.

An estimation algorithm for $\mbf v$ is obtained by discretizing 
\eqref{eq:eopt:scheme}, leading to the following iterations
for each agent $i \in \mathcal{U}$ 
\begin{align}
\mbf w_{i,l+1} = &\mbf w_{i,l}
- \eta_l \Big(
\mu (\diff{s_{l}-1}) \mbf w_{i,l} \nonumber \\ &+  
\beta \sum_{a\in (\mathcal N_i \cup \{i\})\cap \mathcal{U}} \mbf F_{il} \mbf w_{a,l} \Big),
\label{eq:eopt:discr}
\end{align}
where $\eta_l > 0$ is a sufficiently small step-size and $s_{l} := \| \mbf w_l \|^2 / nU$. 
All the terms in \eqref{eq:eopt:discr} can be obtained locally by node $i$ using
one-hop communication with its neighbors, except for the global average $s_l$,
which can be computed by a consensus algorithm as described next. 
The last step is to normalize $\mbf w_\infty$, obtained after convergence 
in \eqref{eq:eopt:discr}. This can again be done by each individual agent, 
since $\mbf v := \mbf w_\infty/\sqrt{nU s_\infty}$
is a unit-norm vector.

\subsubsection{Estimation of $s_{l}$ via a consensus algorithm}
Since $s_{l} = \|\mbf w_l\|^2/(\sdim U) = \frac{1}{U} \sum_{i=1}^{U} (\|\mbf w_{i,l}\|^2/\sdim)$,
this term can be computed by the tags 
using a decentralized averaging consensus algorithm.
We assume for simplicity that the graph of the tags $\mathcal{G}_\mathcal{U}$ is connected. 
To solve the averaging problem, each tag $i$ initializes a variable
$\hat{s}_{i,l,0}:=\|\mbf w_{i,l}\|^2/\sdim$. Then, they execute in a distributed
manner the iterations
\begin{equation}
 \hat{\mbf s}_{l,m+1} = \mbf G \, \hat{ \mbf s}_{l,m}, \forall m \geq 0, 
\label{eq:eoptdistri:consensusFilter}
\end{equation} 
where $\hat{\mbf s}_{l,m} = \col(\hat{s}_{1,l,m},\ldots,\hat{s}_{U,l,m})$,
and $\mbf G$ is a doubly stochastic matrix of weights $G_{ij}$ associated
with the edges of $\mathcal{G}_\mathcal{U}$ 
(i.e., $\sum_{u=1}^U G_{iu} = \sum_{u=1}^U G_{ui}=1$, for $1 \leq i \leq U$,
and $G_{ij} = 0$ if $j \notin \mathcal N_i$),
for instance 
the Metropolis-Hastings weights
\[
\begin{cases}
G_{ij} = \mathsf{1}_{\mathcal N_i \cap \mathcal U}(j)(1+\max(|\mathcal N_i|,|\mathcal N_j|))^{-1}, 
\forall i \neq j, \\
G_{ii} = 1 -  \sum_{u=1}^U G_{iu}.
\end{cases}
\]
We then have $\hat{\mbf s}_{l,m} \to s_l \mbf 1_U$ \cite[p. 58]{Bullo:book09:distributedRobotics}, so
that each tag knows after convergence the scalar value $s_l$ needed for \eqref{eq:eopt:discr}.
\diff{
\begin{remark}
Since $s_l$ is time varying and we need to track its value at each period $l$, 
dynamic consensus methods \cite{Kia:CSM19:dynamicConsensus} may converge faster than 
the solution presented here. We leave the exploration of such schemes for future work.
\end{remark}
}

Algorithm \ref{algo:distriEoptVect} summarizes the decentralized computation of the 
estimate $\hat{\mbf v}_i$ of the $i$-th component of $\mbf v$ by a given tag $i \in \mathcal{U}$. 
After decentralized estimation of $\mbf v$ by the tags, each mobile agent $i$ 
can compute its components of the gradient of $J_E$ from \eqref{eq:gradientJe:sparsity}
or \eqref{eq:gradientJe:sparsity2} by communicating with its neighbors.

\begin{algorithm}
	\KwData{
	$\mbf w_{i,0}$ random,
	$\mbf G$, $\mu,\beta,n_\text{iter},\tilde{n}_\text{iter}$}
	\For{$0 \leq l \leq n_\text{iter}$}
	{
    $\hat{s}_{i,l,0} = \|\mbf w_{i,l}\|^2/n$; \\
    \For{$0 \leq m \leq \tilde{n}_\text{iter}$}
    {
        $\hat{s}_{i,l,m+1} = G_{ii} \hat{s}_{i,l,m} 
        + \sum_{j \in \mathcal N_i \cap \mathcal U} G_{ij} \hat{s}_{j,l,m}$; \\
    }
    \textbf{compute} $\mbf w_{i,l+1}$, setting $s_l := \hat{s}_{i,\tilde{n}_\text{iter}}$ 
    in \eqref{eq:eopt:discr}. 
    }
     \textbf{transmit} 
    $\hat{\mbf v}_i := \frac{\mbf w_{i,n_\text{iter}}}{ \sqrt{ \sdim U 
    \hat s_{\tilde{n}_{i,\text{iter}}}}}$ 
    to the neighborhood;
\caption{Estimation of $\mbf v_i$ by tag $i \in \mathcal U$.}
\label{algo:distriEoptVect}
\end{algorithm}

\begin{remark}
\diff{When the subgraph of $\mathcal G$ with only the tags 
is not connected,
it is still possible to distributively compute the gradient of $J_E$.}  
In this case, there exists a $U\times U$ permutation matrix $\mbf P$
such that $\check{\mbf F}_\mathcal{U}=(\mbf P \otimes \mbf I_n)^{-1} \Fu (\mbf P \otimes \mbf I_n) = \mathrm{diag} 
(\mbf F_{\mathcal{S}_1} \dots \mbf F_{\mathcal{S}_l} \dots)$ is block diagonal, 
where each $\mathcal{S}_l$ represents a subset of connected tags. 
Hence, the minimal eigenvalue $\lambda$ of $\Fu$ is among the minimal eigenvalues $\lambda_{\mathcal{S}_l}$ of the blocks $\mbf F_{\mathcal{S}_l}$.
Therefore, each subset $\mathcal{S}_l$ can use Algorithm \ref{algo:distriEoptVect} 
to compute its eigenvector $\mbf{v}_{\mathcal{S}_l}$ associated to $\lambda_{\mathcal{S}_l} :=\mbf{v}_{\mathcal{S}_l}^\top \mbf F_{\mathcal{S}_l}\mbf{v}_{\mathcal{S}_l}$. 
On the other hand, the graph $\mathcal{G}$ with all nodes is assumed rigid 
and hence fully connected. This allows comparing the $\lambda_{S_l}$ through 
the network $\mathcal{K}$ formed by the anchors in order to find 
$\lambda := \min_{\mathcal{S}_l} \lambda_{\mathcal{S}_{l}}$ corresponding 
to the subset $\mathcal{S}^*$. 
Since $\check{\mbf F}_\mathcal{U}$ is block diagonal, 
its eigenvector associated with $\lambda$ is $\col(0, \dots, \mbf v_{\mathcal{S}^*}, \dots 0)$, 
which then yields 
$\mbf v = (\mbf P \otimes \mbf I_n) \, \col(0, \dots, \mbf v_{\mathcal{S}^*}, \dots 0)$ 
for $\Fu$. \diff{Then, $\mbf v$ gives the gradient of $J_E$ using \eqref{eq:gradientJe:sparsity} and \eqref{eq:gradientJe:sparsity2}}.
\end{remark}

\section{Localizability Optimization for Rigid Bodies}
\label{sec:extensions_rigidity}

\subsection{Constrained Localizability Optimization} 
\label{section: constrainted loc}

\begin{figure}
    \centering
    	\begin{tikzpicture}
    \coordinate (A1) at (-1,0);
    \coordinate (A2) at (-1,1);
    \coordinate (A3) at (2.5,-0.2);
    
    \coordinate (T4) at (0.8,1.3);
    \coordinate (T5) at (1.3,0);
    
    \coordinate (T6) at (3.5,1);
    \coordinate (T7) at (3.7,0.3);
    \coordinate (T8) at (4.5,0);

    \coordinate (G) at (-2,0.5);
    \coordinate (Gx) at (-1.5,0.5);
    \coordinate (Gy) at (-2,1);
    
    \node[blue] at (A1) {$\times$};
    \node[blue] at (A2) {$\times$};
    \node[blue] at (A3) {$\times$};
    \draw[->,purple] ($0.5*(T4)+0.5*(T5)$) -- ($0.5*(T5)+0.5*(T4)+1.9*(-0.26,0.97)$);  
    \draw[purple,fill=purple!25, rotate=15] ($0.5*(T4)+0.5*(T5)$) ellipse (0.5 and 1.2);
    \node[purple] at (T4) {$\times$};
    \node[purple] at (T5) {$\times$};
    
    \draw[dashed,black] ($0.5*(T5)+0.5*(T4)+1.2*(-0.26,0.97)$) -- ($0.5*(T4)+0.5*(T5)+1.2*(-0.26,0.97) + 0.55*(1,0)$) ;  
    \draw[->,red] ($0.5*(T5)+0.5*(T4)+1.2*(-0.26,0.97) + 0.55*(1,0)$) arc (0:105:0.55);
    \node[red,above right] at ($0.5*(T5)+0.5*(T4)+1.2*(-0.26,0.97)$) {$\theta_1$};
    
    \draw[->,orange] ($0.5*(T6)+0.5*(T8)$) -- ($0.5*(T6)+0.5*(T8)+1.9*(-0.707,0.707)$);  

    \draw[orange,fill=orange!25,rotate = 45] ($0.5*(T6)+0.5*(T8)$) ellipse (0.7 and 1.2);
    \node[orange] at (T6) {$\times$};
    \node[orange] at (T7) {$\times$};
    \node[orange] at (T8) {$\times$};
    
    \draw[orange,fill=orange] ($0.5*(T6)+0.5*(T8)$) circle (0.05cm);
    \node[orange,below] at ($0.5*(T6)+0.5*(T8)+(0.4,0.2)$) {$G_2$};

     \draw[dashed,black] ($0.5*(T6)+0.5*(T8)+1.2*(-0.707,0.707)$) -- ($0.5*(T6)+0.5*(T8)+1.2*(-0.707,0.707) + 0.5*(1,0)$) ;  
     \draw[->,red] ($0.5*(T6)+0.5*(T8)+1.2*(-0.707,0.707) + 0.5*(1,0)$) arc (0:135:0.5);
     \node[red,above] at ($0.5*(T6)+0.5*(T8)+1.2*(-0.707,0.707)$) {$\theta_2$};
    \draw[->] (G) -- (Gx);
    \draw[->] (G) -- (Gy);
    \node[below] at (Gx) {$\vec{x}$};
    \node[left] at (Gy) {$\vec{y}$};
    \node[below left] at (G) {$\mathfrak{F}$};
   	\draw[dotted,gray,line width=0.3mm] (T4) -- (A1) -- (T5);
   	\draw[dotted,gray,line width=0.3mm] (T4) -- (A2) -- (T5);
   	\draw[dotted,gray,line width=0.3mm] (T4) -- (A3) -- (T5);
   	\draw[dotted,gray,line width=0.3mm] (T7) -- (A3) -- (T6);
   	\draw[dotted,gray,line width=0.3mm] (T8) -- (A3);
   	\draw[dotted,gray,line width=0.3mm] (T7) -- (T4) -- (T6);
   	\node[blue,below] at (A1) {$6$}; 
   	\node[blue,below] at (A2) {$7$}; 
   	\node[blue,below] at (A3) {$8$}; 
   	
   	\node[purple,below] at (T4) {$2$}; 
   	\node[purple,below] at (T5) {$1$};
   	\draw[purple,fill=purple] ($0.5*(T4)+0.5*(T5)$) circle (0.05cm);
   	\node[purple,below] at ($0.5*(T4)+0.5*(T5)$) {$G_1$};

   	\node[orange,below] at (T6) {$4$}; 
   	\node[orange,below] at (T7) {$5$}; 
   	\node[orange,below] at (T8) {$3$}; 
   	
   	\node[purple] at ($(T4)+(-1,0.5)$) {\large Robot $1$}; 
   	
   	\node[orange] at ($(T6)+(0.8,0.8)$) {\large Robot $2$}; 
   	
	\end{tikzpicture}
    \caption{Setup for two robots, seen as rigid bodies, carrying multiple tags.}
    \label{fig:setup_rigid_body}
\end{figure}
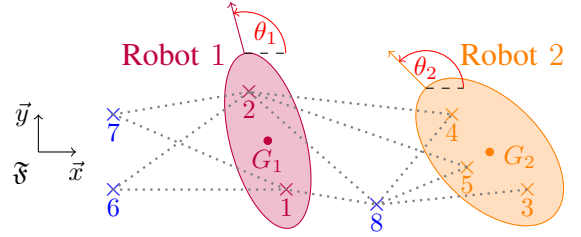

In this section, we consider scenarios where mobile robots can carry several tags, see Fig. \ref{fig:setup_rigid_body}. 
Hence, the relative motion and position of some tags are constrained by the fact
that they are attached to the same rigid body. More generally, let $\mbf f_c: \mbf R^{\sdim U} \to \mbf R^C$
be a known function defining $C$ constraints $\mbf f_c(\mbf p_{\mathcal U})=\mbf 0$ that the tag
positions must satisfy, and define the feasible set
\begin{equation}
\label{eq:feasible set}
\mathcal{C} \coloneqq \left\{ \mbf p = \col(\mbf p_{\mathcal U},\mbf p_{\mathcal K}) \in \r^{\sdim N} \big| 
\mbf f_c(\mbf p_\mathcal{U}) = \mbf 0 \right\}.
\end{equation}
To use the CRLB as localizability potential, the bound should now reflect the fact
that localization algorithms can leverage the information provided by the constraints
to improve their performance. 
We use the following result generalizing Proposition \ref{prop: constrained CRLB for network}.
\begin{proposition}
\label{prop: constrained CRLB for network refined}
Assume that the tag positions are subject to the constraints \eqref{eq:feasible set}.
Let $\mbf A_{\mathcal U}(\mbf p_{\mathcal U})$ be a matrix whose columns span
$\ker \partial \mbf f_c / \partial \mbf p_{\mathcal U}$
(which depends on $\mbf p_{\mathcal U}$ in general). 
Let $\hat{\mbf p}_{\mathcal U}$ be an unbiased estimate of the tag positions 
$\mbf p_{\mathcal U}$, based on the measurements $\tilde{\mbf d}$, the knowledge 
of the anchor positions $\mbf p_{\mathcal K}$, and the knowledge of the constraints
\eqref{eq:feasible set}. Then
\begin{align}   \label{eq: cov tags CRLB rigid}
\mathsf{cov}[\mbf{\hat{p}}_\mathcal{U}] \succeq \mbf B_{\mathcal U}(\mbf p),
\end{align}
where 
\begin{equation}
\label{eq: constrained CRLB network def}
\mbf B_{\mathcal U}(\mbf p) := \mbf A_{\mathcal U}
[\mbf A_{\mathcal U}^T \mbf F_{\mathcal U} \mbf A_{\mathcal U}]^{\dagger} 
\mbf A_{\mathcal U}^T.
\end{equation}
\end{proposition}

\begin{proof}
We have both the trivial constraint 
$\mbf f_t(\mbf p_\mathcal{U}) = \mbf p_{\mathcal K} - \mbf p^*_{\mathcal K} = \mbf 0$ with 
$\mbf p^*_{\mathcal K}$ the known positions of the anchors, 
and the equality constraint $\mbf f_c(\mbf p_{\mathcal U}) = \mbf 0$. 
Define $\mbf h(\mbf p) = \col (\mbf f_c(\mbf p_\mathcal{U}),\mbf f_t(\mbf p_\mathcal{K}))$. 
We then have :
\[
\frac{\partial \mbf h}{\partial \mbf p} = \begin{bmatrix}
\frac{\partial \mbf f_c}{\partial \mbf p_{\mathcal U}} & \mbf 0 \\
\mbf 0 & \mbf I_{\sdim K}
\end{bmatrix}.
\]
We apply the result of Theorem \ref{thm:gorman}, with the matrix $\mbf A$ in \eqref{eq:crlb}
\[
\mbf A = \begin{bmatrix} \mbf A_\mathcal{U} \\ \mbf 0 \end{bmatrix}
\text{ so }
\mbf F_c = \mbf A_\mathcal{U}^\top \mbf F_{\mathcal U} \mbf A_\mathcal{U}, 
\;
\mbf B_c = \begin{bmatrix} \mbf A_\mathcal{U}\mbf F_{c}^\dagger \mbf A_\mathcal{U}^\top  & \mbf 0 \\ \mbf 0 & \mbf 0 \end{bmatrix}.
\]
In \eqref{eq:crlb}, the $nU \times nU$ top-left corner of the matrix inequality gives 
\eqref{eq: cov tags CRLB rigid} for the covariance of $\hat{\mbf p}_{\mathcal U}$.
The other parts of the bound \eqref{eq:crlb} are trivial ($\mbf 0 \succeq \mbf 0$) 
and correspond to the fact that a reasonable estimate 
$\hat{\mbf p} = \text{col}(\hat{\mbf p}_{\mathcal U},\hat{\mbf p}_{\mathcal K})$ should set 
$\hat{\mbf p}_{\mathcal K} = \mbf p_{\mathcal K}$, so that $\hat{\mbf p}_{\mathcal K}$ 
will have zero covariance.
\end{proof}

Note that to simplify the notation, we have omitted in \eqref{eq: constrained CRLB network def} 
to state the dependencies $\mbf A_{\mathcal U}(\mbf p_{\mathcal U})$ and $\mbf F_{\mathcal U}(\mbf p)$.
From the matrix-valued bound \eqref{eq: constrained CRLB network def}, we can define
constrained localizability potentials as in Section \ref{section: localizability potentials unconstrained}.
Here, for conciseness, we only consider the A-Opt potential
\begin{equation}
\label{eq:constrained CRLB potential}
J_c(\mbf p):=\trace{\mbf B_{\mathcal U}(\mbf p)}.
\end{equation}
Moreover, the desired tag positions should also respect the constraints specified by \eqref{eq:feasible set}.
In other words, we aim to adjust the positions of the mobile nodes (anchors or tags) in
order to \diff{minimize, at least locally, the overall potential $J$, which includes the localizability
potential $J_c$ in \eqref{eq:constrained CRLB potential}}, subject to the constraints \eqref{eq:feasible set}.
For this, we can replace the gradient-descent method \eqref{eq:descent_per_agent} by the following first-order 
primal-dual method \cite[p. 528]{dimitri_p_bertsekas_nonlinear_2016}: 
\begin{equation}    \label{eq:lagrangian_descent}
\begin{cases}
\mbf p_{k+1} = 
\mbf p_{k} - \eta_k \left( 
\frac{\partial J(\mbf p_{k})}{\partial \mbf p}
+ 
\pmb \lambda_k^T \frac{\partial \mbf f_c(\mbf p_{\mathcal{U},k})}{\partial \mbf p} \right)^T,
\\
\pmb \lambda_{k+1} = \pmb \lambda_k + \delta  \, \mbf f_c(\mbf p_{\mathcal{U},k}),
\end{cases}
\end{equation}
where $\eta_k \in \r$ is a sequence of stepsizes,
$\delta$ a fixed parameter and $\pmb \lambda_k$ are dual variable iterates. 
The scheme \eqref{eq:lagrangian_descent} provides a sequence of configurations $\mbf p_k$, $k \geq 0$.
Feasibility of the constraints \eqref{eq:feasible set} is not maintained during the iterations
\eqref{eq:lagrangian_descent}, but the algorithm contributes to keeping $\mbf p_{k+1}$
close to $\mathcal C$.
In addition, for each iterate $\mbf p_k$ that we actually want to use as waypoint
for motion planning (some iterates could be skipped),
since \eqref{eq:feasible set} represents rigidity constraints, we can enforce feasibility 
by computing for each robot the pose minimizing the distance between the desired and 
achievable tag locations, in a least-squares sense (this corresponds to a standard 
pose estimation problem \cite[Section 8.1]{Barfoot:book17:stateEst}).

\diff{
A \emph{local} convergence result for the iterations \eqref{eq:lagrangian_descent} to a local 
constrained minimum $\mbf p^*$
and Lagrange multiplier $\boldsymbol \lambda^*$ is stated in \cite[Proposition 5.4.2]{dimitri_p_bertsekas_nonlinear_2016},
for constant stepsizes $\eta_k = \delta$, $k \geq 0$, and $\delta$ sufficiently small. 
Note that this method is not guaranteed to converge starting from any initial configuration $\mbf p_0$.
Hence, it may need to be combined with or replaced by other optimization
methods with global convergence guarantees, such as multiplier methods, as discussed 
in \cite[Section 5.2]{dimitri_p_bertsekas_nonlinear_2016}.
We refer the reader to the literature on nonlinear programming for further
discussion and comparison of available iterative methods, and focus instead in the rest of this section
on the computation of the derivatives $\partial J_c/\partial \mbf p$ and $\partial \mbf f_c / \partial \mbf p$
appearing in \eqref{eq:lagrangian_descent}, which are required for the implementation of all such methods.
}
We specialize the discussion above to the deployment problem 
where some some robots carry multiple tags, which requires evaluating
the cost function \eqref{eq:constrained CRLB potential} and its gradient.
First, we only take into account in the CRLB the constraints on the distances 
between the intra-robot tags, since this leads to somewhat simpler expressions
and computations.
In Section \ref{ss:CRLB_RP}, we include in the CRLB the full information about the relative 
positions of these tags.

\subsection{CRLB with Distance Constraints}
\label{ss:CRLB_rigid}
\label{ss:ext:dist}

Considering Fig. \ref{fig:setup_rigid_body}, as robots carrying multiple tags move,
their tags' relative positions must satisfy rigid displacement constraints. 
We partition the set of tags $\mathcal U$ into $R$ groups $\mathcal U_1, \ldots \mathcal U_R$, 
with $\sum_{r=1}^R |\mathcal U_r|  = U$, such that the tags in group $\mathcal U_r$ 
are rigidly connected (mounted on the same robot).
To simplify the discussion in the following, we assume that each group has
$|\mathcal U_r| \geq \sdim$ tags in dimension $\sdim$ and that these tags are in general position
(no $3$ tags aligned, and no $4$ tags coplanar in dimension $3$). As a result, each group of tags forms
an infinitesimally rigid framework for the complete graph (note that all pairwise distances within
a group $\mathcal U_r$ are known).
For example, we can simply have $2$ tags on each robot if $\sdim = 2$, or $3$ non-aligned 
tags if $\sdim = 3$. We also ignore the possibility of having 
known rigid constraints between anchors and tags. 
The analysis can be extended to mixed networks of robots carrying a single or multiple tags, 
or both anchors and tags, in a straightforward manner.

Since we know the relative positions of the tags in $\mathcal U_r$ in the robot's frame
of reference (\diff{by carefully placing them on the robot}), this information should \diff{in principle} 
be included in the CRLB. First, however, we only include the information about relative \emph{distances} 
between tags in each group, as this leads to simpler algorithms.
In this case, in the framework of Section \ref{section: constrainted loc}, $\mbf f_c$ has
one component for each pair of tags $\{i,j\}$ in the same group $\mathcal U_r$, of the form
\[
\mbf f_c^{\{i,j\}}(\mbf p_{\mathcal U}) = ||\mbf p_{ij}||^2 - d_{ij}^2,
\]
where $d_{ij}$ is perfectly known. 
If we order these components by listing all pairs of tags in the same set $\mathcal U_1$, 
$\mathcal U_2$, \ldots,  $\mathcal U_R$, 
then we obtain for the Jacobian matrix
\begin{align}
\label{eq: jacobian distance constraints}
\frac{\partial \mbf f_c(\mbf p_{\mathcal U})}{\partial \mbf p_{\mathcal U}} = \text{diag}(\mbf R_1,\ldots,\mbf R_R),
\end{align}
where $\mbf R_r$ is the rigidity matrix defined in Section \ref{ss:rigidity_theory}, for the
framework formed by a complete graph among the tags in group $\mathcal U_r$. 
Because the framework within each group is infinitesimally rigid, the kernel of each matrix $\mbf R_r$ 
is spanned by three explicitly known vectors if $\sdim = 2$, or six if $\sdim = 3$, as described in
Proposition \ref{prop:kerthree}. 
Then we can compute the 
matrix $\mbf A_{\mathcal U} = \begin{bmatrix} \mbf A_1 & \ldots & \mbf A_R \end{bmatrix}$ with $\sdim U$ 
rows and $3R$ (if $\sdim = 2$) or $6R$ (if $\sdim = 3$) columns spanning the kernel of \eqref{eq: jacobian distance constraints}.
For example, based on the discussion above Proposition \ref{prop:kerthree}, 
if $\sdim = 2$ we can take
$\mbf A_r = \begin{bmatrix} \mbf v_{T_x}^r & \mbf v_{T_y}^r & \mbf v_{R_z}^r \end{bmatrix}$, 
with $[\mbf v_{T_x}^{r}]_{2i-1}=1$, $[\mbf v_{T_y}^r]_{2i}=1$, $[\mbf v_{R_z}^r]_{2i-1} = -y_i$ 
and $[\mbf v_{R_z}^r]_{2i} = x_i$ for all $i \in \mathcal U_r$ and zeros everywhere else.
From these explicit expressions of $\mbf A_{\mathcal U}$, we can also immediately compute the derivatives
$\partial \mbf A_{\mathcal U} / \partial \xi_i$, for $\xi_i \in \{x_i,y_i,z_i\}$.

Since determining $\mbf A_{\mathcal U}(\mbf p_\mathcal U)$ allows us to compute 
$J_c(\mbf p_{\mathcal U})$ using \eqref{eq: constrained CRLB network def}, the only
missing element to execute the iterations \eqref{eq:lagrangian_descent} is the gradient
of $J_c$. 
\diff{For simplicity, suppose that $\mbf F_c \coloneqq \mbf  A_{\mathcal U}^T \mbf F_{\mathcal U} \mbf A_{\mathcal U}$ 
is invertible. Since $\mbf A_\mathcal{U}$ can be taken to be full column rank, this can be ensured by fulfilling the assumptions 
of Theorem \ref{thm: Fu invertible}, guaranteeing that $\Fu$ is invertible.
Then, we have}
\begin{align}
&\der{J_c}{\xi_i} = \der{}{\xi_i} \trace{
{\mbf A}_{\mathcal U} {\mbf F_c}^{-1} {\mbf A}_{\mathcal U}^T} \label{eq: gradient constrained CRLB}
\\
&= 2\trace{ {\mbf F}_c^{-1} \mbf A_{\mathcal U}^T \der{{\mbf A_{\mathcal U}}}{\xi_i}}
- \trace{{\mbf A_\mathcal{U}}  {\mbf F}_c^{-1} \der{ {\mbf F}_c}{\xi_i}  {\mbf F_c}^{-1} {\mbf A_\mathcal{U}}^\top} 
\nonumber \\
&= 2\trace{ {\mbf F}_c^{-1} \mbf A_{\mathcal U}^T (\mbf I - \mbf B_{\mathcal U} \mbf F_{\mathcal U}) \der{{\mbf A_{\mathcal U}}}{\xi_i} } 
- \trace{ \mbf B_{\mathcal U}^2 \frac{\partial \mbf F_{\mathcal U}}{\partial \xi_i}}.   \nonumber
\end{align}

\subsection{CRLB with Constrained Relative Positions}
\label{ss:CRLB_RP}

\diff{When we place two tags $i$ and $j$ on a robot $r$, we can in fact know the 
relative positions (RP) $\mbf p_{ij}^r$ of these tags in the frame of robot $r$,
not just their distance. Since a position estimator can leverage this information 
to improve its accuracy, we derive in this section the corresponding CRLB}.
To simplify the presentation, we assume here that each robot carries at least two tags.

To obtain the CRLB, let us first introduce $R$ new parameters $\boldsymbol{\theta} \coloneqq \col(\boldsymbol{\theta}_1,\ldots,\boldsymbol{\theta}_R)$, one for each 
robot, where $\boldsymbol{\theta}_i \in \mathbb R^{q}$, with $q = 1$ if $\sdim = 2$ and $q = 3$ if $\sdim = 3$.
Then, for the extended set of parameters 
$\tilde{\mbf{p}}_{\mathcal U} = (\mbf p_{\mathcal U},\boldsymbol \theta)$
and the measurements \eqref{eq:model_meas} or \eqref{eq:model_meas_lognormal}, we denote the 
extended FIM
\begin{equation}    \label{eq: Fu extended}
\tilde{\mbf{F}}_{\mathcal U} = 
-\esp{\der{^2 \ln f(\tilde{\mbf d};\tilde{\mbf p}_{\mathcal U})}{\tilde{\mbf p}_{\mathcal U} \partial 
\tilde{\mbf p}_{\mathcal U}^\top}} 
= \begin{bmatrix}
\Fu & \mbf 0_{\sdim U, qR}  \\
 \mbf 0_{qR, \sdim U} &  \mbf 0_{qR, qR} 
\end{bmatrix}.
\end{equation}
In the following, we add constraints between the tag positions and the parameters $\boldsymbol \theta$,
in such a way that the latter represent the robot orientations in exponential coordinates.
Then, \diff{we compute the constrained FIM from $\tilde{\mbf{F}}_{\mathcal U}$  
using Theorem \ref{thm:gorman} to obtain the final CRLB on position estimates}. 

It is convenient to number and order the tags as follows. Consider robot $r \in \{1,\ldots,R\}$ 
and associated tags $\mathcal U_r$, using the notation of Section \ref{ss:ext:dist}. 
Pick one tag in $\mathcal U_r$, denoted in the following $1^r$. The other tags of $\mathcal U_r$
are denoted $2^r,\ldots,U_r^r$, with $U_r = |\mathcal U_r|$. We group these latter tags by robot and 
list them in the order
\begin{align}   \label{eq: other tags}
\mbf p_{o} \coloneqq \col ( \mbf p_{2^1},\dots \mbf p_{U^1_{1}}, \dots, 
\mbf p_{2^R},\dots \mbf p_{U^R_{R}}) \in \mathbb R^{\sdim (U-R)},
\end{align}
from robot $1$ to robot $R$. The positions of the $R$ tags $1^r$ are also grouped in the vector
\[
\mbf p_c \coloneqq \col(\mbf p_{1^1},\ldots,\mbf p_{1^R}) \in \mathbb R^{\sdim R}.
\]
Then, we have $\tilde{\mbf p}_{\mathcal U} = \col(\mbf p_o,\mbf p_c, \boldsymbol{\theta})$.

Next, for each tag $j^r \in \mathcal U_r$ other than $1^r$, we add the constraint 
$\mbf f^{(r,j^r)}(\mbf p_{1^r},\mbf p_{j^r},\boldsymbol{\theta}_r) = \mbf 0 \in \mathbb R^\sdim$, where
\begin{equation}    \label{eq: RP constraint exp}
\mbf f^{(r,j^r)}(\mbf p_{1^r},\mbf p_{j^r},\boldsymbol{\theta}_r) = 
\mbf p_{j^r} -\mbf p_{1^r} - \exp(\cpop{\boldsymbol{\theta}_r}) \mbf p^r_{j^r 1^r},
\end{equation}
with the notation (depending if $\sdim = 2$ or $\sdim = 3$)
\begin{align*}
\cpop{\theta} &= \begin{bmatrix} 0 & -\theta \\ \theta & 0 \end{bmatrix}, \text{ if } \theta \in \mathbb R, \\
\cpop{\boldsymbol \theta} &= \begin{bmatrix}
0 & -\theta_z & \theta_y \\ \theta_z & 0 & -\theta_x \\ -\theta_y & \theta_x & 0
\end{bmatrix}, \text{ if } \boldsymbol \theta = [\theta_x, \theta_y, \theta_z]^T \in \mathbb R^3.
\end{align*}
There are $U_r-1$ constraints of the form \eqref{eq: RP constraint exp} for robot $r$, each
of dimension $\sdim$, which represent a change from the known coordinates $\mbf p^r_{j^r 1^r}$ 
in the robot frame to the (unknown) coordinates $\mbf p_{j^r 1^r}$ in the world frame $\mathfrak{F}$, 
with the matrix $\exp(\cpop{\boldsymbol{\theta}_r})$ representing the rotation matrix from $\mathfrak{F}$ 
to the frame of robot $r$, using the exponential coordinate representation \cite{Lynch2017ModernRM}. 
\newcommand{\tr}{\boldsymbol{\theta_r}}
\newcommand{\rot}{\mbf R_{\tr}}
Define in the following the notation $\exp(\cpop{\boldsymbol{\theta}_r}):=\rot$ and
\[
\boldsymbol \Phi_{\boldsymbol \theta_r}^{(r,j^r)} \coloneqq 
\rot \, \mbf p^r_{j^r 1^r}, \;\;
\text{ for } j^r \in \mathcal U_r, 1 \leq r \leq R.
\]

\begin{remark}
Recall that when $\sdim = 2$, we have simply
\[
\exp(\cpop{\theta}) = \begin{bmatrix} \cos(\theta) & -\sin(\theta) \\ \sin(\theta) & \cos(\theta) \end{bmatrix},
\]
and when $\sdim = 3$, $\exp(\cpop{\boldsymbol \theta})$ can be computed efficiently using
Rodrigues' formula \cite[Proposition 3.1]{Lynch2017ModernRM}.
\end{remark}

Considering \eqref{eq: RP constraint exp} for all $R$ robots, we obtain $U-R$ constraints on 
the parameters $\tilde{\mbf p}_{\mathcal U}$, each of dimension $\sdim$. We list these constraints
in the same order as for $\mbf p_o$ in \eqref{eq: other tags} and denote them
$\mbf f_{\text{RP}}(\mbf{p}_o,\mbf p_c,\boldsymbol{\theta}) = \mbf 0$. For the constrained
CRLB, we are interested in the kernel of the Jacobian matrix of $\mbf f_{\text{RP}}$.
Remark that with the chosen ordering of tags and constraints, we have
$\frac{\partial \mbf f_{\text{RP}}}{\partial \mbf{ p}_o} = \mbf I_{\sdim (U-R)}$.
If we define 
\begin{equation}    \label{eq: partial Jacobian RP}
\mbf N \coloneqq \begin{bmatrix} \frac{\partial \mbf f_{\text{RP}}}{\partial \mbf{ p}_c} & 
\frac{\partial \mbf f_{\text{RP}}}{\partial \boldsymbol{\theta}} \end{bmatrix},
\end{equation}
\diff{and $\mbf A_\text{RP} \coloneqq \spanv { \ker \frac{\partial \mbf f_{\text{RP}}}{\partial \mbf{\tilde p}_\mathcal U} }$,}
then immediately
\begin{align}
\mbf A_\text{RP}
&= \spanv { \ker \begin{bmatrix} \mbf I_{\sdim (U-R)} & \mbf N \end{bmatrix} } \nonumber \\
&= \col \left( -\mbf N, \mbf I_{(n+q)R} \right).  \label{eq: A_RP}
\end{align}
Indeed, $\frac{\partial \mbf f_{\text{RP}}}{\partial \mbf{\tilde p}_\mathcal U}$ is of rank $\sdim(U-R)$,
so $\mbf A_{RP}$ should have $\sdim U + q R - \sdim(U-R) = (n+q)R$ independent columns, and clearly
\[
\frac{\partial \mbf f_{\text{RP}}}{\partial \mbf{\tilde p}_\mathcal U} \mbf A_{\text{RP}} = 
-\mbf N + \mbf N = \mbf 0.
\]
Hence, it is sufficient to compute $\mbf N$ to obtain $\mbf A_{\text{RP}}$.

\begin{proposition}
The matrix $\mbf N$ in \eqref{eq: partial Jacobian RP} is defined by
\[
\mbf N = \col \left( \{\mbf N^{(r,j^r)} \}_{1 \leq r \leq R,2^r \leq j^r \leq U_r^r} \right)
\; \in \mathbb R^{\sdim (U-R) \times (\sdim+q) R}.
\]
where the blocks $\mbf N^{(r,j^r)} \in \mathbb R^{\sdim \times (n+q)R}$ are stacked in the same 
order as $\mbf p_o$ in \eqref{eq: other tags} and are of the form
\[
\mbf N^{(r,j^r)} = - \begin{bmatrix}
\mbf 0_{\sdim, \sdim(r-1)} & \mbf I_{\sdim} & \mbf 0_{\sdim, s} & \mbf N_{\boldsymbol{\theta}_r}^{(r,j^r)}
& \mbf 0_{\sdim, (R-r)q}
\end{bmatrix}
\]
with $s = (R-r) \sdim +(r-1) q$, where
\diffblock
\begin{equation}
\mbf N_{\boldsymbol \theta_r}^{(r,j^r)} =
\begin{cases}
[1]_\times \boldsymbol \Phi_{\boldsymbol \theta_r}^{(r,j^r)}
\in \mathbb R^{2} \text{ if } \sdim = 2, \\
\left[ \boldsymbol \Phi_{\boldsymbol \theta_r}^{(r,j^r)} \right]_\times 
\boldsymbol \Omega_{\boldsymbol{\theta}_r}  \in \r^{3\times 3} \text{ if } \sdim = 3,
\end{cases}
\label{eq:RP:ntheta}
\end{equation}
with
$\boldsymbol \Omega_{\boldsymbol{\theta}_r} := 
(\boldsymbol \theta_r \boldsymbol \theta_r^\top + ( \mbf I_3  - \rot) [\boldsymbol \theta_r]_\times) \| \boldsymbol \theta_r \|^{-2}
$.  
\end{proposition}
\diffend

\begin{proof}
Decompose $\mbf N^{(r,j^r)}$ by blocks
\[
\mbf N^{(r,j^r)} = \begin{bmatrix} 
\mbf G_1 & \ldots \mbf G_R & \mbf H_1 & \ldots & \mbf H_R \end{bmatrix}
\]
with $\mbf G_i \in \mathbb R^{n \times n}$ and $\mbf H_i \in \mathbb R^{n \times q}$.
The matrix $\mbf N^{(r,j^r)}$ is obtained by taking
the partial derivatives of $\mbf f^{(r,j^r)}$ in \eqref{eq: RP constraint exp}
with respect to the coordinates of $\mbf p_{1^r}$, which gives the block 
$\mbf G_r = -\mbf I_{\sdim}$, and with respect to the coordinates 
of $\boldsymbol{\theta}_r$, which gives the block 
$\mbf H_r = -\mbf N_{\boldsymbol{\theta}_r}^{(r,j^r)} \in \mathbb R^{n \times q}$. 
All other blocks are zero.
\diffblock
The expression of $\mbf H_r$ comes from the fact that 
$\cpop{\boldsymbol \theta_r} = \theta_r \cpop{1}$ when $n = 2$, whereas
when $n=3$, we have 
\[
\frac{\partial \boldsymbol \Phi_{\boldsymbol \theta_r}^{(r,j^r)}}{\partial \boldsymbol{\theta}_r}
= - \rot [\mbf p_{j^r1^r}^r]_\times \frac{\boldsymbol{\theta}_r \boldsymbol{\theta}_r^\top + (\rot^\top - \mbf I_3)[\boldsymbol{\theta}_r]_\times}{\|\boldsymbol{\theta}_r\|^2}
\]
from \cite[Result 1]{gallego_compact_2015}. This expression is further reduced to the one
in \eqref{eq:RP:ntheta} using elementary properties of rotation matrices.
\end{proof}
\diffend
With the matrices $\tilde{\mbf F}_{\mathcal U}$ and $\mbf A_\text{RP}$ defined in \eqref{eq: Fu extended}
and \eqref{eq: A_RP}, we can follow the discussion of Section \ref{section: constrainted loc} and define
$\mbf B_\text{RP} :=  \mbf A_\text{RP} [\mbf A_\text{RP}^\top \tilde{\mbf{F}}_{\mathcal U} 
\mbf A_\text{RP}]^{\dagger} \mbf A_\text{RP}^\top$ to obtain a CRLB taking the RP constraints
into account. 
\diff{We can build a cost function providing a lower bound on MSE of the tag positions as}
\begin{equation}    \label{eq:constrained CRLB potential - bis}
J_c (\mbf p) 
= \trace{\mbf C \mbf B_\text{RP} \mbf C^\top},
\end{equation}
similarly to \eqref{eq:constrained CRLB potential}, where \diff{
$\mbf C = [\mbf I_{nU} \; \mbf 0_{nU, qR}]$ is introduced here to 
select the $nU\times nU$ first block of $\mbf B_\text{RP}$} and hence 
consider only the uncertainty in the estimate $\hat{\mbf p}_{\mathcal U}$.
\diff{Alternatively, the uncertainty in the estimate of the whole
extended state $\tilde{\mbf p}_{\mathcal U}$ can be considered by
using the matrix $\mbf C = \text{diag}(\mbf I_{nU}, w_\theta \mbf I_{qR})$,
with $w_\theta$ a weight to select.} 
To compute the gradient with respect to $\mbf p$ for \eqref{eq:lagrangian_descent}, similarly to
\eqref{eq: gradient constrained CRLB}, we have, for $\xi \in \{x,y,z\}$ : 
\begin{equation}
\der{J_c}{\xi_i}=2\trace{\mbf C \der{\mbf A_{\text{RP}}}{\xi_i}\mbf D^\top}
- \trace{\mbf D \der{\mbf F_c}{\xi_i} \mbf D^\top},
\label{eq:derJcRP}
\end{equation}
with $\mbf D := \mbf C \mbf A_{\text{RP}} \mbf F_c^{-1}$, assuming
$\mbf F_c = \mbf A_\text{RP}^\top \tilde{\mbf{F}}_{\mathcal U} \mbf A_\text{RP}$
to be invertible. To compute the derivative $\partial \mbf A_{\text{RP}} / \partial \xi_i$, 
it is sufficient to know how to compute the terms
\diffblock
$\partial \mbf N_{\boldsymbol \theta_r}^{(r,j^r)} / \partial \xi_i$.
Then, noting that $\boldsymbol \Phi_{\boldsymbol \theta_r}^{(r,j^r)} = \mbf p_{j^r} -\mbf p_{1^r}$, the differentiation of \eqref{eq:RP:ntheta} yields
\begin{equation*}
	\der{\mbf N_{\tr}^{(r,j^r)}}{\xi_i}
	=
	\begin{cases}
	\cpop{1} \mbf e_\xi \, \psi_{j^r}(i) \text{ if } \sdim = 2; \\
	\cpop{\mbf e_\xi} \boldsymbol{\Omega}_{\tr} \, \psi_{j^r}(i) \text{ if } \sdim=3,
	\end{cases}
\end{equation*}
for $\xi \in \{x,y,z\}$, where $\mbf e_x, \mbf e_y, \mbf e_z$ forms the canonical basis of 
$\r^3$, and we introduced the notation
$\psi_{j^r}(i)$ equals to $1$ if $i=j^r$, to $-1$ if $i=1^r$ and to zero otherwise.

\section{Simulations}

\label{sec:sim_standard}

In this section, we present simulation results for two deployment scenarios. The first scenario
is a structure inspection problem by a multi-robot network maintaining localizability while 
the task is performed. The second concerns the deployment of an Unmanned Ground
Vehicle (UGV) carrying several tags, where we include the distance and relative 
position constraints in the CRLB-based potential.

\subsection{Cooperative Structure Inspection}

\diffblock

Consider a system composed of $N=16$ agents, with $U=12$ tags carried by mobile robots 
(\textit{i.e.}, $\mathcal{U} = \{1,\dots,12\}$) and $K=4$ fixed anchors 
with known positions (\textit{i.e.}, $\mathcal{K} = \{13,\dots,16\}$). 
Each robot carries an UWB transceiver to communicate and take ranging measurements
with any other robot or UWB anchor, following the model \eqref{eq:model_meas},
via a Two Way Ranging (TWR) protocol \cite{mai_local_2018,prorok_models_2013}.

\begin{figure}[h]
	\centering
	\includegraphics[width=0.9\linewidth]{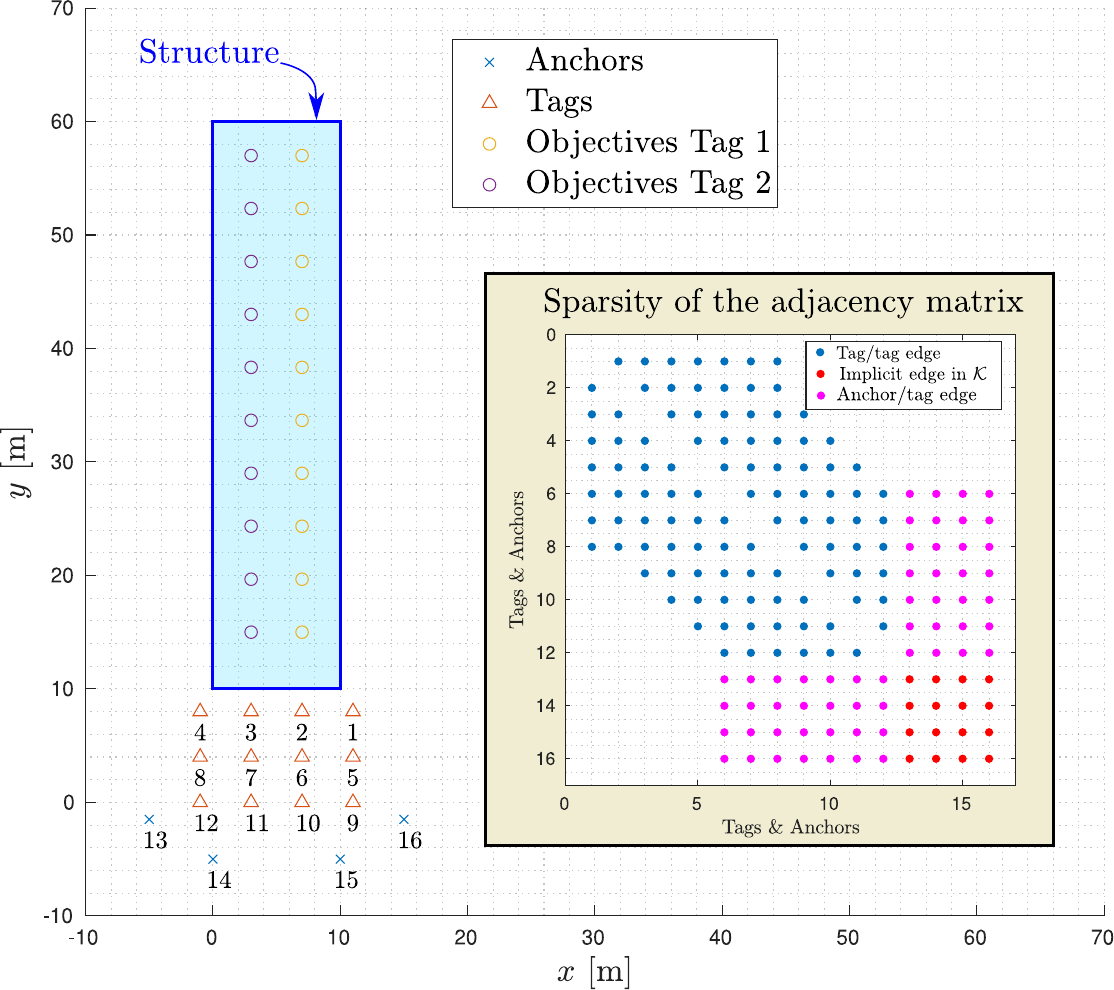}
	\diffcaption{Initial system configuration, waypoints for the leaders $1$ et $2$ and ranging network sparsity.}
	\label{fig:sim:setup}
\end{figure}

We assign an inspection task to the two first robot-tags $1$ and $2$, called ``leaders'', 
while the remaining robot-tags $\mathcal{U}_F =\{i\in \mathcal{U},i>2\}$ are called ``followers''
and deploy to support accurate localization. 
The leaders are required to visit ten waypoints each, underneath a $50~\mathrm{m} \times 10 ~\mathrm{m}$ 
rectangular structure represented in blue in Fig. \ref{fig:sim:setup}, in order to inspect it.
The links in the ranging/communication network are represented on Fig. \ref{fig:sim:setup} 
by the sparsity pattern of the network adjacency matrix, i.e., showing its non-zero entries.
In particular, we stress that the leaders cannot communicate directly with the anchors. 

\subsubsection{Motion Planner for the Follower Robots}
\label{sss:motionPlannerSim}
We follow the motion planning framework based on artificial potentials presented
in Section \ref{sec:statement}.
To enhance the localizability of the robots, we chose to include in the overall potential
the cost $J_D(\mbf p)=-\log\det\Fu$ introduced in Section \ref{sec:loca_potentials}.
\jcm{This choice is motivated in particular by the fact that in a decentralized system, computing 
the gradient of $J_D$ via Algorithm \ref{algo:dopt} requires a single distributed matrix inversion.}
We add safety margins between robots by introducing a collision avoidance potential
$$
J_\text{avd}(\mbf p) =  \frac{1}{2} \sum_{i\in \mathcal{U}} \sum_{j\in \mathcal{U} \cup \mathcal{K}} \left(d_{ij}^{-1}-{d_a}^{-1}\right)^2 \mathsf{1}_{d_{ij}<d_a}.
$$ 
We also encourage ranging tags to maintain proximity, in order to limit the potential 
deterioration of ranging measurements at long distances, e.g., due to power fading.
To do so, we use the potential
$$
J_\text{con}(\mbf p) =  \frac{1}{2} \sum_{i\in \mathcal{U}} \sum_{j\in \mathcal{N}_i} \left( d_{ij}-d_c\right)^2 \mathsf{1}_{d_{ij}>d_c}.
$$
In our simulations, we set $d_a=2~\text{m}$ and $d_c=50~\text{m}$.

Therefore, the overall potential is defined as 
$J(\mbf p) := K_l J_D(\mbf p) +  K_c J_{con}(\mbf p) + K_a J_\text{avd}(\mbf p)$  
where $K_l, K_a, K_c>0$ are constant parameters. 
The leaders travel directly to their prespecified waypoints. Meanwhile,
each follower $i \in \mathcal{U}_F$ implements the following \diff{gradient descent} scheme
\begin{equation}
\mbf p_{i,k+1}^d = \hat{\mbf p}_{i,k} - \frac{\partial J(\hat{\mbf p}_{\mathcal{U},k})}{\partial {\mbf p}_{i,k}} \times 
\min \left\{1,\frac{\Delta_\text{vel}}{\|\partial J/\partial {\mbf p}_{i,k}\|}\right \},
\label{eq:simulations:potdesc}
\end{equation}
i.e., with robot $i$ at its current position $\mbf p_{i,k}$ at period $k \geq 0$, 
a gradient step provides the next desired position $\mbf p^d_{i,k+1}$.
The $\min$ term bounds the stepsizes so that $\|\mbf p^d_{i,k+1}-\mbf p_{i,k}\|\leq \Delta_\text{vel}$,
for some specified value of $\Delta_\text{vel}$.
For $\xi_i \in \{x_i,y_i\}$, we compute $\partial J_D/\partial \xi_i$ by \eqref{eq:der:Dopt}, 
possibly using Richardson iterations presented in Algorithm \ref{algo:dopt} for a decentralized
implementation.
The expressions of the derivatives $\partial J_\text{con} / \partial {\xi_i}$ and 
$\partial J_\text{avd} / \partial {\xi_i}$ of the other potentials
are standard \cite{Lynch2017ModernRM} and can be distributively computed since they only depend 
on each tag's neighborhood. Note that in \eqref{eq:simulations:potdesc} we do not
assume that the true positions are accessible but compute the gradients at the
estimates $\hat{\mbf p}_{i,k}$ (see \eqref{eq:quadratic_cost}).

\begin{figure}[htbp!]
	\centering
	\includegraphics[width=0.9\linewidth]{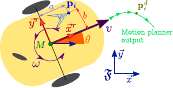}
	\caption{Robot and tag configuration for trajectory tracking. $(M,\vec{x}^r,\vec{y}^r)$ is the robot frame.}
	\label{fig:controleptapt}
\end{figure}

\diffend
The gradient descent scheme is used to obtain desired waypoints
for the tags, 
which we can track using controllers on the robots.
For concreteness, assume that all robots are identical with \diff{unicycle} kinematics \cite[Chap. 4]{corke_robotics_2011}
\begin{equation}
\dot{x}_M = v\cos(\theta), \quad
\dot{y}_M = v\sin(\theta), \quad
\dot{\theta} = \omega
\label{eq:kine_monocycle}
\end{equation}
where $\omega$ and $v$ are the rotational and translational velocities 
and $\theta$ is the robot's heading with respect to $\mathfrak{F}$.
The coordinates of the tag in the robot's frame (for any $i$) are $\mbf p_i^r = [a,b]^\top$, 
with $a \neq 0$, see Fig. \ref{fig:controleptapt}.
With
$\dot{\mbf p}_i \in \mathbb R^2$
the velocity of tag $i$ in $\mathfrak{F}$, 
implementing the following Proportional-Integral (PI) controller
\begin{equation}
\dot{\mbf p}_i = K_P(\mbf p^d_i(t) - \mbf p_i(t)) + K_I \int_{\tau = 0}^t (\mbf p_i^d(\tau) - \mbf p_i(\tau))d\tau,
\label{eq:velcontroller}
\end{equation}
with $K_P, K_I >0$, 
allows the tags to track the desired \diff{(piecewise constant)} trajectory $\mbf p^d$. 
This \diff{corresponds to} a velocity command $\mbf u_i := [v_i,\omega_i]^\top$ for robot $i$,
since 
$\mbf u_i = \mbf T (\theta_i) \dot{\mbf p}_i$ \cite[Section 13.3.1.4]{Lynch2017ModernRM} with
\[
\mbf T(\theta) =
\frac{1}{a} 
\begin{bmatrix}
a \cos \theta - b \sin \theta & a \sin \theta + b \cos \theta \\
- \sin \theta & 
\cos \theta 
\end{bmatrix}.
\]
\diffblock

\diffblock

\subsubsection{Simulation and Performance Analysis}
\label{ss:sim:perf}

We choose the
weights in the potential $J$ as $K_l = 5\times 10^4$, $K_a = K_c = 1 \times 10^3$
and the maximal step length $\Delta_{vel} = 2~\mathrm{m}$. 
When the leaders reach their $o$-th waypoint, we repeat the iterations \eqref{eq:simulations:potdesc}
$N_\text{iter}=30$ times to compute sufficiently distant waypoints for the followers. 
Then, we only transmit the desired position $\mbf p_{i,o N_\text{iter}}^{d}$ to the controller of
each follower $i \in \mathcal{U}_F$ in order to enhance the tags' localizability.  
The tags are positioned on the robots so that $a = b = 0.5~m$, 
and the PI controller gains are $K_p = 3$, $K_i = 0.5$.
The controller \eqref{eq:velcontroller} follows the trajectory computed 
from \eqref{eq:simulations:potdesc} with a maximum tracking error of about $10$ cm.

To illustrate the performance of our deployment scheme we perform 
$M = 1000$ Monte Carlo simulations, using the measurement model \eqref{eq:model_meas} with $\sigma = 5~\mathrm{cm}$. 
At simulation $\rho$, the position estimates $\hat{\mbf p}_{\mathcal{U},k}^\rho$ 
used in \eqref{eq:simulations:potdesc}
are obtained by solving the least-squares problem
\begin{align}
\hat{\mbf p}_{\mathcal{U},k}^\rho 
&= \underset{\mbf p_{\mathcal{U}} \in \r^{2 U}}{\text{argmin}} \; Q(\mbf p_\mathcal{U}), \nonumber \\
\text{with }
Q(\mbf p_\mathcal{U}) &\coloneqq
\sum_{i\in \mathcal{U}} \sum_{j\in \mathcal N_i} ( \| {\mbf p}_{i,k} -  {\mbf p}_{j,k}\| -\tilde{d}_{ij,k}^\rho )^2,
\label{eq:quadratic_cost}
\end{align}
where 
${\mbf p}_{j,k}$ is the anchor position in \eqref{eq:quadratic_cost} if $j \in \mathcal{K}$
and $\tilde{d}_{ij,k}^\rho$ are the range measurements.

\begin{figure}[h]
	\centering
	\includegraphics[width=\linewidth]{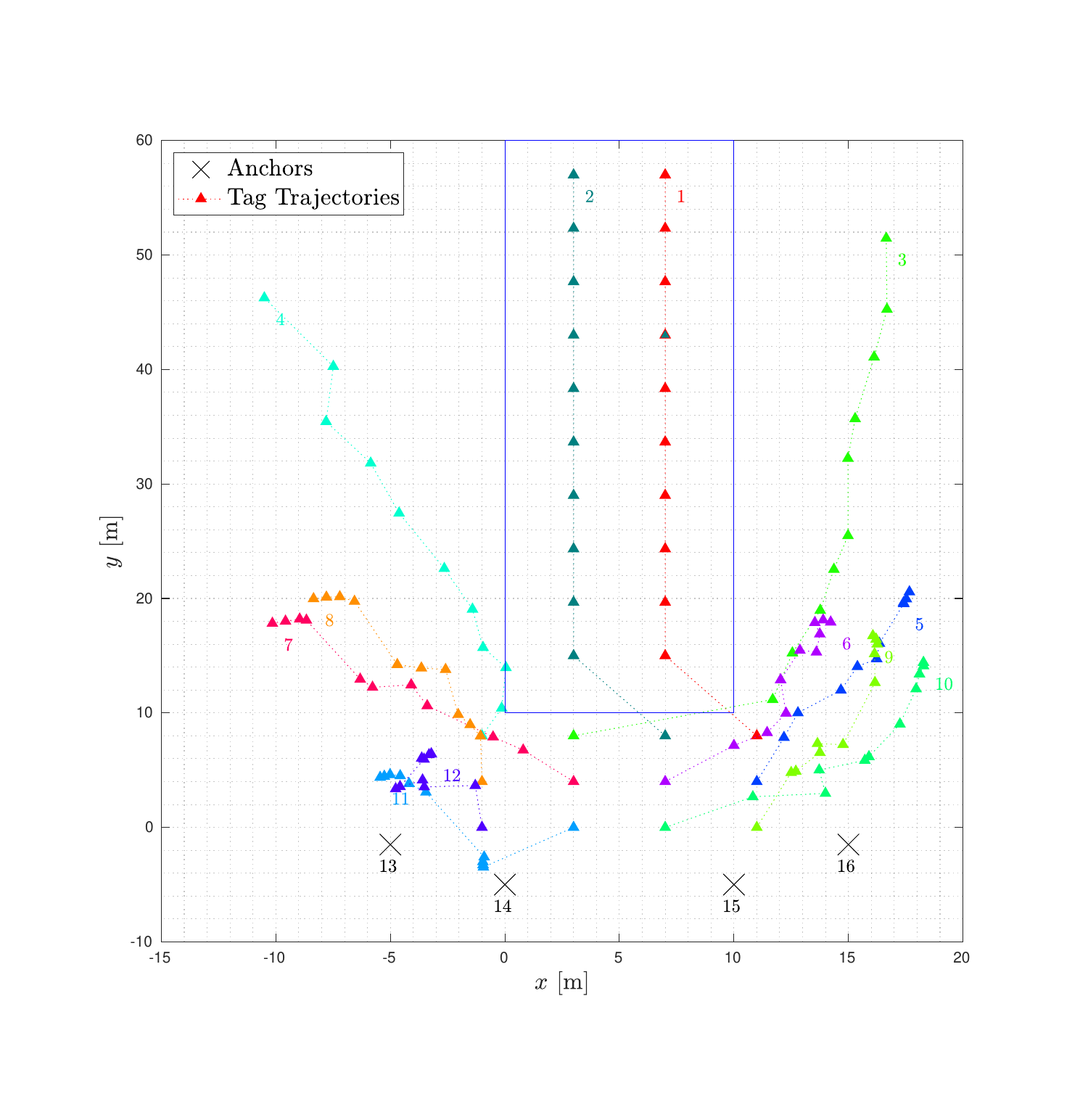}
	\vspace{-15mm}
	\diffcaption{Tag trajectories in the workspace.}
	\label{fig:sim:xydepl}
\end{figure}

\begin{figure}[htbp!]
	\centering
	\includegraphics[width=0.9\linewidth]{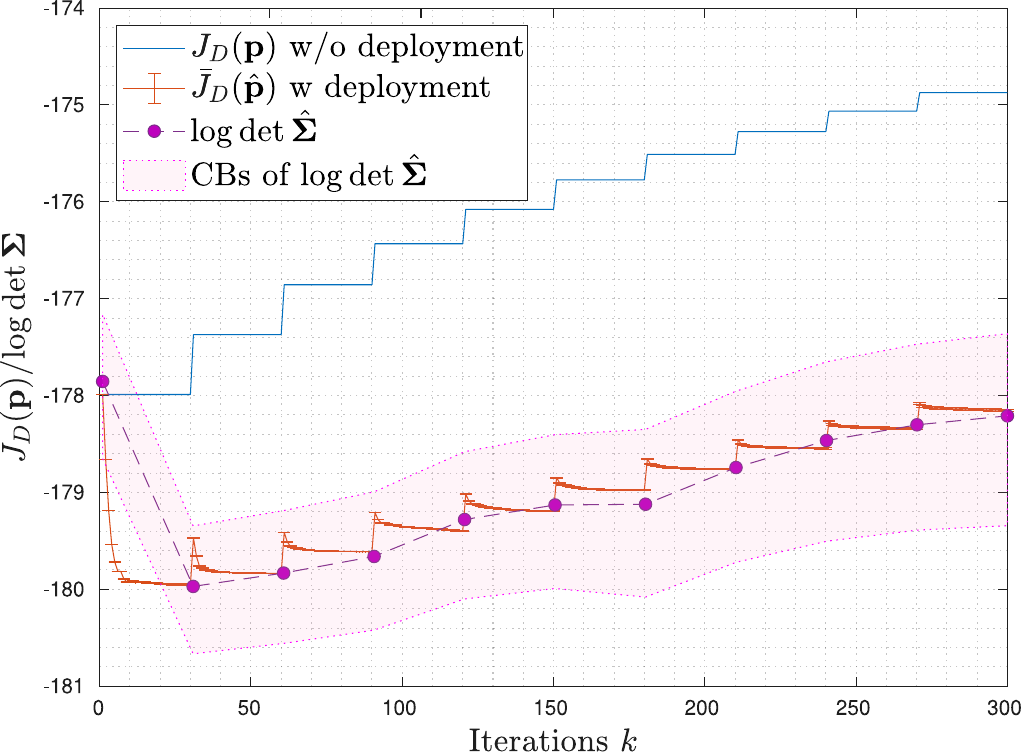}
	
	\diffcaption{Localizability potential with and without follower deployment, 
	empirical entropy and $3\sigma$ confidence bounds obtained from the Monte-Carlo simulations. 
	The leaders' waypoints are updated every $N_\text{iter}=30$ iterations of the 
	gradient descent scheme.}
	\label{fig:sim:potsuperp}
\end{figure}

As shown by the trajectories on Fig. \ref{fig:sim:xydepl}, the leaders 
follow their assigned paths and the followers maintain the network's localizability. 
Initially, all robots are aligned, a geometry with poor localizability. 
On Fig. \ref{fig:sim:potsuperp}, we plot the empirical average $\bar{J}_D$ and $3\sigma$
confidence bounds (CBs) for the potential $J_D$ over the $M$ simulations. 
Initially, the localizability potential decreases as the followers deploy.
The following step increases occur when the leaders move to their next waypoints
and the network extends, while the anchors remain fixed and far away, see Fig. \ref{fig:sim:xydepl}. 
Overall however, the followers manage to keep the localizability at a low value.
For comparison, we plot in blue on Fig. \ref{fig:sim:potsuperp} the evolution of 
the localizability potential without deployment of the followers.
We also plot the empirical 
statistical entropy $\ln \det \tilde{\Sigma}_k$ and its CBs,
with $\tilde{\Sigma}_k$ the empirical covariance of the estimates 
$\hat{\mbf p}_{\mathcal{U},k}$ obtained by solving \eqref{eq:quadratic_cost}. 
The plot highlights that the entropy remains close to the theoretical
lower bound provided by $J_D$, as discussed 
in Section \ref{section: localizability potentials unconstrained}.

\begin{figure}[h]
	\centering
	\includegraphics[width=0.8\linewidth]{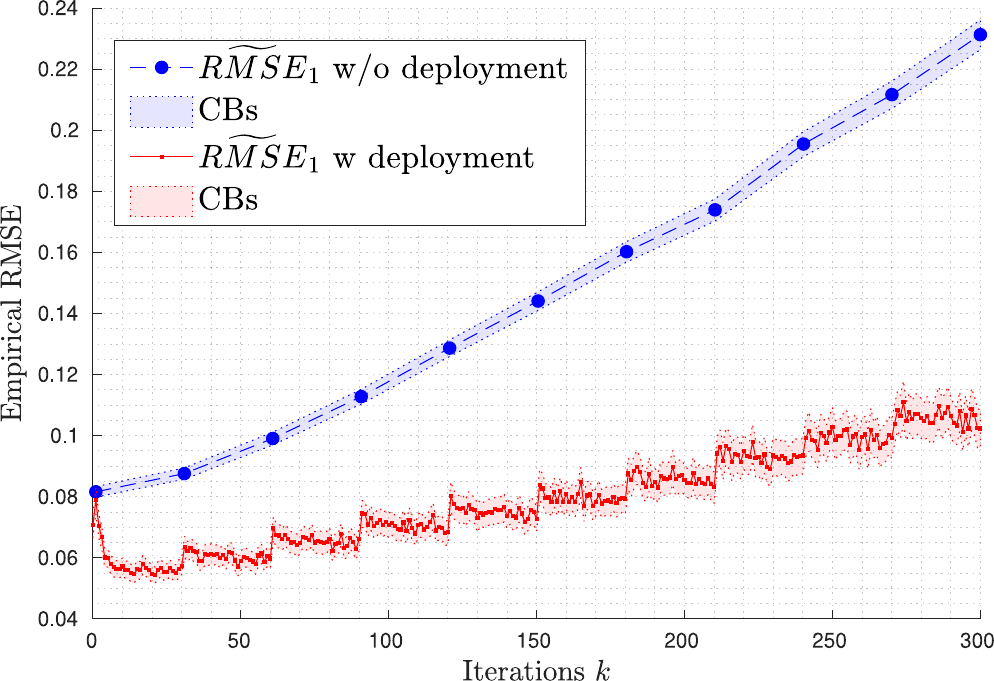}
	\diffcaption{Plot of the empirical RMSE over the trajectory.}
	\label{fig:sim:rmse}
\end{figure}

Even though the deployment is performed here using $J_D$ to measure localizability, 
which is related to entropy, Fig. \ref{fig:sim:rmse} shows that other localization
accuracy measures are improved as well. In this case, we plot the empirical
Root Mean Squared Error (RMSE) for the location estimate of the first leader tag,
the plot for the second leader being similar.
Namely, at each iteration $k$ of \eqref{eq:simulations:potdesc}, 
we compute the empirical MSE 
$\widetilde{MSE}_{1,k} := \frac{1}{M} \sum_{\rho=1}^M \|\hat{\mbf p}_{1,k}^\rho -\mbf p_{1,k}\|^2$,
with $\hat{\mbf p}_{1,k}^\rho$ the estimate of $\mbf p_{1,k}$ for simulation $\rho$. 
Then $\widetilde{RMSE}_1 \coloneqq (\widetilde{MSE}_{1,k})^{1/2}$.
The CBs shown on Fig. \ref{fig:sim:rmse} are defined by $b_{\pm,k}=s^{1/2}_{\pm,k}$,
where $s_{\pm,k} =  \widetilde{MSE}_{1,k} \pm 3 \widetilde{\sigma}_{1,k}/\sqrt{M}$,
with $\widetilde{\sigma}_{1,k}^2 
= \frac{1}{M-1} \sum_{\rho=1}^M [\|\hat{\mbf p}_{1,k}^\rho -\mbf p_{1,k}\|^2 - \widetilde{MSE}_{1,k}]^2$ 
the empirical variance of the samples.
For comparison, we also plot $\widetilde{RMSE}_1$ without deployment.
The empirical RMSE is significantly reduced by the motion of the followers, 
remaining below $12$ cm even when the leader $1$ is at its farthest waypoint.

\subsubsection{Distributed Gradient Computations}

\begin{figure}[h]
	\begin{subfigure}{0.49\linewidth}
		\centering
		\includegraphics[width=\linewidth]{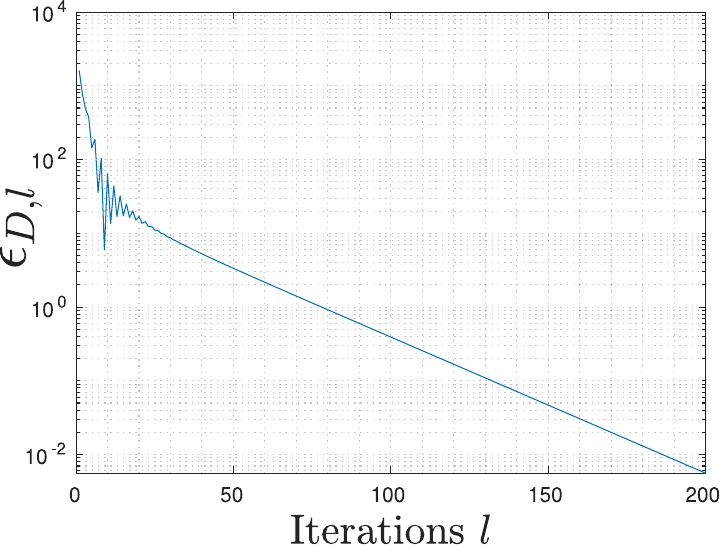}
		\diffcaption{Errors on D-Opt gradient.}
		\label{fig:sim:convDopt}
	\end{subfigure}
	\begin{subfigure}{0.49\linewidth}
		\centering
		\includegraphics[width=\linewidth]{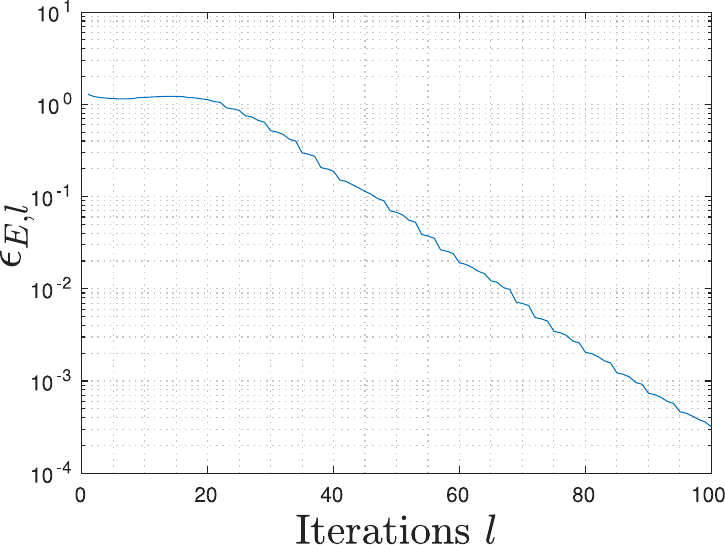}
		\diffcaption{Errors on E-Opt gradient.}
		\label{fig:sim:convEopt}
	\end{subfigure}
	\diffcaption{Convergence of the D-Opt and E-Opt gradient estimates for the last configuration in the trajectory.}
	\label{fig:sim:convDecentralizedPotential}
\end{figure}

Here we illustrate the convergence of the distributed algorithms of Section \ref{sec:gradients} 
estimating the gradients of the localizability potentials, 
more specifically Algorithm \ref{algo:dopt} (D-Opt) and Algorithm \ref{algo:distriEoptVect} (E-Opt).
Define the relative error $\epsilon_{S_l}$ on the gradients at the $l$-th iteration as follows
\[
\epsilon_{S,l} =  \frac{\| \widehat{[\lder{J_S}{\mbf p_\mathcal{U}}}]_l - \lder{J_S}{\mbf p_\mathcal{U}} \|_2}{\|\lder{J_S}{\mbf p_\mathcal{U}}\|_2} 
\]
for each scheme $S \in \{D,E\}$ producing the estimates $[\widehat{\lder{J_S}{\mbf p_\mathcal{U}}}]_l$. 
On Fig. \ref{fig:sim:convDecentralizedPotential} we plot the errors $\epsilon_{E,l}$ and $\epsilon_{D,l}$
for increasing values of $l$
at the last (fixed) configuration $\mbf p_\mathcal{U}$ of the trajectory shown on Fig. \ref{fig:sim:xydepl}.
For the D-Opt scheme, we arbitrarily choose the initial condition $\mbf x_0 = \mbf I_{n\times u}$ 
in Algorithm \ref{algo:dopt}, which is far from the ideal value $\Fu^{-1}$. 
Nonetheless, an error of $10\%$ on the gradient is obtained after about $120$ iterations. 
To estimate the gradient of $J_E$, we arbitrarily set $\mbf w(0)=\mbf 1_{n U}$ in Algorithm \ref{algo:distriEoptVect}. 
In this case a relative error of $10\%$ is obtained after $50$ iterations, with the inner loop to compute 
the squared norm of the eigenvector set to $\tilde{n}_\text{iter}=10$. 

The convergence speed of both algorithm depends on the structure of $\Fu$ and the chosen initial condition. 
In practice, for $k=0$ we can initialize the decentralized gradient estimation schemes with arbitrary values 
and wait for a sufficient number of iterations, until some stopping condition
of the form 
$\max_{i\in \mathcal{U}}\|[\widehat{\lder{J_S}{\mbf p_i}}]_l - [\widehat{\lder{J_S}{\mbf p_i}}]_{l-1}\|<\epsilon$ 
is reached, for some tolerance threshold $\epsilon > 0$. 
Then, for the next periods $k>0$ of the trajectory, we can use for initialization the values obtained after 
convergence at the end of the previous period $k-1$, which should lead to faster convergence.

\diffend

\subsection{Deployment of a UGV Carrying Several Anchors}

\begin{figure}[htbp!]
    \centering
    \includegraphics[width=0.5\linewidth]{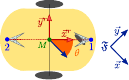}
    \caption{Robot equipped with two tags.}
    \label{fig:robotsetup}
\end{figure}

Here we illustrate
the results of Section \ref{sec:extensions_rigidity} and the performance 
difference between 
leveraging information only on relative distances or on the full relative positions.
Consider the robot shown in Fig. \ref{fig:robotsetup}, following the kinematic 
model \eqref{eq:kine_monocycle} and carrying two tags $\mathcal{U}=\{1,2\}$ placed 
at positions $\mbf p_1^r = [1,0]^\top$ and $\mbf p_2^r = [-1,0]^\top$ in the robot frame,
centered at $\mbf p_M = \frac{1}{2}(\mbf p_1+\mbf p_2)$.
Three fixed anchors $\mathcal{K}=\{3,4,5\}$ are placed at the coordinates
$\mbf p_3 = [-5, 5]^\top$, $\mbf p_4=[5, -5]^\top$ and $\mbf p_5 = [5, 5]^\top$ 
in the absolute frame. All nodes communicate and obtain range measurements with 
each other, following the Gaussian additive model \eqref{eq:model_meas} 
with $\sigma=0.1~\mathrm{m}$. The heading of the robot is $\theta$ and $\exp \cpop{\theta}$ 
is the rotation matrix between $\mathcal{F}$ and the robot frame.

\begin{figure}[h]
	\centering
	\includegraphics[width=0.65\linewidth]{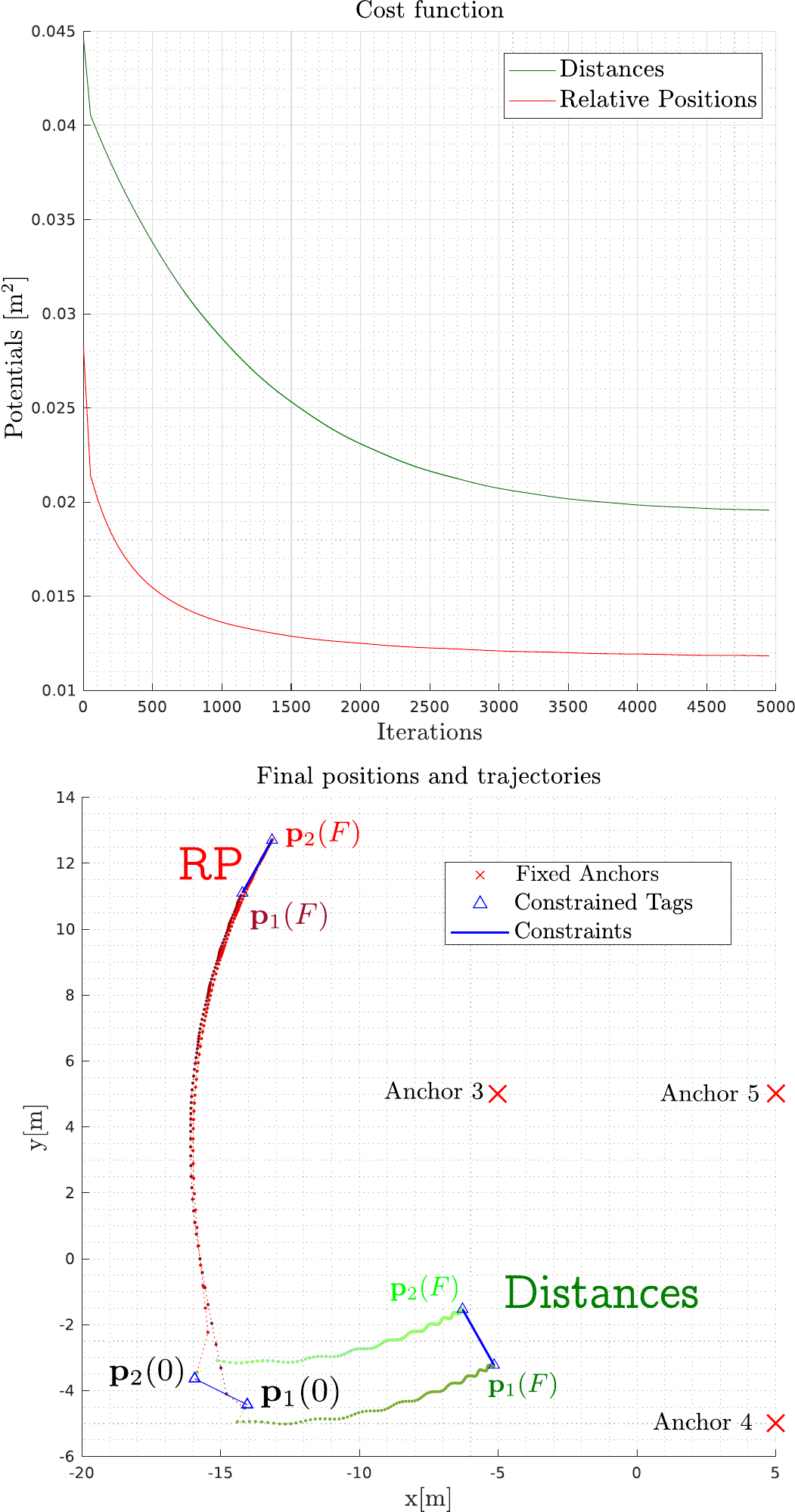}
	\caption{\diff{Deployment results for (D) and (RP) scenarios. The cost functions are plotted  as well as the positions during the trajectory.}} 
	\label{fig:robotMultitagConv}
\end{figure}

In scenario (D), we include the constraint $d_{12}=2~\mathrm{m}$ as in Section \ref{ss:ext:dist},
and define the cost function as \eqref{eq:constrained CRLB potential}. In scenario (RP), we include 
the constraint $\mbf p_{12}^r=[2,0]^\top$ as in Section \ref{ss:CRLB_RP} and define the cost
function as \eqref{eq:constrained CRLB potential - bis}, so that it can be compared to the previous one.
We compute the potentials and their derivatives with the results of 
Section \ref{sec:extensions_rigidity} and implement the scheme
\eqref{eq:lagrangian_descent} to compute a sequence of desired poses. 
\diff{The robot reaches them by using the pose controller presented in
\cite{astolfi_exponential_1999}, which includes heading control, in contrast to \eqref{eq:velcontroller}}.
At $k=0$, the initial configuration of the robot in both cases is given 
by $\mbf p_M(0) = [-15,-4]^\top$ and $\theta(0) = -\pi/8$.
The cost and robot trajectories are shown in Fig. \ref{fig:robotMultitagConv}, \diff{ denoting $F=5000$ the last iteration index of \eqref{eq:lagrangian_descent}}.
\diff{Thanks to the dual penalization of the rigidity constraint, the steady state configuration of the tags provided by \eqref{eq:lagrangian_descent}
is feasible for the robot.}

The following constrained least-squares estimators 
$\hat{\mbf p}_\mathcal{U}^\text{D}$ and $\hat{\mbf p}_\mathcal{U}^\text{RP}$ 
of $\mbf p_\mathcal{U}$ are implemented in scenarios (D) and (RP)
\[
\begin{cases}
\hat{\mbf p}_\mathcal{U}^\text{D}  
=
\underset{\hat{\mbf p}_\mathcal{U}}{\text{argmin}} ~Q(\hat{\mbf p}_\mathcal{U}), \\
\text{s.t }\hat{d}_{12}-d_{12}=0  
\end{cases}
\text{and}
\begin{cases}
\hat{\mbf p}_\mathcal{U}^\text{RP}  
=
\underset{\hat{\mbf p}_\mathcal{U}}{\text{argmin}} ~Q(\hat{\mbf p}_\mathcal{U}), \\
\text{s.t }\hat{\mbf p}_{21}- \exp \cpop{\hat \theta} \mbf p_{12}^r=\mbf 0  
\end{cases}
\]
where $\hat \theta := \text{atan2}(\hat y_{21}, \hat x_{21})$ and 
$Q(\mbf p_\mathcal{U})$ is defined in \eqref{eq:quadratic_cost}. 
We evaluate the localization performance by computing the empirical MSE $\widetilde{MSE}_{\mathcal{U},k}:=\frac{1}{2}[\widetilde{MSE}_{1,k}+\widetilde{MSE}_{2,k}]$ 
for the two tag positions, using
the same process as in \diff{Section} \ref{ss:sim:perf}, with $M = 500$ simulations. 

\begin{table}[h]
	\caption{Monte Carlo Simulation Results. \diff{Empirical MSE at the initial and
	terminal point, with $3\sigma$ confidence bounds.}} 
	\label{tab:monte}
	\centering 
	\begin{tabular}{|c|c|c|c|c|c|}
		\hline 
		 & $\widetilde{MSE}_{\mathcal{U},0}$ & Confidence & $\widetilde{MSE}_{\mathcal{U},F}$ & Confidence & ET \\ 
		\hline 
		(D) & $4.28~\mathrm{m^2}$ & $\pm 0.03~\mathrm{m^2}$ & $0.93~\mathrm{m^2}$ & $\pm 0.02~\mathrm{m^2}$ & $1.70~\mathrm{s}$ \\ 
		\hline 
		(RP) & $2.97~\mathrm{m^2}$ & $\pm 0.04~\mathrm{m^2}$ & $0.63~\mathrm{m^2}$ & $\pm 0.002~\mathrm{m^2}$ & $1.89~\mathrm{s}$\\ 
		\hline 
	\end{tabular}
\end{table}

\diff{The results shown in Table \ref{tab:monte} indicate that the motion
significantly improves the estimate accuracy in both cases: around $78\%$ for (D) and $79\%$ for (RP). Moreover, the relative
position constraints provides a clear improvement to the MSE compared to 
only using the relative distance information.}
Table \ref{tab:monte} also provides the Execution Times (ET) of the deployment algorithms 
for all the steps shown in Fig. \ref{fig:robotMultitagConv}. 
The simulation is coded in \texttt{Matlab R2018b} and runs on 
a computer equipped with an 
\texttt{Intel I7} processor.
The ET for the (RP) scenario is about 10\% larger than for (D), due to the
increased complexity to evaluate $\mbf A$ and its derivative.
In summary, compared to (D), deployment using (RP) leads to a significant improvement 
of the precision and a moderate increase of the ET.

\diffblock
\section{Experiments}
\label{sec:exp}

\begin{figure}[h]
	\centering
	\includegraphics[width=0.7\linewidth]{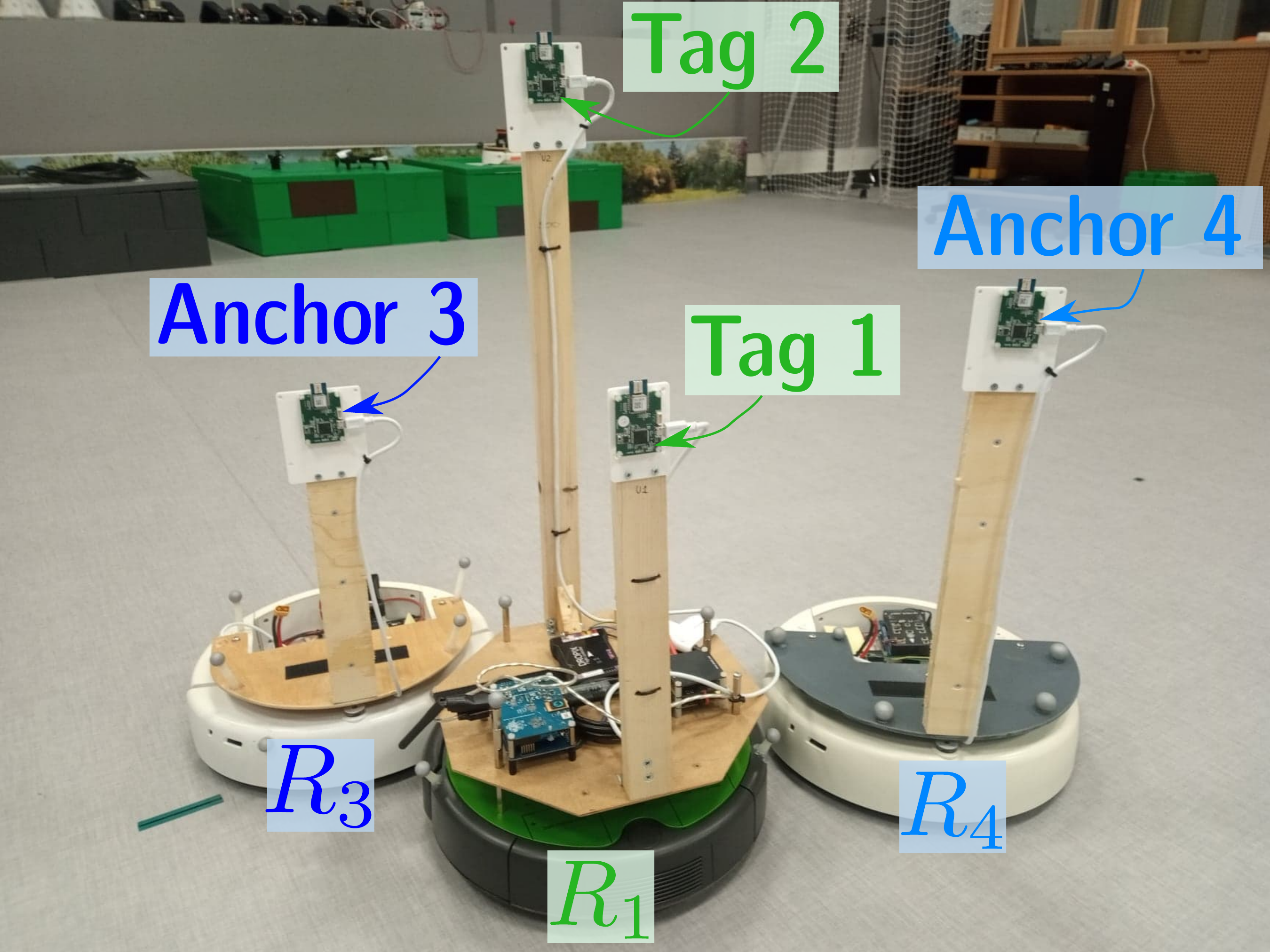}
	\diffcaption{Robots, anchors and tags.}
	\label{fig:robotsetupexp}
\end{figure}

\jln{To validate experimentally some of the ideas presented in this paper, we placed}
two tags $\mathcal{U}=\{1,2\}$ on the \emph{same} ground robot $R_1$ and two anchors $\mathcal{K}=\{3,4\}$ 
on two other robots $R_3$ and $R_4$, as shown on Fig. \ref{fig:robotsetupexp}. The anchors are externally 
positioned with a motion capture system, which is also used in the following to provide the true positions 
of the tags and evaluate the accuracy of position estimates.
The anchors and tags are based on Qorvo's \texttt{DW1000} UWB modules \cite{qorvodecawave_dwm1000_2022}.
Each tag-anchor pair $(u,k) \in \mathcal{U} \times \mathcal{K}$ is measuring its distance ${d}_{uk}$ 
using a bias-compensated
single-sided two-way ranging
protocol described in \cite{cano_clock_2022}.
The modules are placed at known height on masts, to limit signal reflections on the ground.

\begin{figure}[htbp!]
	\centering
	\includegraphics[width=\linewidth]{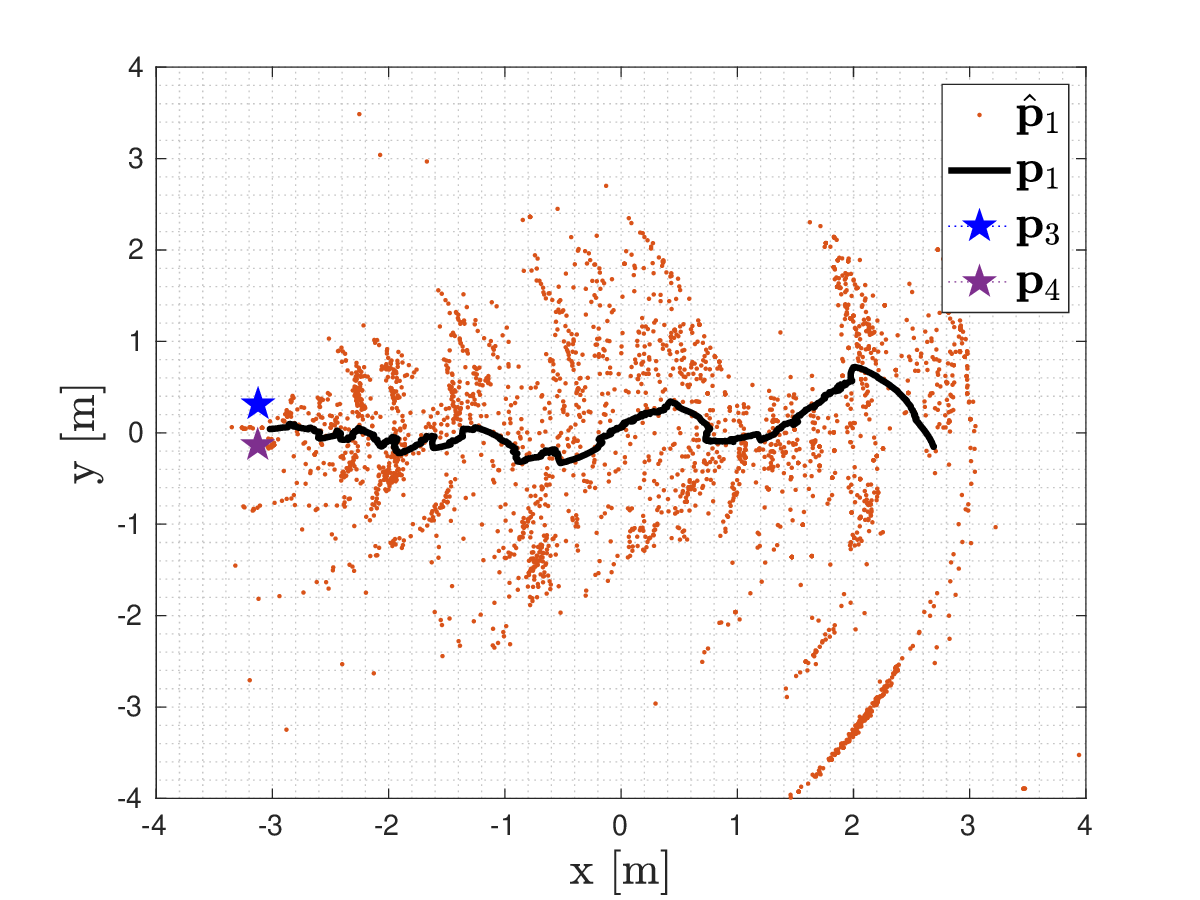}
	\diffcaption{Trajectory $\mbf p_1$ of tag $1$ and its estimates $\hat{\mbf p}_1$ 
	in the Cartesian plane while the anchors remain fixed.}
	\label{fig:nodeplxy}
	\diffblock
\end{figure}

\begin{figure}[h]
	\centering
	\includegraphics[width=\linewidth]{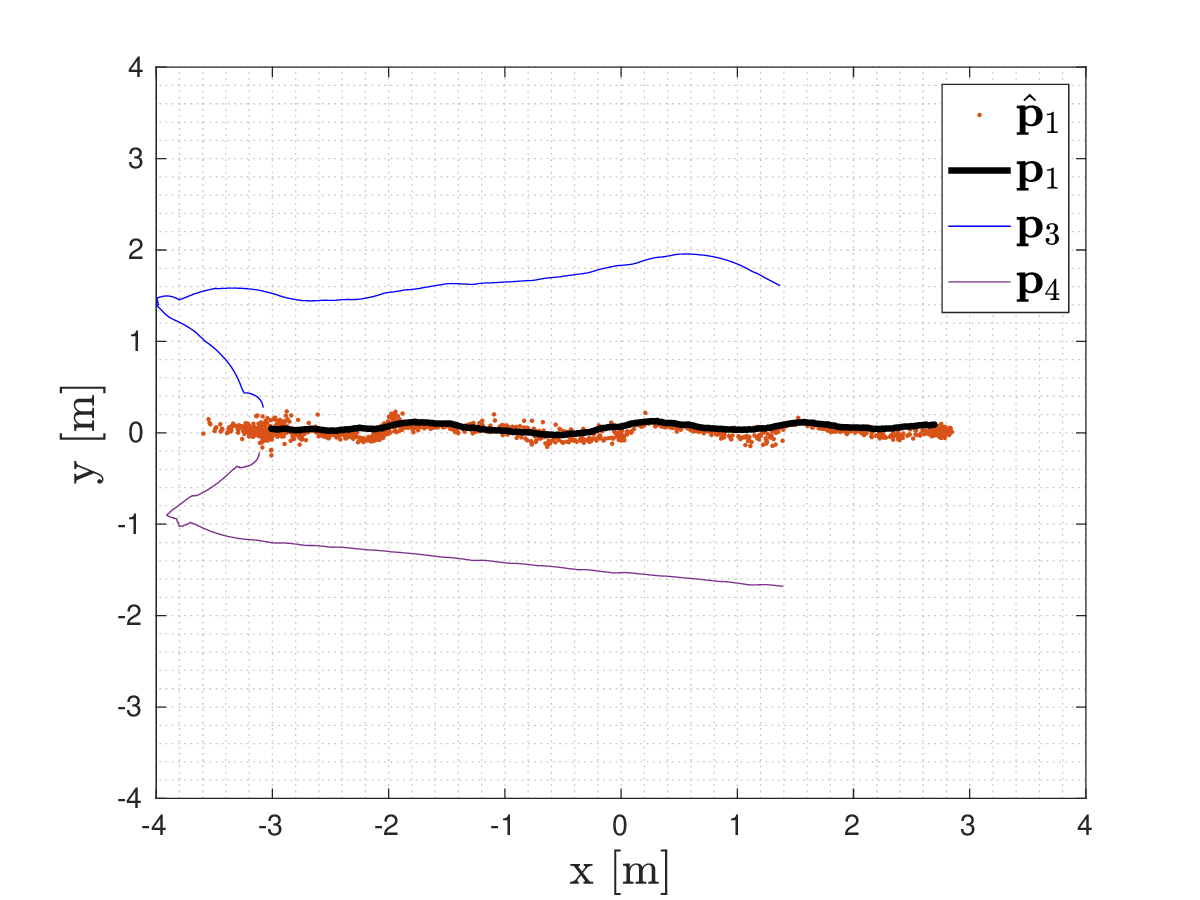}
	\diffcaption{Anchor and tag $1$ trajectories when the anchors are mobile.} 
	\label{fig:deplxy}
\end{figure}

Robot $R_1$ is initially placed at location $[-3,0]^\top$ in the world frame and 
is expected to follow the $x$-axis of that frame until reaching the neighborhood of
the final location at coordinates $[3,0]^\top$, see Fig. \ref{fig:nodeplxy} and \ref{fig:deplxy}.
To do so, the robot's position is controlled by the low-level trajectory tracking controller described 
in Section \ref{sss:motionPlannerSim}, using estimates $\hat{\mbf p}_\mathcal{U}$ of the tags' locations. 
These estimates are computed by collecting the four UWB-based ranging measurements $\tilde{d}_{uk}$ between tags 
and anchors and solving the least-squares problem
\begin{equation}
\label{eq:expLS}
\hat{\mbf p}_\mathcal{U}  = 
\underset{\mbf p_\mathcal{U} \in \mathcal{C}}{\text{argmin}} 
\sum_{k\in \mathcal{K}} \sum_{u \in \mathcal{U}} \left(||\mbf p_u - \mbf p_{k}||-\tilde{d}_{uk}\right)^2, 
\end{equation}
where 
$\mathcal{C} := \{\text{col}(\mbf p_1,\mbf p_2) \in \r^4 | \mbf p_{21} = \exp([\theta]_\times) \mbf p_{21}^1\}$
captures constraint \eqref{eq: RP constraint exp},
with the relative position $\mbf p_{21}^1 = [0.3,0]^\top$
of the tags in the robot frame centered at $\mbf p_1$ known. 
Here 
$\theta$ is the heading of $R_1$.
\jln{Note that we do not attempt to improve the location estimates \eqref{eq:expLS}
by filtering them, in order to emphasize the effect of the network geometry
on the localizability from the distance measurements alone.}

\begin{figure}[htbp!]
	\centering
	\includegraphics[width=0.7\linewidth]{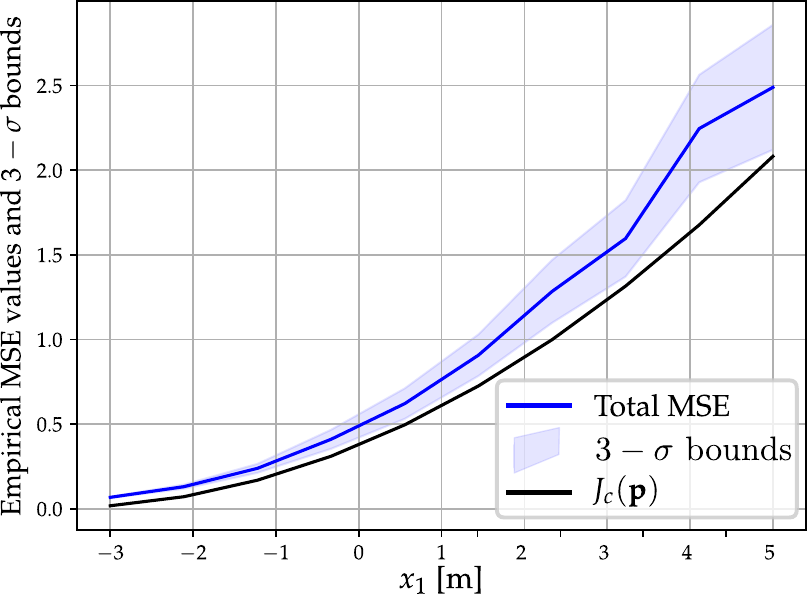}
	\diffcaption{Motion of robot $R_1$ with static anchors:
	empirical MSE \jln{of $\hat{\mbf p}_\mathcal{U}$} obtained from solving 
	\eqref{eq:expLS} $500$ times at each location, 
	and localizability potential \eqref{eq:constrained CRLB potential - bis} 
	for an ideal trajectory of $R_1$ with $y_1 = 0$.}
	\label{fig:deplmontecarlo}
\end{figure}

\begin{figure}[h]
	\centering
	\includegraphics[width=\linewidth]{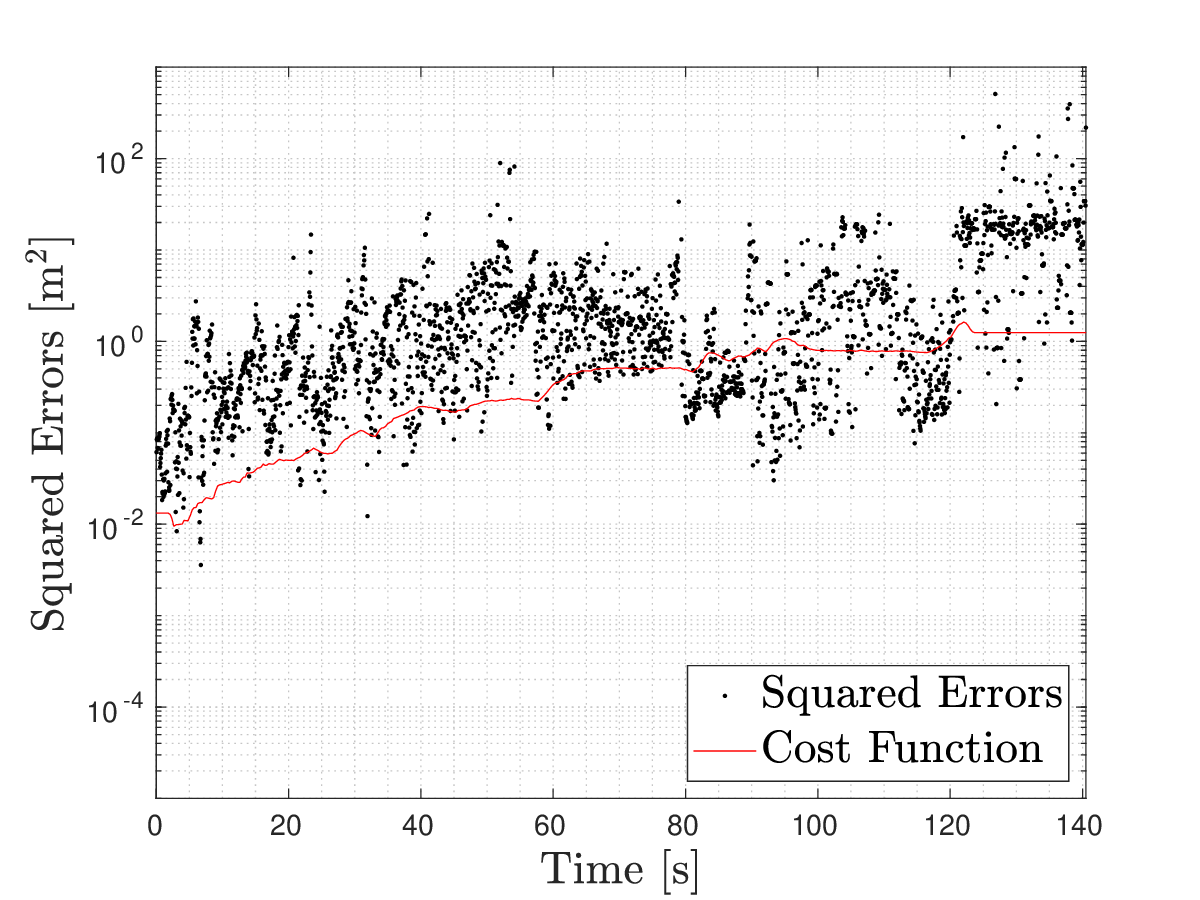}
	\diffcaption{Localizability potential and squared errors over the tags' trajectory, with fixed anchors.}
	\label{fig:nodeplcostse}
\end{figure}

First, $R_1$ attempts to follow its path 
while the anchors remain fixed at
$[-3.3,\pm 0.3]^\top$ in the Cartesian plane, as shown on Fig. \ref{fig:nodeplxy}. 
\jln{After each small motion, $R_1$ stops and repeatedly computes estimates of 
$\hat{\mathbf p}_{\mathcal U}$ using \eqref{eq:expLS}, each time using fresh measurements.
The resulting estimates for tag $1$ are shown by orange dots on Fig. \ref{fig:nodeplxy}.}
The \jln{position} estimates are increasingly noisy as $R_1$ moves toward the positive x-axis, 
with the $y$-coordinate in particular becoming increasingly uncertain.
This is intuitive because the inter-anchor distance $d_{34}$ becomes small compared 
to the measured anchor-tag distances.
The trajectory of the robot becomes increasingly erratic as a result of using
poor estimates, which motivates improving the localizability.
\jln{Although the estimates could be filtered over time to improve their accuracy and better track 
the desired path, this would lead to a slower system.}
Fig. \ref{fig:deplmontecarlo} shows in blue the empirical average MSE obtained after 
solving \eqref{eq:expLS} $500$ times, together with the $3\sigma$ confidence bounds
on this MSE value. It also shows the localizability potential $J_c$ defined 
in \eqref{eq:constrained CRLB potential - bis}, which is a theoretical lower bound on the MSE.
We see that $J_c$ predicts an increasingly poor localizability as the robot
moves toward the positive $x$-axis, which is confirmed by the empirical MSE measurements.
Fig. \ref{fig:nodeplcostse} shows the squared errors $||\hat{ \mbf p}_\mathcal{U}-\mbf p_\mathcal{U}||^2$ 
and the potential $J_c$ over the tags' trajectory, on a semi-logarithmic plot.
We note that $J_c$ is generally a good indicator of the order of magnitude of the 
expected uncertainties, which however are amplified in practice by other effects
such as multipath and non-line of sight measurements \cite{cano_kalman_2019,sahinoglu_ultra-wideband_2008}.

\diffblock

\begin{figure}[h]
	\centering
	\includegraphics[width=\linewidth]{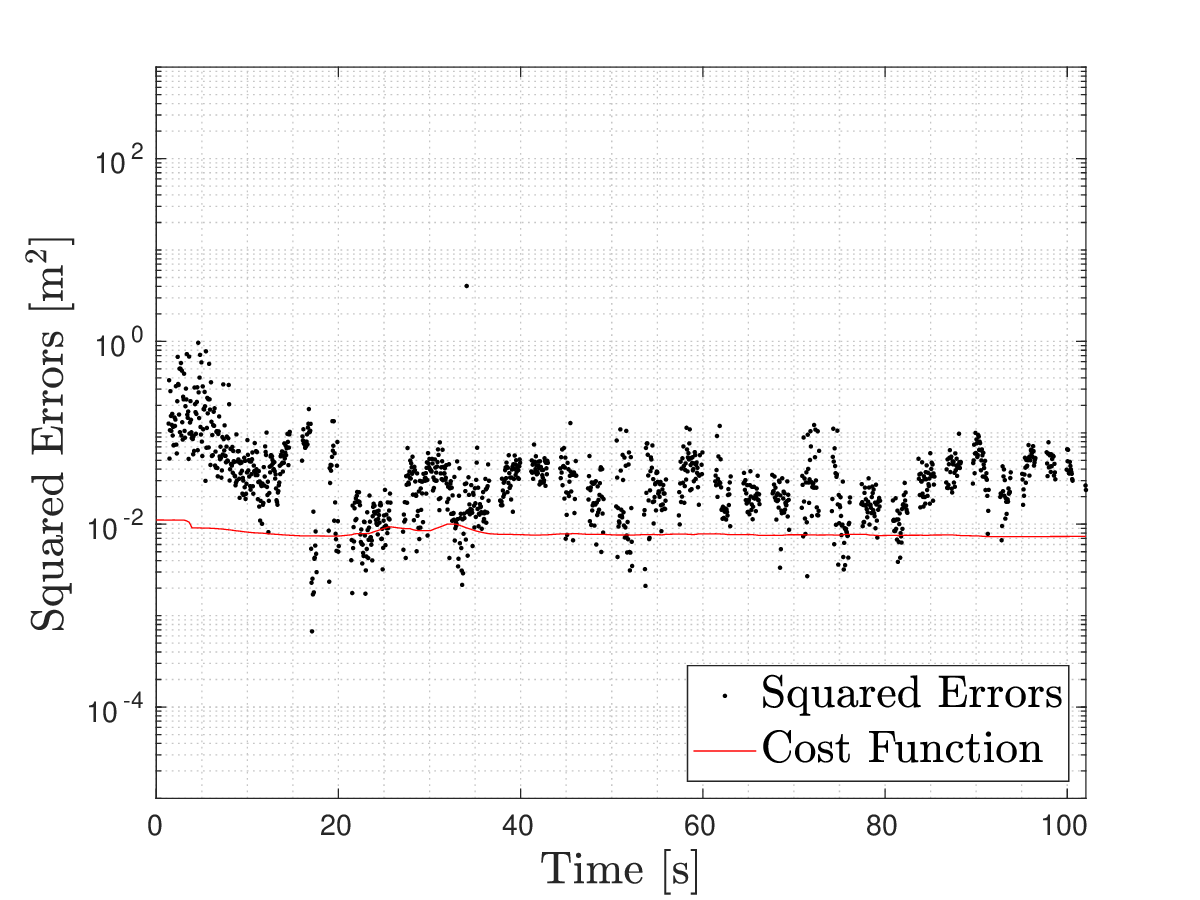}
	\diffcaption{Localizability and squared errors during deployment.}
	\label{fig:deplcostse}
\end{figure}

\diffblock
Next, we illustrate on Fig. \ref{fig:deplxy} the trajectory tracking results when the anchors are deployed 
simultaneously with $R_1$, using the gradient descent scheme described in Section \ref{ss:CRLB_RP}, with the 
gradient expression \eqref{eq:derJcRP}.
In this case, the position estimates produced by \eqref{eq:expLS} exhibit much less variance,
which is confirmed also by Fig. \ref{fig:deplcostse}. This figure also shows that the
localizability potential $J_c$ is kept at a much lower value during the motion.
The reduced variance allows us to efficiently reject measurement outliers and maintain
an empirical MSE of about $12$ cm along the trajectory, which is appropriate for
indoor navigation.
Hence, this experiment highlights that 
localizability can be improved automatically in real-time, even when using the position estimates 
to replace the true position in the gradient-based deployment algorithm.

\diffend

\section{Conclusion and Perspectives}

This paper presents deployment methods applicable to Multi Robots Systems (MRS)
with relative distance measurements, which \diff{maximize} localizability.
Constrained Cram\'er-Rao Lower Bounds (CRLB) are used to predict the localization error
of a given configuration, assuming Gaussian ranging measurement models. 
A connection between Fisher information matrices and rigidity matrices is highlighted, 
which yields useful \diff{invertibility} properties, e.g., for initial MRS placement.  

The CRLB is used to design artificial potentials,
so that gradient descent schemes can be developed to plan robot
motions that enhance the overall localizability of the network.
Moreover, we show how to distribute the execution of the \diff{gradient estimation}
algorithms among the robots, so that they only need to communicate 
with their neighbors in the ranging graph.
Finally, we extend the methodology to MRS with robots carrying multiple tags, 
again leveraging the theory of equality-constrained CRLBs.
Future work 
\diff{could consider also optimizing the network topology, since maintaining ranging links typically
entails a cost (consuming bandwidth, computation resources, etc.). 
Developing formal closed-loop stability properties for the gradient-based control law
with noisy position estimates is also of interest.
}

\section*{Acknowledgements}

The authors thank Drs. \'Eric Chaumette, Ga\"el Pag\`es and Ali Naouri 
from ISAE-Supa\'ero (France) for helpful discussions.

\bibliographystyle{IEEEtran}
\bibliography{ms}

\begin{thebibliography}{10}
\providecommand{\url}[1]{#1}
\csname url@samestyle\endcsname
\providecommand{\newblock}{\relax}
\providecommand{\bibinfo}[2]{#2}
\providecommand{\BIBentrySTDinterwordspacing}{\spaceskip=0pt\relax}
\providecommand{\BIBentryALTinterwordstretchfactor}{4}
\providecommand{\BIBentryALTinterwordspacing}{\spaceskip=\fontdimen2\font plus
\BIBentryALTinterwordstretchfactor\fontdimen3\font minus
  \fontdimen4\font\relax}
\providecommand{\BIBforeignlanguage}[2]{{%
\expandafter\ifx\csname l@#1\endcsname\relax
\typeout{** WARNING: IEEEtran.bst: No hyphenation pattern has been}%
\typeout{** loaded for the language `#1'. Using the pattern for}%
\typeout{** the default language instead.}%
\else
\language=\csname l@#1\endcsname
\fi
#2}}
\providecommand{\BIBdecl}{\relax}
\BIBdecl

\bibitem{le_ny_jerome_localizability-constrained_2018}
J.~{Le Ny} and S.~Chauvi{\`e}re, ``Localizability-constrained deployment of
  mobile robotic networks with noisy range measurements,'' in \emph{American
  Control Conference (ACC)}, Milwaukee, WI, Jun. 2018, pp. 2788--2793.

\bibitem{Cano:ICRA21:constrainedCRLB}
J.~Cano and J.~Le~Ny, ``\BIBforeignlanguage{en}{Improving ranging-based
  location estimation with rigidity-constrained {CRLB}-based motion
  planning},'' in \emph{\BIBforeignlanguage{en}{IEEE Internationnal Conference
  on Robotics and Automation (ICRA)}}, Xi'An (China), 2021.

\bibitem{sheu_distributed_2010}
J.~Sheu, W.~Hu, and J.~Lin, ``Distributed localization scheme for mobile sensor
  networks,'' \emph{IEEE Transactions on Mobile Computing}, vol.~9, no.~4, pp.
  516--526, Apr. 2010.

\bibitem{Prorok:ICRA12:relativeLoc}
A.~Prorok, A.~Bahr, and A.~Martinoli, ``Low-cost collaborative localization for
  large-scale multi-robot systems,'' in \emph{IEEE International Conference on
  Robotics and Automation (ICRA)}, Saint Paul, MN, USA, 2012.

\bibitem{xu_aoa_2008}
J.~Xu, M.~Ma, and C.~L. Law, ``{AOA} cooperative position localization,'' in
  \emph{{IEEE} Global Telecommunications Conference ({GLOBECOM})}, New Orleans,
  LA, Nov. 2008.

\bibitem{wei_noisy_2015}
M.~Wei, R.~Aragues, C.~Sagues, and G.~C. Calafiore, ``Noisy range network
  localization based on distributed multidimensional scaling,'' \emph{IEEE
  Sensors}, vol.~15, no.~3, pp. 1872--1883, Mar. 2015.

\bibitem{carlino_robust_2018}
L.~Carlino, D.~Jin, M.~Muma, and M.~Zoubir, ``Robust distributed cooperative
  {RSS}-based localization for directed graphs in mixed {LoS/NLoS}
  environments,'' \emph{EURASIP Journal on Wireless Communications and
  Networking}, vol. 2019, Jan. 2019.

\bibitem{sahinoglu_ultra-wideband_2008}
Z.~Sahinoglu, S.~Gezici, and I.~Guvenc,
  \emph{\BIBforeignlanguage{en}{Ultra-wideband Positioning Systems: Theoretical
  Limits, Ranging Algorithms, and Protocols}}.\hskip 1em plus 0.5em minus
  0.4em\relax Cambridge: Cambridge University Press, 2008.

\bibitem{mueller_fusing_2015}
M.~W. Mueller, M.~Hamer, and R.~D'Andrea, ``\BIBforeignlanguage{en}{Fusing
  ultra-wideband range measurements with accelerometers and rate gyroscopes for
  quadrocopter state estimation},'' in \emph{\BIBforeignlanguage{en}{IEEE
  International Conference on Robotics and Automation ({ICRA})}}, Seattle, WA,
  USA, May 2015, pp. 1730--1736.

\bibitem{cano_kalman_2019}
J.~Cano, S.~Chidami, and J.~{Le Ny}, ``\BIBforeignlanguage{en}{A {K}alman
  filter-based algorithm for simultaneous time synchronization and localization
  in {UWB} networks},'' in \emph{\BIBforeignlanguage{en}{IEEE International
  Conference on Robotics and Automation ({ICRA})}}, Montreal, QC, Canada, May
  2019.

\bibitem{Buehrer:IEEE18:collaborativeLocSurvey}
R.~M. Buehrer, H.~Wymeersch, and R.~M. Vaghefi, ``Collaborative sensor network
  localization: Algorithms and practical issues,'' \emph{Proceedings of the
  IEEE}, vol. 106, no.~6, pp. 1089--1114, 2018.

\bibitem{tay_generating_1985}
T.-S. Tay and W.~Whiteley, ``Generating isostatic frameworks,''
  \emph{Structural Topology}, vol.~11, Jan. 1985.

\bibitem{cao_ratio--distance_2020}
K.~Cao, Z.~Han, X.~Li, and L.~Xie, ``Ratio-of-{Distance} {Rigidity} {Theory}
  {With} {Application} to {Similar} {Formation} {Control},'' \emph{IEEE
  Transactions on Automatic Control}, vol.~65, no.~6, pp. 2598--2611, Jun.
  2020.

\bibitem{Aspnes:TMC06:theoryLoc}
J.~Aspnes, T.~Eren, D.~K. Goldenberg, A.~S. Morse, W.~Whiteley, Y.~R. Yang,
  B.~D. Anderson, and P.~N. Belhumeur, ``A theory of network localization,''
  \emph{IEEE Transactions on Mobile Computing}, vol.~5, no.~12, pp. 1663--1678,
  2006.

\bibitem{patwari_locating_2005}
N.~Patwari, J.~N. Ash, S.~Kyperountas, A.~O. Hero, R.~L. Moses, and N.~S.
  Correal, ``Locating the nodes: cooperative localization in wireless sensor
  networks,'' \emph{IEEE Signal Processing Magazine}, vol.~22, no.~4, pp.
  54--69, Jul. 2005.

\bibitem{groves_principles_2013}
P.~D. Groves, \emph{Principles of {GNSS}, inertial, and multisensor integrated
  navigation systems}, 2nd~ed.\hskip 1em plus 0.5em minus 0.4em\relax Artech
  House, 2013.

\bibitem{Kim:TAC06:maximizingEigenvalue}
Y.~Kim and M.~Mesbahi, ``On maximizing the second smallest eigenvalue of a
  state-dependent graph {L}aplacian,'' \emph{IEEE Transactions on Automatic
  Control}, vol.~51, no.~1, p. 117, 2006.

\bibitem{siciliano_maintaining_2009}
N.~Michael, M.~M. Zavlanos, V.~Kumar, and G.~J. Pappas,
  ``\BIBforeignlanguage{en}{Maintaining connectivity in mobile robot
  networks},'' in \emph{\BIBforeignlanguage{en}{Experimental {Robotics}}},
  B.~Siciliano, O.~Khatib, F.~Groen, O.~Khatib, V.~Kumar, and G.~J. Pappas,
  Eds.\hskip 1em plus 0.5em minus 0.4em\relax Springer, 2009, vol.~54, pp.
  117--126, {S}pringer Tracts in Advanced Robotics.

\bibitem{decentralized_2010}
P.~Yang, R.~Freeman, G.~Gordon, K.~Lynch, S.~Srinivasa, and R.~Sukthankar,
  ``\BIBforeignlanguage{en}{Decentralized estimation and control of graph
  connectivity for mobile sensor networks},''
  \emph{\BIBforeignlanguage{en}{Automatica}}, vol.~46, pp. 390--396, 2010.

\bibitem{Shames:Automatica09:LocMinimization}
I.~Shames, B.~Fidan, and B.~D. Anderson, ``Minimization of the effect of noisy
  measurements on localization of multi-agent autonomous formations,''
  \emph{Automatica}, vol.~45, no.~4, pp. 1058--1065, 2009.

\bibitem{zelazo_rigidity_2012}
D.~Zelazo, A.~Franchi, P.~Allgöwer, H.~Bülthoff, and P.~Robuffo~Giordano,
  ``Rigidity maintenance control for multi-robot systems,'' in \emph{Robotics:
  Science and Systems VIII}, Jul. 2012.

\bibitem{zelazo_decentralized_2015}
D.~Zelazo, A.~Franchi, H.~H. Bülthoff, and P.~Robuffo~Giordano,
  ``\BIBforeignlanguage{en}{Decentralized rigidity maintenance control with
  range measurements for multi-robot systems},''
  \emph{\BIBforeignlanguage{en}{The International Journal of Robotics
  Research}}, vol.~34, no.~1, pp. 105--128, Jan. 2015.

\bibitem{sun_distributed_2015}
Z.~Sun, C.~Yu, and B.~D.~O. Anderson, ``Distributed optimization on proximity
  network rigidity via robotic movements,'' in \emph{{Chinese} {Control}
  {Conference} ({CCC})}, Hangzhou, China, Jul. 2015, pp. 6954--6960.

\bibitem{haug_bayesian_2012}
A.~J. Haug, \emph{\BIBforeignlanguage{en}{Bayesian estimation and tracking: a
  practical guide}}.\hskip 1em plus 0.5em minus 0.4em\relax Wiley, 2012.

\bibitem{choset_principles_2005}
H.~Choset, K.~Lynch, S.~Hutchinson, K.~George, W.~Burgard, L.~Kavraki, and
  S.~Thrun, \emph{Principles of Robot Motion}.\hskip 1em plus 0.5em minus
  0.4em\relax The MIT Press, 2005.

\bibitem{mcaulay_barankin_1971}
R.~McAulay and E.~Hofstetter, ``{B}arankin bounds on parameter estimation,''
  \emph{IEEE Transactions on Information Theory}, vol.~17, no.~6, pp. 669--676,
  Nov. 1971.

\bibitem{Ucinski:book04:optimalSensing}
D.~Uci{\'n}ski, \emph{Optimal measurement methods for distributed parameter
  system identification}.\hskip 1em plus 0.5em minus 0.4em\relax CRC press,
  2004.

\bibitem{pukelsheim_optimal_2006}
F.~Pukelsheim, \emph{\BIBforeignlanguage{en}{Optimal {Design} of
  {Experiments}}}.\hskip 1em plus 0.5em minus 0.4em\relax SIAM, 2006.

\bibitem{LeNy:CDC09:activeSensingGP}
J.~Le~Ny and G.~J. Pappas, ``On trajectory optimization for active sensing in
  {G}aussian process models,'' in \emph{IEEE Conference on Decision and Control
  (CDC)}, Shanghai, China, 12 2009.

\bibitem{carrillo_comparison_2012}
H.~Carrillo, I.~Reid, and J.~A. Castellanos, ``\BIBforeignlanguage{en}{On the
  comparison of uncertainty criteria for active {SLAM}},'' in
  \emph{\BIBforeignlanguage{en}{{IEEE} {International} {Conference} on
  {Robotics} and {Automation} (ICRA)}}, St Paul, MN, USA, May 2012, pp.
  2080--2087.

\bibitem{hero_lower_1990}
J.~D. Gorman and A.~O. Hero, ``Lower bounds for parametric estimation with
  constraints,'' \emph{IEEE Transactions on Information Theory}, vol.~36,
  no.~6, pp. 1285--1301, 1990.

\bibitem{bonnabel_intrinsic_2015}
S.~Bonnabel and A.~Barrau, ``An intrinsic {C}ram{\'e}r-{R}ao bound on lie
  groups,'' in \emph{Geometric Science of Information}.\hskip 1em plus 0.5em
  minus 0.4em\relax Springer International Publishing, 2015, pp. 664--672.

\bibitem{chirikjian_wirtinger_2018}
G.~S. Chirikjian, ``\BIBforeignlanguage{en}{From {W}irtinger to {F}isher
  information inequalities on spheres and rotation groups},'' in
  \emph{\BIBforeignlanguage{en}{IEEE {International} {Conference} on
  {Information} {Fusion} ({FUSION})}}, Cambridge, United Kingdom, Jul. 2018,
  pp. 730--736.

\bibitem{Bensky:book16:wirelesPos}
A.~Bensky, \emph{Wireless positioning technologies and applications},
  2nd~ed.\hskip 1em plus 0.5em minus 0.4em\relax Artech House, 2016.

\bibitem{coulson_statistical_1998}
A.~J. Coulson, A.~G. Williamson, and R.~G. Vaughan, ``A statistical basis for
  lognormal shadowing effects in multipath fading channels,'' \emph{IEEE
  Transactions on Communications}, vol.~46, no.~4, pp. 494--502, Apr. 1998.

\bibitem{Khatib:art86:potentials}
O.~Khatib, ``Real-time obstacle avoidance for manipulators and mobile robots,''
  in \emph{Autonomous robot vehicles}.\hskip 1em plus 0.5em minus 0.4em\relax
  Springer, 1986, pp. 396--404.

\bibitem{Bullo:book09:distributedRobotics}
F.~Bullo, J.~Cort{\'e}s, and S.~Martinez, \emph{Distributed control of robotic
  networks: a mathematical approach to motion coordination algorithms}.\hskip
  1em plus 0.5em minus 0.4em\relax Princeton University Press, 2009.

\bibitem{Cano:IROS22:distanceCRLB}
J.~Cano, G.~Pages, E.~Chaumette, and J.~{Le Ny}, ``Optimal localizability
  criterion for positioning with distance-deteriorated relative measurements,''
  in \emph{International Conference on Intelligent Robots and Systems (IROS)},
  Kyoto, Japan, October 2022.

\bibitem{kay_fundamentals_1993}
S.~M. Kay, \emph{Fundamentals of {Statistical} {Signal} {Processing}:
  {Estimation} {Theory}}.\hskip 1em plus 0.5em minus 0.4em\relax Englewood
  Cliffs, NJ, USA: Prentice-Hall, 1993.

\bibitem{petersen_matrix_2012}
K.~B. Petersen and M.~S. Pedersen, ``\BIBforeignlanguage{en}{The matrix
  cookbook},'' Technical University of Denmark, Tech. Rep., 2012.

\bibitem{bonin_matroids_1996}
W.~Whiteley, ``\BIBforeignlanguage{en}{Some matroids from discrete applied
  geometry},'' in \emph{\BIBforeignlanguage{en}{Contemporary {Mathematics}}},
  J.~E. Bonin, J.~G. Oxley, and B.~Servatius, Eds.\hskip 1em plus 0.5em minus
  0.4em\relax Providence, Rhode Island: American Mathematical Society, 1996,
  vol. 197, pp. 171--311.

\bibitem{Godsil:book01:algebraicGraphTheory}
C.~Godsil and G.~F. Royle, \emph{Algebraic graph theory}.\hskip 1em plus 0.5em
  minus 0.4em\relax Springer, 2001.

\bibitem{Moore:Sensys04:networkLoc}
D.~Moore, J.~Leonard, D.~Rus, and S.~Teller, ``Robust distributed network
  localization with noisy range measurements,'' in \emph{International
  Conference on Embedded Networked Sensor Systems}, Baltimore, MD, USA, 2004,
  p. 50–61.

\bibitem{harville_matrix_1997}
D.~A. Harville, \emph{\BIBforeignlanguage{en}{Matrix {Algebra} from a
  {Statistician's} {Perspective}}}.\hskip 1em plus 0.5em minus 0.4em\relax New
  York, NY: Springer, 1997.

\bibitem{Bertsekas:2015:parallel}
D.~Bertsekas and J.~Tsitsiklis, \emph{Parallel and distributed computation:
  numerical methods}.\hskip 1em plus 0.5em minus 0.4em\relax Athena Scientific,
  2015.

\bibitem{Kia:CSM19:dynamicConsensus}
S.~S. Kia, B.~Van~Scoy, J.~Cortes, R.~A. Freeman, K.~M. Lynch, and S.~Martinez,
  ``Tutorial on dynamic average consensus: The problem, its applications, and
  the algorithms,'' \emph{IEEE Control Systems Magazine}, vol.~39, no.~3, pp.
  40--72, 2019.

\bibitem{dimitri_p_bertsekas_nonlinear_2016}
D.~P. Bertsekas, \emph{\BIBforeignlanguage{en}{Nonlinear programming}},
  3rd~ed.\hskip 1em plus 0.5em minus 0.4em\relax Belmont, MA: Athena
  Scientific, 2016.

\bibitem{Barfoot:book17:stateEst}
T.~D. Barfoot, \emph{State estimation for robotics}.\hskip 1em plus 0.5em minus
  0.4em\relax Cambridge University Press, 2017.

\bibitem{Lynch2017ModernRM}
K.~Lynch and F.~Park, \emph{Modern Robotics: Mechanics, Planning, and
  Control}.\hskip 1em plus 0.5em minus 0.4em\relax Cambridge University Press,
  2017.

\bibitem{gallego_compact_2015}
G.~Gallego and A.~Yezzi, ``\BIBforeignlanguage{en}{A compact formula for the
  derivative of a 3-{D} rotation in exponential coordinates},''
  \emph{\BIBforeignlanguage{en}{Journal of Mathematical Imaging and Vision}},
  vol.~51, no.~3, pp. 378--384, Mar. 2015.

\bibitem{mai_local_2018}
V.~Mai, M.~Kamel, M.~Krebs, A.~Schaffner, D.~Meier, L.~Paull, and a.~R.
  Siegwart, ``Local positioning system using {UWB} range measurements for an
  unmanned blimp,'' \emph{IEEE Robotics and Automation Letters}, vol.~3, no.~4,
  pp. 2971--2978, Oct. 2018.

\bibitem{prorok_models_2013}
A.~Prorok, ``Models and {Algorithms} for {Ultra}-{Wideband} {Localization} in
  {Single}- and {Multi}-{Robot} {Systems},'' {PhD} {Thesis}, Ecole
  Polytechnique Fererale de Lausanne (EPFL), 2013.

\bibitem{corke_robotics_2011}
P.~Corke, \emph{\BIBforeignlanguage{en}{Robotics, {Vision} and {Control}}},
  ser. Springer {Tracts} in {Advanced} {Robotics}, B.~Siciliano and O.~Khatib,
  Eds.\hskip 1em plus 0.5em minus 0.4em\relax Springer, 2011, vol.~73.

\bibitem{astolfi_exponential_1999}
A.~Astolfi, ``\BIBforeignlanguage{en}{Exponential stabilization of a wheeled
  mobile robot via discontinuous control},''
  \emph{\BIBforeignlanguage{en}{Journal of Dynamic Systems, Measurement, and
  Control}}, vol. 121, no.~1, pp. 121--126, Mar. 1999.

\bibitem{qorvodecawave_dwm1000_2022}
\BIBentryALTinterwordspacing
\emph{{DWM1000} {datasheet}}, Qorvo (formerly Decawave), 2022. [Online].
  Available: \url{https://www.qorvo.com/products/p/DWM1000}
\BIBentrySTDinterwordspacing

\bibitem{cano_clock_2022}
J.~Cano, G.~Pages, E.~Chaumette, and J.~{Le Ny}, ``Clock and {Power}-{Induced}
  {Bias} {Correction} for {UWB} {Time}-of-{Flight} {Measurements},'' \emph{IEEE
  Robotics and Automation Letters}, pp. 2431--2438, 2022.

\end{thebibliography}

\end{document}